\documentclass{jair}

\acmDOI{10.1613/jair.1.18890}

\JAIRAE{Alessandro Farinelli}
\JAIRTrack{} 
\acmVolume{83}
\acmArticle{25}
\acmMonth{8}
\acmYear{2025}

\usepackage{booktabs}
\usepackage{enumitem}

\usepackage{graphicx}
\usepackage{color}
\usepackage{caption}
\usepackage{tikz}

\usepackage[verbose]{placeins}
\usepackage{float}
\floatstyle{plaintop}
\restylefloat{table}

\newcommand{\sigmoid}{\mathsf{sig}}

\begin{document}


\title[Learning How to Vote with Principles]{Learning How to Vote with Principles: Axiomatic Insights Into the Collective Decisions of Neural Networks}


\author{Levin Hornischer}
\authornote{Corresponding Author.}
\orcid{0000-0003-2087-527X}
\email{levin.hornischer@lmu.de}
\affiliation{%
  \institution{Munich Center for Mathematical Philosophy, LMU Munich}
  \city{Munich}
  \country{Germany}
}

\author{Zoi Terzopoulou}
\orcid{0000-0001-5289-434X}
\email{zoi.terzopoulou@cnrs.fr}
\affiliation{%
  \institution{GATE, CNRS, Universit\'e Jean Monnet, Universit\'e Lumiere Lyon 2}
  \city{Saint-Etienne}
  \country{France}
}

\renewcommand{\shortauthors}{Hornischer \& Terzopoulou}

\begin{abstract}
Can neural networks be applied in voting theory, while satisfying the need for transparency in collective decisions? We propose \emph{axiomatic deep voting}: a framework to build and evaluate neural networks that aggregate preferences, using the well-established axiomatic method of voting theory.
Our findings are: 
(1)~Neural networks, despite being highly accurate, often fail to align with the core axioms of voting rules, revealing a disconnect between mimicking outcomes and reasoning. 
(2)~Training with axiom-specific data does not enhance alignment with those axioms. 
(3)~By solely optimizing axiom satisfaction, neural networks can synthesize new voting rules that often surpass and substantially differ from existing ones.
This offers insights for both fields: For AI, important concepts like bias and value-alignment are studied in a mathematically rigorous way; for voting theory, new areas of the space of voting rules are explored.
\end{abstract}

\received{21 October 2024}
\received[revised]{16 April 2025}
\received[accepted]{14 June 2025}

\maketitle
 

\section{Introduction}
\label{sec: introduction}

Artificial intelligence (AI) is increasingly applied in many domains, including not just scientific and technological but also societal problems. This poses a dilemma when it comes to \emph{social choice}, i.e., voting, preference aggregation, and other processes of collective decisions. On the one hand, voting systems should be transparent, but the neural networks on which modern AI is built are notoriously opaque. On the other hand, neural networks could unearth novel and tailor-made collective decision procedures.
Already, state-of-the-art techniques for alignment of Large Language Models (LLMs) with human values---like RLHF\footnote{Bai et al., 2022. Training a Helpful and Harmless Assistant with Reinforcement Learning from Human Feedback. arXiv:2204.05862.} or DPO~\citep{Rafailov2024}---rely on the aggregation of human preferences about the generated outputs to fine-tune LLMs. This triggered recent research in guiding such AI alignment using social choice~\citep{conitzerposition}.

In this paper, we study how neural networks aggregate votes and preferences. When they form such collective decisions, do they adhere to the normative principles that social choice theory formulates as axioms? 
This is fundamental both for a discussion of the dilemma and for using social choice for AI alignment. Moreover, it offers new insights for both AI and voting theory.
For AI, this provides a rich testing ground to study pressing machine learning concepts like bias, value-alignment and interpretability in a mathematically rigorous way. For example, a network is not biased towards specific individuals if it aggregates their preferences in accordance with the axiom of anonymity; the so-called Pareto principle requires the neural network to align with any preference shared among all individuals; and the well-known axiom of independence entails a certain compositional interpretability of the network. For voting theory, axiomatic deep voting provides a new method for the central quest of exploring the space of voting rules.

\begin{figure}
\centering
\includegraphics[width=.7\linewidth]{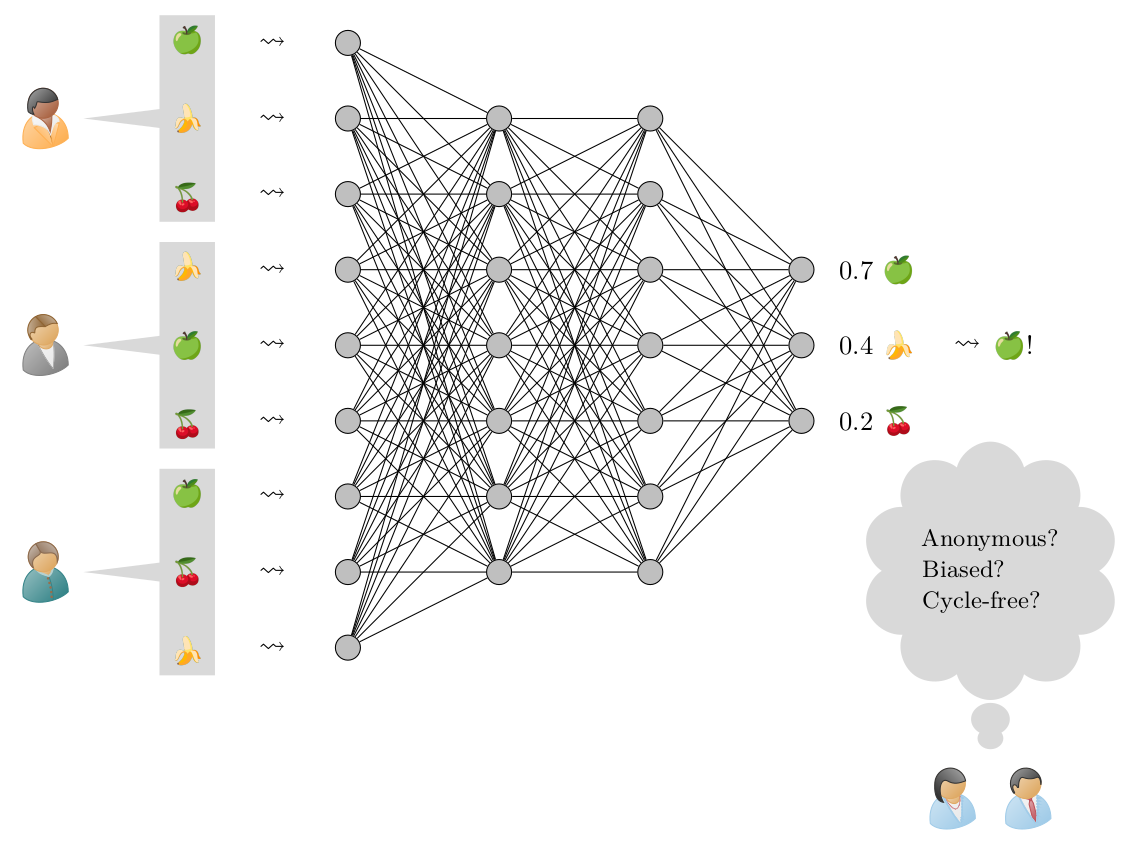}
\Description{A stylized neural network with preferences of three agents inputted on the left and a winning set outputted on the right.}
\caption{Can neural networks learn to vote with principles?}
\label{fig: overview of paper}
\end{figure}

\paragraph{Social choice.} 
How are individual preferences best turned into a collective decision? This question is studied by \emph{social choice theory} \citep{HBCOMSOC2016, List2022}, and specifically \emph{voting theory} \citep{zwicker2016introduction}. A \emph{voting rule} is a function that takes as input a \emph{profile}---i.e., a list of each individual's preferences among a set of alternatives---and produces as output a \emph{collective decision}, i.e., the alternative(s) that the rule takes to be most preferred for the group as a whole (see Section~\ref{sec: preliminaries} for the formal definitions). The most straightforward rule is \emph{Plurality} (picking the top alternative of most individuals); other classic rules include \emph{Borda} and \emph{Copeland}, while a recent suggestion is \emph{Stable Voting}.

\paragraph{Axiomatic deep voting.} 
To study the collective choices of neural networks, we develop the \emph{axiomatic deep voting} framework (sketched in Figure~\ref{fig: overview of paper}).\footnote{The source code is available here: \url{https://github.com/LevinHornischer/AxiomaticDeepVoting}.}
Deep neural networks are (parametrized) functions that map vectors (typically of a high dimension) to vectors (typically of a low dimension). So, after suitably \emph{encoding} profiles and collective decisions as vectors, neural networks realize voting rules, i.e., functions from profiles to collective decisions.
Discovering a voting rule can then be seen as an \emph{optimization problem}: updating the neural network parameters until a given desired property is fulfilled.
We \emph{evaluate} a trained neural network in terms of accuracy and axiom satisfaction. While the former is standard in machine learning, the latter is specific to voting theory and its \emph{axiomatic method}~\citep{Thomson2001, List2011}. Different axioms describe different desirable properties of voting rules. An example is the already mentioned anonymity axiom which requires that the names of the voters should \emph{not} influence the collective decision.

\paragraph{Research questions.} With this framework, we investigate the general question of how neural networks aggregate preferences via three more specific questions.
\begin{itemize}
\item[(1)] 
\emph{Correct for the right reasons?} 
Neural networks can accurately learn standard voting rules, but do they adhere to the normative principles expressed by voting axioms?
\end{itemize}
We observe eminent violations of the axioms, despite high accuracy in mimicking voting rules. So we focus on teaching neural networks the expert knowledge expressed by axioms. There are two common ways to do this. The first is via \emph{dataset augmentation} \citep{xia2013designing}: 
\begin{itemize}
\item[(2)]
\emph{Learning principles by example?}
Can neural networks be trained to adhere to voting axioms by training with data exemplifying the axioms? 
\end{itemize}
The second way is via \emph{semantic loss functions} \citep{Xu2018}. For this, we develop a translation of the axioms into loss functions; so, by optimizing this loss during training, the network increases the corresponding axiom satisfaction. 
Importantly, though, perfect axiom satisfaction is impossible according to the infamous theorem by \citet{arrow1951}. So we search for the best possible axiom satisfaction:
\begin{itemize}
\item[(3)] 
\emph{Rule synthesis guided by principles?} 
When neural networks optimize axiom satisfaction, can they develop new voting rules that surpass existing ones?
\end{itemize}
We compare the discovered rules to a wide range of known voting rules, to test if neural networks can advance the current state of the art in voting theory.

\paragraph{Key findings.}
In three experiments, we answer these questions in turn. In each, we test three paradigmatic neural network architectures: multi-layer perceptrons, convolutional neural networks, and word embedding based classifiers. We also check a variety of standard distributions of voter preferences. 
Our three experiments find, respectively:
\begin{enumerate}
\item[(1)]
The employed architectures demonstrate similar behavior both regarding accuracy and axiom satisfaction. Importantly, despite high accuracy, they markedly violate critical axioms like anonymity---yet, the news is not as bad for other axioms. 
\item[(2)]
Data augmentation does not seem to boost the principled learning of neural networks. However, it drastically decreases the amount of required training data.
\item[(3)]
Neural networks that perform the unsupervised learning task of optimizing axiom satisfaction discover voting rules that are substantially different from existing ones and are comparable---and often better---in axiom satisfaction. 
\end{enumerate}
Thus, we fruitfully combine two approaches to studying the space of voting rules: Drawing on machine learning, we use neural networks qua universal function approximators to \emph{explore} that space; and drawing on voting theory, we \emph{evaluate} points in that space---i.e., voting rules---by their axiom satisfaction, thus guiding the exploration.

\section{Related Work}
\label{sec: related work}

We identify three main streams of relevant literature. 

\subsection{Axiomatic Evaluation of Voting Rules} 

Social choice theory has extensively quantified the axiom satisfaction of various voting rules, with a significant focus on the concept of \emph{manipulability}, i.e., the propensity of voters to be untruthful in order to sway the outcome in their favor \citep{favardin2002borda,favardin2006some,nitzan1985vulnerability}. Numerous studies \citep{fishburn1982majority,merrill1984comparison,nurmi1988discrepancies} examine how often voting rules elect the \emph{Condorcet winner} (i.e., the alternative representing a majoritarian consensus) for relatively small elections all having the same probability of materializing (i.e., assuming the Impartial Culture distribution).
In line with our findings, the Borda rule is found to elect the Condorcet winner more often than the Plurality rule \citep{nurmi1988discrepancies}.
When considering the axiom of independence---the main trigger of Arrow's impossibility theorem---the Borda rule fulfills it more frequently than Copeland, which in turn satisfies it more than Plurality \citep{dougherty2020probability}. 
For the special case of 3 voters and 3 alternatives, an anonymous voting rule satisfies independence between 1.3$\%$ and 25.5$\%$ of the time \citep{powers2007number}.

Overall, our work aligns with the traditional concept of evaluating voting rules based on axioms. However, we also consider learning voting rules and not just evaluating them.

\subsection{Neural Networks and Voting} 

 The synergy between voting and machine learning has recently garnered more and more attention. \cite{kujawska2020predicting} use, among others, multi-layer perceptrons (MLPs) on elections of 20 alternatives and 25 voters to predict the winners of different voting rules. 
The study's primary aim is to identify an effective computational technique on top of the classical ones of the voting literature. The authors find  that the Borda rule is predicted by the neural networks with high accuracy (up to 99$\%$), but more complex rules are predicted with lower accuracy (up to 85$\%$ for Kemeny and $89\%$ for Dodgson).
\citet{burka2022voting} employ MLPs to investigate the relation between sample size and accuracy when learning different voting rules, including Plurality, Borda, and Copeland. In that work, up to 3000 data points are used based on the Impartial Culture assumption, with at most 5 alternatives and 11 voters. The MLP is found to mimic more closely Borda, no matter on which rule it is trained: e.g., for 3 alternatives and 7 voters, trained on Plurality, the MLP mimics Borda with 95$\%$ accuracy and Plurality with 86$\%$ accuracy. However, the size of the training data exhibits an impact on the results: e.g., when trained on elections with a Condorcet winner, the MLP mimics more closely Borda in sample-size up to 1000, and Copeland in larger samples. Increasing the size of the MLP by adding layers does not seem important. \cite{anil2021learning} study more complex neural network architectures (such as Set Transformers and DeepSets),  improving the accuracy of MLPs by up to $4\%$ in learning Plurality and Copeland. With sufficiently many data points, those networks are shown to match almost perfectly each voting rule, and to  generalize to elections with unseen numbers of voters. 

Similarly to all these works, our first experiment considers precisely the problem of using neural networks to learn existing rules from voting theory. However, we systematically study this with axioms: instead of only targeting the right outcomes, we test whether they are obtained via the right principles.
In an initial exploration towards the same direction, \citet{armstrong2019machine} use a single axiom---prescribing the election of a Condorcet winner when one exists---to train normatively appealing neural networks. This work relies on real data from Canadian federal elections, while ours builds on extensive synthetic data. Additionally, our third experiment illustrates original interactions between sets of different axioms that have not been explored in the literature yet. 

After observing a theoretical trade-off between fairness (in particular anonymity variations, demanding that different types of voters be treated equally) and certain notions of economic efficiency (some related to the Condorcet winner), \citet{mohsin2022learning} train two machine learning models on synthetic data and discover new voting rules that compete well against both Plurality and Borda. Although rule synthesis and axiomatic analysis is not a main focus of that work, the obtained results enforce the idea that machine learning methods can beat existing ones from economic theory when optimized for principled learning. In the slightly different framework of probabilistic voting,  MLPs are used to
learn voting rules that output distributions over the set of alternatives (rather than the certain winning alternatives), for elections of up to 7 alternatives and 29 voters.\footnote{Matone et al., 2024. DeepVoting: Learning Voting Rules with Tailored
Embeddings. arXiv:2408.13630.} It is shown that the discovered rules can lead to novel ones with improved axiomatic properties, after the appropriate embedding of the input and some adjustments of the output. Another work  studies the setting of participatory budgeting (PB), where citizens vote about the projects that they would like to implement in their neighbourhoods,  while taking into account their respective costs.\footnote{Fairstein, Vilenchik, and Gal, 2024. Learning Aggregation Rules in Participatory Budgeting: A Data-Driven Approach. arXiv:2412.01864.} Set Transformer neural networks trained
on PB instances with both real and synthetic data are found to both learn existing voting rules and to discover new ones that satisfy societal objectives (such as fair representation). Overall, our paper adds to this literature by offering a concrete framework to study combinations of basic axioms from the ML perspective,  in the most standard voting setting.

Other promising lines of research target learning an abstract voting rule given examples about its choices \citep{procaccia2009learnability} and designing a voting rule that maximizes some notion of social welfare \citep{anil2021learning}. \citet{holliday2024learning} explore the strategic manipulation of voting rules by MLPs of different sizes, generating elections of up to 6 alternatives and 21 voters. They find that sufficiently large MLPs learn to profitably manipulate all examined voting rules only with information about the pairwise majority victories between
alternatives. But some rules like Split Cycle seem more resistant than other rules (e.g., Plurality and Borda). Moreover, a systematic study of the connections between social choice and RLHF is conducted by \citet{dai2024mapping}, including fundamental axiomatic notions such as the Condorcet winner.

A different approach, rather orthogonal to ours, is to consider AI models as the individuals who vote, instead of using them as the aggregation mechanisms. In this vein, \citet{yang2024llm} 
consider a human voting experiment with 180 participants to
establish a baseline for human preferences and conducted a
corresponding experiment with LLM (e.g., GPT-4) agents. 
The voting behavior of the networks seems to be affected by the presentation
order of the alternatives, as well as the numerical ID assigned to each LLM representing a voter.
Some voting rules such as Borda show that LLMs may lead to less
diverse collective outcomes. Importantly GPT-4 seems
to over-rely on stereotypical demographics of the voters it is supposed to mimic.
Similarly, using data from Brazil's 2022 presidential election, \citet{gudino2024llms} tests the accuracy with which LLMs predict an individual's vote. They find that LLMs are more accurate than a naive rule guessing that individuals simply vote for the proposals of the candidate most aligned with their political orientation.

\subsection{Social Choice for AI Alignment} 

A growing research area studies how social choice theory can be used to guide the alignment of modern AI methods with human values and moral judgments. \citet{conitzerposition}  highlight a series of technical connections---for example, the alternatives in a voting context could be treated as all possible parameterizations of a network, or as all its possible answers.  As an indication, in a popular work about a controversial topic,  \citet{noothigattu2018voting} use data from the online `moral machine experiment' to build a model of aggregated moral preferences aimed at guiding the decision making of autonomous vehicles. On a more theoretical level,  Arrow's theorem can be utilized to prove that there does not exist any AI system that can treat all its users and human supervisors equally.\footnote{Mishra, 2023. AI Alignment and Social Choice: Fundamental Limitations and Policy Implications. arXiv:2310.16048.} Based on an investment game where participants are asked to manage wealth, \citet{koster2022human} find that by optimizing for human preferences, a reinforcement learning model can propose a wealth redistribution mechanism that is supported by the majority of participants.
We do not directly engage with the ethical dimension of this research area; still, we participate in the related foundational discussion by studying whether neural networks can learn to vote with principles.

\section{Preliminaries on Voting Theory} 
\label{sec: preliminaries}

We work in the standard setting of voting theory, where a finite set $N$ of \emph{voters} have preferences that are linear orders (also called \emph{rankings}) over a finite set $A$ of \emph{alternatives} \citep{zwicker2016introduction}. Set $m := |A|$ and $n := |N|$. We denote by $\boldsymbol{P} = (P_1,...,P_n)$ a preference \emph{profile}, i.e., a vector with the preference $P_i$ for every voter $i\in N$. This is illustrated in Figure~\ref{fig:profile}.

\begin{figure}
\centering
\begin{footnotesize}
\begin{tabular}{cccccc}
\toprule
  1  	&  2  	 &  3  	  &  4     &  5     &  6  \\
\midrule
  $c$   &  $d$   &  $d$   &  $c$   &  $a$   &  $d$   \\
  $a$   &  $b$   &  $b$   &  $b$   &  $b$   &  $a$   \\
  $b$   &  $c$   &  $a$   &  $d$   &  $c$   &  $b$   \\
  $d$   &  $a$   &  $c$   &  $a$   &  $d$   &  $c$   \\
\bottomrule
\end{tabular}
\end{footnotesize}
\Description{An example of a profile.}
\caption{A voting profile, with voters $N=\{1,\ldots,6\}$ and alternatives $A=\{a,b,c,d\}$. Each column depicts the preference of the individual voter; e.g., voter~1 prefers alternative $c$ most, followed by alternative $a$, etc.}
\label{fig:profile}
\end{figure}

For a permutation of the alternatives $\sigma: A\to A$, the ranking~$\sigma(P)$ is obtained by applying $\sigma$ elementwise to the ranking $P$, and $\sigma(\boldsymbol{P}) = (\sigma(P_1),\ldots ,\sigma(P_n))$. For a permutation of the voters $\pi: N\to N$, we define  $\pi(\boldsymbol{P}) = (P_{\pi(1)},...,P_{\pi(n)})$. A \emph{voting rule} is a function~$F$ that determines the winning alternatives for each such profile. Formally, $F: \boldsymbol{P} \mapsto S$, where $\emptyset \neq S \subseteq A$.\footnote{We use the Python package \href{https://pref-voting.readthedocs.io}{pref-voting} in all our experiments \citep{holliday2025pref_voting}.}

\subsection{Voting Rules} 

Voting rules usually fit into one of two categories: scoring rules and tournament solutions. \emph{Scoring rules} assign a score to each alternative depending on its position in the linear preference of each voter and declare as winners those alternatives with the highest score across all voters. The two primary scoring rules are \emph{Plurality} (assigning score~1 to an alternative each time it is ranked first by a voter, and score~0 otherwise) and \emph{Borda} (assigning score~$m-1$ to an alternative ranked first by a voter, score~$m-2$ to an alternative ranked second, and so on, until score~0 is assigned to an alternative ranked last by a voter).\footnote{Plurality and Borda are often contrasted in voting  \citep{hatzivelkos2018borda,terzopoulou2023}.} Another, less popular scoring rule, is \emph{Anti-Plurality}, which assigns score~0 to an alternative each time it is ranked last by a voter, and score~1 otherwise.

\emph{Tournament solutions} on the other hand are based on tournaments that capture pairwise comparisons between the alternatives, induced by the voters' preferences. For $x,y\in A$, let $N^{\boldsymbol{P}}_{x\succ y}$ be the set of voters~$i$ in the profile~$\boldsymbol{P}$ that consider $x$ better than $y$ in $P_i$, and $n^{\boldsymbol{P}}_{x\succ y} := |N^{\boldsymbol{P}}_{x\succ y}|$.  A classical tournament solution is the \emph{Copeland} rule, which selects as winners the alternatives that beat the most other alternatives in a pairwise majority contest:
$ \mathrm{argmax}_{x\in A} |\{y\in A : n^{\boldsymbol{P}}_{x\succ y} \geq n^{\boldsymbol{P}}_{y\succ x}\}|  $. In other words, an alternative can be thought to be assigned a Copeland score of 1 for every other alternative to which it is majority preferred, with the winner being the alternative with the highest score overall. Analogously, the \emph{Llull} rule assigns  to an alternative a score 1 for every other alternative to which it is majority preferred, and score $1/2$ for every other alternative to which it is majority tied---if there are no majority ties, then the Llull winners coincide with the Copeland winners.  The \emph{Top Cycle} rule selects the smallest set of alternatives such that each alternative in the set is majority preferred to each alternative outside the set. In a sense, this set captures the notion of a Condorcet winner when a single such alternative does not exist. For the \emph{Banks} rule, we say that a chain $(x,y,z,\ldots)$ in a tournament is a subset of alternatives that are linearly connected by the majority relation (i.e., $x$ is majority preferred to $y$, $y$ is majority preferred to $z$, and so on). Then $x$ is a Banks winner if it is the maximum element of a maximal chain. To define the recently proposed \emph{Stable Voting} rule \citep{holliday2023stable}, we first need to describe---the more computationally expensive, and thus left aside in our analysis---\emph{Split Cycle} \citep{holliday2023split}.  The weighted tournament of a profile is a weighted directed graph the nodes of which are alternatives with an edge from $x$ to $y$
of weight $n^{\boldsymbol{P}}_{x\succ y}$. Suppose that in each cycle of
the graph, we simultaneously delete the edges with minimal weight.
Then the alternatives with no incoming edges are the winners of Split Cycle. If there is only one Split Cycle winner in a profile $\boldsymbol{P}$, then this also is the winner of Stable Voting; otherwise $x$ is a winner of Stable Voting if for some alternative~$y$ it holds that $x$ is a Split Cycle winner with the maximal margin  $n^{\boldsymbol{P}}_{x\succ y}$ such that $x$ is a
Stable Voting winner in the profile $\boldsymbol{P}_{-y}$ obtained from $\boldsymbol{P}$ after deleting alternative~$y$.

Other prominent rules that do not fit into the two above categories are \emph{Blacks}, \emph{Baldwin}, and \emph{Weak Nanson}. Black returns the Condorcet winner (i.e., the alternative beating every other alternative in a pairwise strict majority contest) if one exists, otherwise it returns the Borda winners. Weak Nanson (resp., Baldwin) is defined iteratively on voting profiles of various sizes. In each round, all alternatives with below-average (resp., the lowest) Borda score are removed. Whenever all alternatives have the same Borda score, they all win; otherwise the alternative that remains in the last round wins. More voting rules can be defined in iterative terms, based on the idea that less popular alternatives in one round be dropped from all preferences in the next round, until some surviving alternative achieves majority support: \emph{Plurality with Runoff} consists of at most two rounds---if no alternative is ranked on top by a majority of voters in the first round, then the second round selects the majority winner between  the two alternatives that achieved the highest Plurality score in the first round; \emph{Instant Runoff} (resp., \emph{Coombs}) is such that in each round the alternatives with the lowest Plurality score (resp., Anti-Plurality score) are eliminated.

A rule different in spirit, the \emph{Uncovered Set}, relies on the idea of undefeated alternatives. Let us say that alternative $x$ left-covers $y$ if all alternatives $z$ that are majority preferred to $x$ are also majority preferred to $y$. Then, $x$ defeats $y$ if $x$ is majority preferred to $y$ and also left-covers $y$. The winners of Uncovered Set are the undefeated alternatives. Finally, the \emph{Kemeny-Young} rule is based on a notion of distance between the preferences of the voters: it constructs the linear order that minimises the sum of Kendal Tau distances for all voters' preferences and elects as winner the alternative on the top of that linear order. 

Note that all the aforementioned rules are included in our last experiment, which aims at comparing a wide pool of different voting rules that vary in nature. Our other two experiments are focused on the three most standard voting rules, Plurality, Borda, and Copeland, which are the most frequently studied in the literature to date at the intersection of voting and machine learning.
For more detailed discussions of the rules, we refer to the introductory chapter of \cite{zwicker2016introduction} and to the \texttt{pref-voting} documentation.

\subsection{Axioms}

We define axioms as functions that map a voting rule and a preference profile to a value in $\{-1,1,0\}$, where $0$ means that the axiom is not applicable, $-1$ means that the axiom is violated, and $1$ that it is satisfied.
The \emph{satisfaction degree} of a rule with respect to a given axiom is the ratio of the number of sampled profiles in which the axiom is satisfied to the number of sampled profiles in which it is applicable. We focus on axioms that capture basic and diverse normative properties of a voting rule~$F$.

\begin{itemize}
\item 
\emph{Anonymity} is always applicable; it is satisfied in~$\boldsymbol{P}$ if for all permutations of voters $\pi: N\to N$, $F(\pi(\boldsymbol{P})) = F(\boldsymbol{P})$. In words, the winners should be invariant under permutations of the voters.
\item 
\emph{Neutrality} is always applicable; it is satisfied in~$\boldsymbol{P}$ if for all permutations of alternatives $\sigma: A \to A$,  $F(\sigma(\boldsymbol{P})) = \sigma (F(\boldsymbol{P}))$. In words, under permutations of the alternatives, the winners  should be permuted respectively.
\item 
\emph{Condorcet principle} is applicable in~$\boldsymbol{P}$  if some $x\in A$ is such that $n^{\boldsymbol{P}}_{x\succ y} > n/2$ for all $y\in A \setminus \{x\}$; it is satisfied  if $F(\boldsymbol{P})=\{x\}$. In words, if a Condorcet winner exists, then it should be the unique winner of the voting rule.
\item 
\emph{Pareto principle} is applicable in~$\boldsymbol{P}$  if there exist two alternatives $x,y\in A$  such that $n^{\boldsymbol{P}}_{x\succ y} =n$; it is satisfied if $y \notin F(\boldsymbol{P})$. In words, if an alternative is considered inferior to a certain other alternative by all voters, then it should not win.
\item 
\emph{Independence} is applicable in~$\boldsymbol{P}$  if $F(\boldsymbol{P}) \neq A$; it is satisfied if for all $x\in F(\boldsymbol{P})$,  $y\notin F(\boldsymbol{P})$, and $\boldsymbol{P'}$ such that $N^{\boldsymbol{P}}_{x\succ y} = N^{\boldsymbol{P'}}_{x\succ y}$, it holds that $y \notin F(\boldsymbol{P'})$.    In words, if the relative ranking between a winning alternative and a losing alternative remains the same for all voters, then the losing alternative should not win.
\end{itemize}

All voting rules defined above satisfy anonymity and neutrality, as well as the Pareto principle, for all preference profiles. They all violate independence for some preference profile. Several of them such as Copeland, Llull, Blacks, Banks, Stable voting, and Weak Nanson  always satisfy the Condorcet principle.

\subsection{Distributions of Preference Profiles}

Specifying the distribution of preference data is essential to studying the voting behavior of a society. Indeed, there is increasing interest within the computational social choice community towards a line of work called `map of elections' that attempts to systematize simulation experiments relying on synthetic voting data \citep{boehmer2021putting, boehmer2023properties}. In this vein, we aim to ensure that our results are independent of the specific choice of the distribution. We thus employ four representative distributions that aim to cover elections of different nature, as explained below. In Section~\ref{ssec: voting theoretic parameters} we specify the choice of parameters of these distributions for our experiments, which allows them to capture a wide range of realistic scenarios. It is also important to note that all our experimental results are found to be robust across distributions; thus we do not expect that by examining additional distributions we would discover significantly different insights.

\emph{Impartial Culture (IC)} assumes that all preference profiles have the same probability of appearing. Each preference of a voter in a profile is sampled uniformly at random.
The \emph{Mallows} distribution \citep{mallows1957non} fixes a reference ranking $P$  and assumes that each voter's preference is close to that ranking. Closeness to the reference ranking is defined using the Kendall Tau distance, parameterized by a dispersion parameter $\phi \in (0,1]$.\footnote{\label{ftn: definition Kendall Tau distance} The \emph{Kendall Tau distance} between two rankings $P$ and $Q$ over the same set of alternatives is the number of pairs of alternatives $(a, b)$ such that $a$ is preferred over $b$ in $P$ but not so in $Q$.} This distribution reduces to IC when $\phi =1$ and concentrates all mass on $P$ as $\phi$ tends to 0.  

The IC and Mallows distributions are complementary: IC is simplistic and widely employed in theoretical works on voting rules as discussed earlier in the literature review; it captures an extreme case with no correlation between preferences of voters. Mallows is often employed in numerical studies of voting rules that use artificial data but wish to capture more realistic voting scenarios \citep{caragiannis2017learning,lee2014crowdsourcing}. 

The next two distributions also capture more intricate relationships between the preferences in a profile. According to the \emph{2D-Euclidean} distribution, voters and alternatives are distributed randomly in 2-dimensional Euclidean space, and the closer an alternative is to a voter the more the voter prefers that alternative.
Finally, the \emph{Urn} distribution \citep{eggenberger1923statistik}  generates a profile given a parameter $\alpha \in [0, \infty)$. Voters randomly draw their ranking from an urn. Initially, the urn includes all possible rankings over the alternatives. After a voter randomly draws  from the urn, we add  to the urn $\alpha n!$ copies of that ranking. When $\alpha=0$, this reduces to IC.

\section{Method}
\label{sec: method}

To answer our research questions, we develop the \emph{axiomatic deep voting} framework, visualized in Figure~\ref{fig: deep voting architecture}. It is built around a neural network, which is a function $f_w : \mathbb{R}^i \to \mathbb{R}^j$ parametrized by weights $w \in \mathbb{R}^k$. We will instantiate this with three different neural network architectures (see Section~\ref{ssec: architectures}). Every profile $\boldsymbol{P}$ is mapped, via an \emph{encoding} function~$e$ (see Section~\ref{ssec: encoding}), to a vector $x = e(\boldsymbol{P}) \in \mathbb{R}^i$, for which the neural network produces an output $\hat{y} \in \mathbb{R}^j$.\footnote{For our third architecture, the encoding function is part of the neural network, while for the first two it is independent (see Section~\ref{ssec: encoding}); hence we treat $e$ as a separate entity here.} The \emph{decoding} function~$d$ (see Section~\ref{ssec: decoding}) turns this output into a winning set $S = d(\hat{y})$.
Thus, this setup realizes the voting rule: 
\begin{align*}
F_w \big( \boldsymbol{P} \big) 
:=
d \Big( f_w \big( e(\boldsymbol{P} ) \big) \Big).
\end{align*}
The network is trained, as usual, using backpropagation with respect to a \emph{loss function} (in Section~ \ref{ssec: loss functions}), which relies on training data.
Finally, we \emph{evaluate} (in Section~\ref{ssec: evaluation}) the trained network not only with respect to its accuracy (how well it fits the test dataset), but, crucially, also by how much it satisfies the various voting axioms.

\begin{figure*}
\centering
\tikzset{>=latex}
\tikzstyle{box} = [draw, rectangle, text=black, minimum height=3em, minimum width=3em, font=\footnotesize, text opacity = 1]
\tikzstyle{inference} = [draw=black, fill=green, fill opacity=0.2]
\tikzstyle{training} = [draw=black, fill=violet, fill opacity=0.2]
\tikzstyle{evaluation} = [draw=black, fill=yellow, fill opacity=0.2]
\tikzstyle{input} = [draw=black, fill=cyan, fill opacity=0.2]
\tikzstyle{output} = [draw=black, fill=orange, fill opacity=0.2]
\tikzstyle{boxlabel} = [font=\itshape\footnotesize , inner sep=0pt]
\begin{tikzpicture}[scale=.9]
\draw[inference] (0,0.1) rectangle ++(8,1.8);
\node[boxlabel] at (-.3,1) (infbox) {\rotatebox{90}{inference}};
\node[box, input] at (-2,1) (prof) {profile};
\node[box] at (1,1) (enc) {encoding};
\draw[->] (prof)-- (infbox) -- (enc);
\node[box] at (4,1) (nn) {neural network};
\draw[->] (enc)--(nn);
\node[box] at (7,1) (dec) {decoding};
\draw[->] (nn)--(dec);
\node[box, output] at (9.5,1) (ws) {winning set};
\draw[->] (dec)--(ws);

\draw[training] (0,2.1) rectangle ++(8,1.8);
\node[boxlabel] at (-.3,3) (trainbox) {\rotatebox{90}{training}};
\node[box] at (4,3) (loss) {loss function};
\node[box, input] at (-2,3) (data) {training data};
\draw[->] (data.east) -- (trainbox) -- (.9,3) -- ([shift={(-.1,0)}]enc.north);
\draw[->] ([shift={(.1,0)}]enc.north) -- (1.1,3) -- (loss.west);
\draw [->] ([shift={(-.5,0)}]nn.north) -- ([shift={(-.5,0)}]loss.south);
\draw [<-] ([shift={(.5,0)}]nn.north) -- ([shift={(.5,0)}]loss.south);

\draw[evaluation] (0,-1.9) rectangle ++(8,1.8);
\node[boxlabel] at (-.3,-1) (evalbox) {\rotatebox{90}{evaluation}};
\node[box, input] at (-2,-1) (test) {test data};
\node[box] at (4,-1) (eval) {\parbox{3cm}{\centering accuracy \& \\ axiom satisfaction}};

\draw [-] ([shift={(0,.1)}]test.east) -- ([shift={(0,.1)}]evalbox.west);
\draw [->] ([shift={(0,.1)}]evalbox.east) -- (1,-.9) -- (enc.south);
\draw [-] ([shift={(0,-.1)}]test.east)  -- ([shift={(0,-.1)}]evalbox.west);
\draw [->] ([shift={(0,-.1)}]evalbox.east) -- ([shift={(0,-.1)}]eval.west);
\draw [->] (dec.south) -- (7,-1) -- (eval.east);

\end{tikzpicture}
\Description{The axiomatic deep voting architecture in diagrammatic form.}
\caption{The axiomatic deep voting architecture.}
\label{fig: deep voting architecture}
\end{figure*}

\subsection{Architectures}
\label{ssec: architectures}

We use three paradigmatic neural network architectures from modern machine learning.

First, \emph{multi-layer perceptrons} (MLPs)---also known as feed-forward neural network---are the classic deep neural net \citep[see, e.g.,][ch.~6]{Goodfellow2016}. They consist of an input layer of neurons, one or more hidden layers, and an output layer. 

Second, \emph{convolutional neural networks} (CNNs) are a standard architecture  to process grid-like input data such as images \citep[see, e.g.,][ch.~9]{Goodfellow2016}, and in our case profiles. Compared to MLPs, they additionally use so-called convolutional layers to capture local, invariant patterns in the input.

Third, we devise an architecture that satisfies the anonymity axiom by design: We view profiles as sentences whose words are the rankings. We use the \emph{word embedding} algorithm \emph{Word2vec} \citep{Mikolov2013} to map each ranking to a high-dimensional embedding vector. These vectors are averaged---hence we get anonymity---and an MLP then classifies this average into a winning set. This combined architecture we call here \emph{word embedding classifiers} (WECs).

\subsection{Encoding}
\label{ssec: encoding}

To ensure our neural networks learn general patterns, we do not work with a fixed number of voters and alternatives, but only with a maximal number of voters $n_\mathrm{max}$ and a maximal number of alternatives $m_\mathrm{max}$. So the model should allow as input any profile $\boldsymbol{P}$ over the set of voters $N = \{ 0 , \ldots , n-1 \}$ with $n \leq n_\mathrm{max}$ and set of alternatives $M = \{ 0 , \ldots , m-1 \}$ with $m \leq m_\mathrm{max}$. (For readability, we also write $a,b,c, \ldots$ for the alternatives.)
We write  $a_r^s$ for the $r$-th most preferred alternative of voter $s$, so the profile $\boldsymbol{P}$ is represented as the matrix $(a_r^s)_{r,s}$, whose columns are the rankings as in Figure~\ref{fig:profile}. 
We write $\tilde{\boldsymbol{P}} = (\tilde{a}_r^s)_{r,s}$ for the result of padding the $m \times n$ matrix $\boldsymbol{P}$ with the symbol $\sim$ to the maximal input dimensions $m_\mathrm{max} \times n_\mathrm{max}$. (So $\tilde{a}_r^s$ is $a_r^s$ if $r \leq m$ and $s \leq n$, and otherwise it is $\sim$.)

How should we encode $\tilde{\boldsymbol{P}}$ so it can be inputted to a neural network? The most straightforward way is to read each alternative $a_r^s \in M$ as the number that it is and the padding symbol $\sim$ as, say, $-1$. Then the matrix $\tilde{\boldsymbol{P}}$ is regarded as a vector of dimension $m_\mathrm{max} n_\mathrm{max}$.
However, this does not perform well, so, following \citet{anil2021learning}, we represent an alternative not as a number but as a one-hot vector. For $a \in \{ 0 , \ldots , m_\mathrm{max} - 1 \}$, let $\overline{a}$ be the vector of length $m_\mathrm{max}$ that is $1$ at position $a$ and $0$ everywhere else. For the padding symbol, let $\overline{\sim}$ be the vector of length $m_\mathrm{max}$ that is $0$ everywhere.
We write $\overline{\boldsymbol{P}} = (\overline{\tilde{a}_r^s })_{r,s}$.

The \emph{encoding function for MLPs}, $e_\mathrm{MLP}$, maps profile $\boldsymbol{P}$ to the vector $x$ obtained by casting the matrix $\overline{\boldsymbol{P}}$ column by column into a flattened vector (of dimension $m_\mathrm{max}^2 n_\mathrm{max}$). This vector $x$ can then be inputted into the MLP.

The \emph{encoding function for CNNs} regards the matrix $\overline{\boldsymbol{P}}$ as a pixel image: the `pixel' at position $(r,s)$ has the `color value' $\overline{\tilde{a}_r^s}$.
Thus, $e_\mathrm{CNN}$ maps profile $\boldsymbol{P}$ to the matrix $\overline{\boldsymbol{P}}$ recast as a tensor with dimensions 
$(\mathit{channel}, \mathit{height}, \mathit{width}) = (m_\mathrm{max}, m_\mathrm{max}, n_\mathrm{max})$. This tensor can then be inputted into the CNN.

The \emph{encoding function for WECs} regards the profile $\boldsymbol{P} = (P_1 , \ldots, P_n)$ as a sentence with words $P_i$. We train it to embed these words into vectors of a fixed high dimension. Thus, unlike the previous encoding functions, this one is not separate from the neural network but rather forms the first layer of the WEC, with the remaining layers processing the embedding vectors.  
More precisely, we first pre-train the embeddings as follows.
For a given corpus size $c$, we sample $c$-many profiles from a given distribution of profiles (e.g., IC) to form our corpus (i.e., a set of sentences). The rankings occurring in the profiles form the vocabulary of this corpus, to which we add the \texttt{unk} token (to later represent \emph{unknown} rankings, i.e., rankings that are not in the vocabulary) and the \texttt{pad} token (to \emph{pad} a profile to length $n_\mathrm{max}$). Due to the \texttt{unk} token, this encoding applies to \emph{all} profiles, even if it contains rankings that are not part of the model's vocabulary. Using Word2vec, we train embeddings which represent words in the vocabulary as vectors. When instantiating the WEC architecture, these embeddings form the first layer: it maps the profile $(P_1 , \ldots, P_n)$ to the corresponding embedding vectors $(v_1 , \ldots , v_n)$. The next layer averages these vectors into a single vector $v$, followed by several linear layers ending with the output layer.

\subsection{Decoding}
\label{ssec: decoding}

Given a profile $\boldsymbol{P}$ as input, all neural network architectures produce as output the logits $\hat{y} = (\hat{y}_0 , \ldots , \hat{y}_{m_\mathrm{max}})$ in $\mathbb{R}^{m_\mathrm{max}}$. We apply the sigmoid function $\sigmoid$ elementwise to obtain the probability that alternative $r$ is in the winning set.\footnote{
We use `$\sigmoid$' instead of `$\sigma$' to denote the sigmoid function, in order to not confuse it with the previous use of `$\sigma$' for permutations of alternatives.}
With $m$ the number of alternatives in profile $\boldsymbol{P}$, we define the decoding function
 \begin{align*}
d_m ( \hat{y} ) 
&:=
\big\{ 
r \in \{ 0 , \ldots , m \}
:
\sigmoid ( \hat{y}_r ) > 0.5
\big\}.\footnotemark
\end{align*}%
\footnotetext{A priori, it can happen that the neural network does not assign any winner, in contrast to our definition of a voting rule. We check (and train) that this happens, if at all, only with a negligible probability.}%
In experiment~3, we will consider further versions of this decoding function (see Section~\ref{ssec: experiment 3}).

\subsection{Loss Functions}
\label{ssec: loss functions}

Since multiple alternatives can win, we cast the task of finding a voting rule as a \emph{multi-label classification} problem.
Each input profile $\boldsymbol{P}$ is associated with $m$ binary labels (where $m$ is the number of alternatives in $\boldsymbol{P}$), and the $r$-th label is $1$ if and only if the $r$-th alternative is in the winning set associated with $\boldsymbol{P}$. Hence we use \emph{binary cross entropy} as loss function.

A main contribution of this paper is that, for each axiom, we also design a loss function that enforces satisfaction of that axiom. So, for each axiom $\mathsf{ax}$, we define a function $L_\mathsf{ax} (f_w, \boldsymbol{P})$ that takes as input the function $f_w$ computed by the neural network with weights $w$ and a profile $\boldsymbol{P}$. It outputs a non-negative real number describing numerically how much the axiom is satisfied: $0$ means perfect axiom satisfaction, while higher numbers mean worse axiom satisfaction. 
We now define these loss functions, which we will use later.

\emph{Anonymity}.
Given the network $f_w$ and profile $\boldsymbol{P}$, uniformly sample $N$-many permutations $\pi_1 , \ldots , \pi_N$ of the set of voters of $\boldsymbol{P}$ and define   
\begin{align*}
L_A ( f_w , \boldsymbol{P} ) 
:=
\frac{1}{N} \sum_{r = 1}^N 
\mathrm{KL} \Big( 
f_w \big( e ( \boldsymbol{P} ) \big) , 
f_w \big( e ( \pi_r ( \boldsymbol{P} ) ) \big)
\Big),
\end{align*}
where $\mathrm{KL}$ is Kullback--Leibler divergence.\footnote{Though, in principle, other distance/similarity functions can be considered.}

\emph{Condorcet}.
If $\boldsymbol{P}$ has no Condorcet winner, $L_C(f_w , \boldsymbol{P}) := 0$, and otherwise, if that Condorcet winner is alternative $a$, define (recall $\overline{a}$ is the one-hot vector for alternative $a$)
\begin{align*}
L_C(f_w , \boldsymbol{P}) 
:=
\mathrm{KL} \big(
f_w ( e( \boldsymbol{P} ) ) , \overline{a}
\big).
\end{align*} 

\emph{Pareto}.
We define (recall that $n_{a \succ b}^{\boldsymbol{P}} = n$ means that all voters in $\boldsymbol{P}$ rank $a$ above $b$)
\begin{align*}
L_P(f_w , \boldsymbol{P}) 
:=
\sum_{a,b \text{ with } n_{a \succ b}^{\boldsymbol{P}} = n}
\sigmoid \Big( 
f_w ( e( \boldsymbol{P} ) )_b
\Big).
\end{align*}

\emph{Independence}.
Define $L_I(f_w , \boldsymbol{P}) := 0$ if $\boldsymbol{P}$ does not have at least two alternatives. Otherwise, randomly sample $N$-many pairs $(a_r , b_r)$ of distinct alternatives in $\boldsymbol{P}$ and randomly sample, for each ranking $P_k$ of $\boldsymbol{P} = (P_1 , \ldots , P_n)$, a shuffling $P'_k$ of $P_k$ in which, however, the order of $a_r$ and $b_r$ is the same as in $P_k$, and set $\boldsymbol{P}_r := (P'_1 , \ldots , P'_n)$.
Write
$\hat{y} := f_w ( e ( \boldsymbol{P} ) )$ and
$\hat{y}^r := f_w ( e ( \boldsymbol{P}_r ) )$, 
and define
\begin{align*}
L_I(f_w , \boldsymbol{P}) 
:=
\sum_{r = 1}^N
\mathrm{KL} \Big(
\big( \hat{y}_{a_r} \hat{y}_{b_r} \big) ,
\big( \hat{y}^r_{a_r} \hat{y}^r_{b_r} \big) 
\Big).
\end{align*}

\emph{No winner}.
Recall that voting rules are required to output at least one winner. 
This is usually not called an axiom, and we did not hard-code this into our architectures. 
So we also want to optimize our neural networks to align with this requirement.
Hence we define the `no winner' loss as follows. 
Writing 
$\hat{y} = f_w ( e ( \boldsymbol{P} ) )$, 
we want that at least one of the numbers in 
$p := \big( \sigmoid ( \hat{y}_1 ), \ldots , \sigma ( \hat{y}_m ) \big)$
is above $0.5$, i.e., the maximum norm $\Vert p \Vert_\infty$ should be above $0.5$. Hence the more it is below that, the worse the loss:
\begin{align*}
L_{NW} ( f_w , \boldsymbol{P} ) 
:=
\max \big(
0.5 - 
\Vert p \Vert_\infty
,
0
\big).\footnotemark
\end{align*}
\footnotetext{To see almost-everywhere differentiability of the loss functions, use the distributivity of the differential operator over sums, the chain rule, and the almost-everywhere differentiability of the involved functions ($\mathrm{KL}$, $\sigmoid$, $\max$, $\Vert \cdot \Vert_\infty$).}

\subsection{Evaluation Metrics}
\label{ssec: evaluation}

We have two ways of evaluating the model: accuracy and axioms. 
First, we calculate the accuracy of the trained neural network on a given test set in two ways: \emph{Identity} (or \emph{hard}) \emph{accuracy} is the percentage of pairs $(\boldsymbol{P} , S)$ in the test set for which $F_w ( \boldsymbol{P}) = S$. \emph{Subset} (or \emph{soft}) \emph{accuracy} is defined in the same way but replacing the identity with $F_w ( \boldsymbol{P}) \subseteq S$.
Second, we calculate the satisfaction degrees for the various axioms of the voting rule $F_w$ that the trained neural network realizes (see Section~\ref{ssec: evaluating axiom satisfaction} for details).

\section{Experimental Setup}
\label{ssec: experimental setup}
We describe all details for designing and evaluating our experiments.

\subsection{Voting-Theoretic Parameters}
\label{ssec: voting theoretic parameters}

We work with all four profile distributions and with $n_\mathrm{max} = 77$ and $m_\mathrm{max} = 7$ in the first experiment and $n_\mathrm{max} = 55$ and $m_\mathrm{max} = 5$ in the other experiments. The first experiment does not show a qualitative difference between these settings, but the latter is computationally more efficient.

For all experiments, we use the Mallows distribution with a parameter rel-$\phi$ (randomly generated) that, together with the number of alternatives, determines the value of the dispersion parameter~$\phi$.  According to \cite{boehmer2021putting} and \cite{boehmer2023properties}, this methodology generates data that more closely resemble those of real elections.
We use the Urn-R distribution \citep{boehmer2021putting}, where, for each generated profile, $\alpha$ is chosen according to a Gamma distribution with shape parameter $k=0.8$ and scale parameter $\theta =1$.
The other distributions do not need further parameters.

\subsection{Synthetic Data Generation}
\label{ssec: datasets}

We can sample profiles in a controlled and realistic manner and produce their corresponding winning sets with existing voting rules (see Section~\ref{sec: preliminaries}). So we generate synthetic data:
Given a profile distribution $\mu$ and a voting rule $F$, we randomly pick integers $n \in [1, n_\mathrm{max}]$ and $m \in [1, m_\mathrm{max}]$ and $\mu$-sample a profile $\boldsymbol{P}$ with $n$ voters and $m$ alternatives and compute $S = F(\boldsymbol{P})$.
Thus, we generate a dataset $D = \big\{ (\boldsymbol{P}_1 , S_1) , \ldots , (\boldsymbol{P}_k , S_k) \big\}$.

\subsection{Evaluating Axiom Satisfaction}
\label{ssec: evaluating axiom satisfaction}

To evaluate the axiom satisfaction of a voting rule (be it realized by a neural network or an existing one), we sample $400$ profiles on which the axioms are applicable. We use the same profile distribution $\mu$ as was used for training the neural network, and we again randomly choose integers $n \in [1, n_\mathrm{max}]$ and $m \in [1, m_\mathrm{max}]$ before $\mu$-sampling a profile with $n$ voters and $m$ alternatives.
To compute whether an axiom is satisfied for a profile, the axioms of anonymity, neutrality, and independence require sampling of permutations. We sample, per profile, $50$, $50$, and $w (m - w) 256$ permutations, respectively (where $w$ is the number of winners according to the rule on the profile, and hence $m - w$ is the number of losers).\footnote{
Independence considers more `degrees of freedom', so we take more samples. Specifically, we go through all pairs $(x,y)$ where $x$ is a winner and $y$ a loser, and we sample, for each voter, $4$ alternative rankings that, however, have the same relative order of $x$ and $y$ as the actual ranking submitted by that voter, and then build $4^4 = 256$ profiles out of these alternative rankings and check that $y$ still does not win.}

\subsection{Hyperparameters}
\label{ssec: hyperparamters}

All models use ReLU as the activation function.
Our MLP has four hidden layers with 128 neurons each, like those of~\cite{anil2021learning}. 
The CNN has two convolution layers with kernel size $(5,1)$ and $(1,5)$, respectively (and 32 or 64 channels), followed by three linear layers with 128 neurons.
Thus, the first kernel can pick up local patterns in the rankings of the voters, while the second kernel can pick up local patterns among the $i$-th preferred alternatives of the voters. (Appendix~\ref{ssec: app CNN kernel size} establishes the optimality of this choice when compared to other kernel sizes and additional pre-processing.)
The WEC has the word embedding layer, then the averaging layer, and then three linear layers with 128 neurons. For pre-training the word embedding layer with word2vec, we use a corpus size of $10^5$, an embedding dimension of $200$, and a window size of $7$.\footnote{That is in the setting $n_\mathrm{max} = 77$ and $m_\mathrm{max} = 7$. When $n_\mathrm{max} = 55$ and $m_\mathrm{max} = 5$, we reduce this to a corpus size of $2 \times 10^4$, an embedding dimension of $100$, and a window size of $5$.}
The corpus size is chosen large enough so that no occurrences of the \texttt{unk} token are observed in $1,000$ sampled profiles.

This results in the following numbers of parameters in the setting $n_\mathrm{max} = 77$ and $m_\mathrm{max} = 7$:
$500,487$ (MLP),
$1,834,439$ (CNN), and 
$1,226,143$ (WEC).
In the setting $n_\mathrm{max} = 55$ and $m_\mathrm{max} = 5$ this reduces to:
$193,285$ (MLP),
$232,165$ (CNN), and
$45,585$ (WEC).
Thus, the models have roughly comparable capacities.
Section~\ref{sec: app_hyperparamter} in the Appendix motivates these choices via hyperparameter tuning.

For training, we use the \emph{AdamW} algorithm~\citep{Loshchilov2019}. We use a batch size of 200. Since we have synthetic data, we do not use epochs and hence only specify the number of gradient steps. In experiments~1, 2, and 3, these are 15,000, 5,000, and 15,000, respectively. 
Similar to \citet{anil2021learning}, we use as a learning rate scheduler cosine annealing with warm restarts~\citep{loshchilov2017}.  
All results are reported for one fixed seed. (In the Appendix, Table~\ref{tbl: exp1 appendix cross validation} performs cross validation and Tables~\ref{tbl: exp 3 averaged training IC} and~\ref{tbl: exp 3 averaged training Mallows} report averaged results across different seeds.) All experiments were run on a laptop without GPU.

\section{Results and Analysis} 
\label{sec: experiments}

Within our axiomatic deep voting framework, we answer our three research questions: 
(1)~Are preferences-aggregating neural networks correct for the right reasons? No. 
(2)~Can they learn voting-theoretic principles by example? No.
(3)~Can they synthesize new rules guided by the principles? Yes.

\subsection{Experiment 1: Correct for the Right Reasons?}
\label{ssec: experiment 1}

Recent work in computer science has studied the capabilities of neural networks to learn voting rules \citep{anil2021learning,burka2022voting}, but without asking whether ``the system performs well for the right reasons''~\cite[p.~5192]{Bender2020}.
Here we use voting-theoretic axioms to shed light on the learning behavior of neural networks, specifically aiming to distinguish solely accurate versus principled learning. 

\paragraph{Design.}

We train each of the three neural network architectures (MLP, CNN, and WEC) on data from each one of the three basic voting rules (Plurality, Borda, and Copeland) using four different sampling distributions (IC, Urn, Mallows, and Euclidean) with the parameters mentioned in Section~\ref{ssec: voting theoretic parameters}. 
We report the results as \emph{relative} accuracy and axiom satisfaction, i.e., 
\begin{equation*}
\langle \text{relative evaluation} \rangle
=
\langle \text{rule evaluation} \rangle - \langle \text{model evaluation} \rangle.
\end{equation*}
For example, if the model has $95\%$ accuracy, then, since the rule has $100\%$ accuracy, there is a relative accuracy \emph{loss} of $100\% - 95\% = 5\%$.
If the model has $35\%$ satisfaction of the independence axiom and the rule only $30\%$, then the relative independence satisfaction is $30\% - 35\% = -5 \%$, so there is a relative independence \emph{gain} of $5 \%$.

\paragraph{Results.}

\begin{figure*}
\centering
\includegraphics[width=0.49\linewidth]{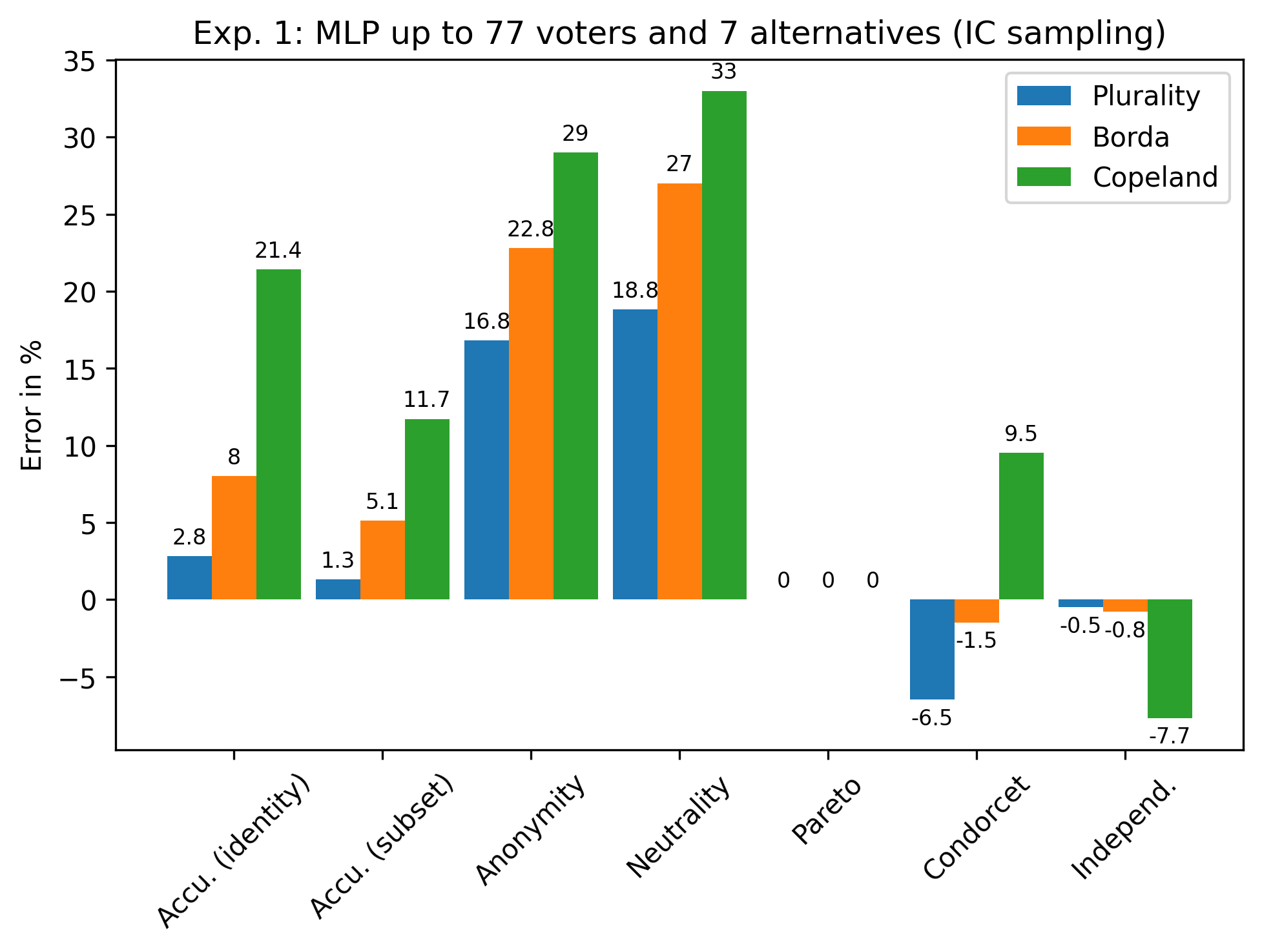}
\includegraphics[width=0.49\linewidth]{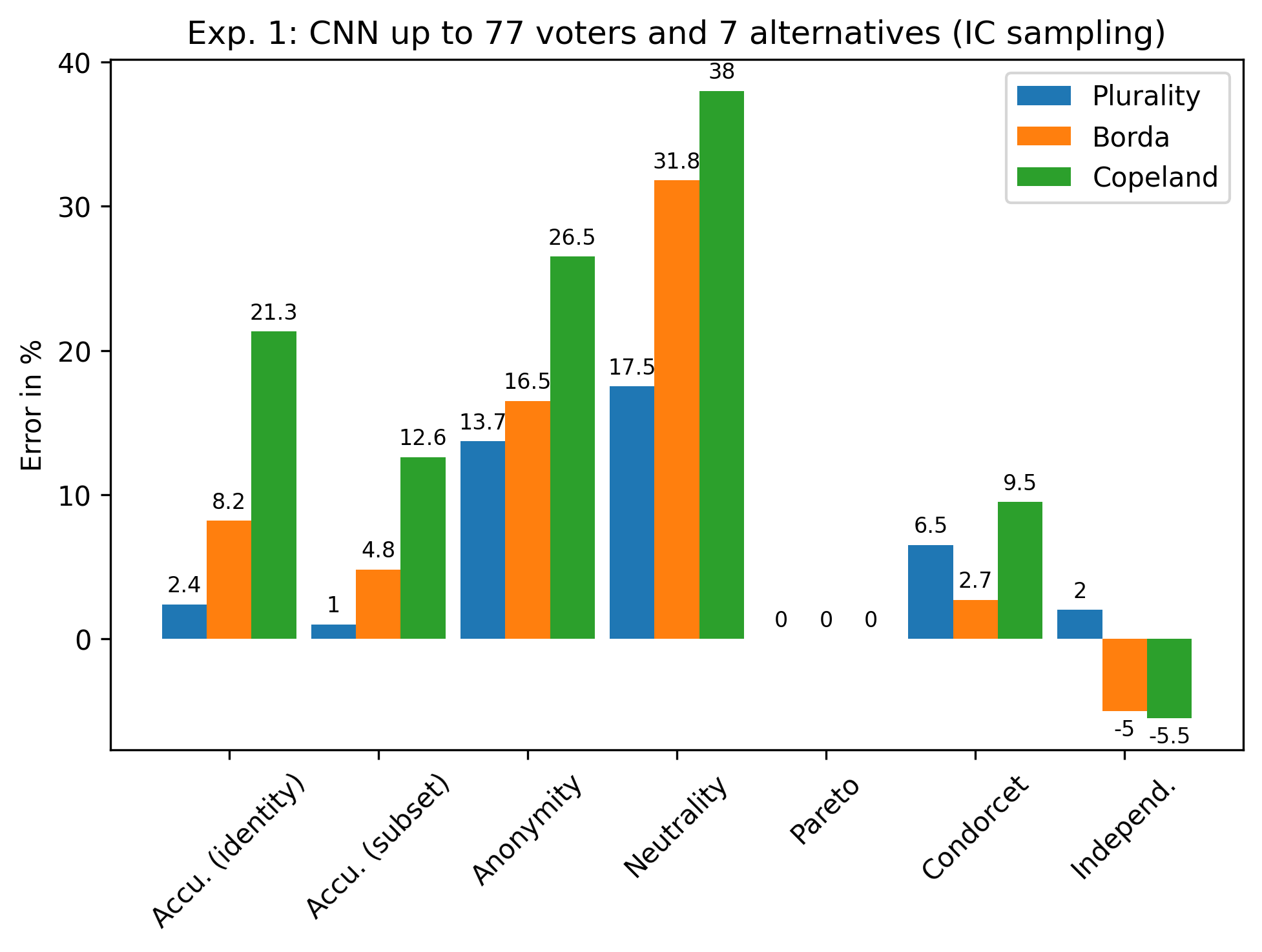}\\
\includegraphics[width=0.49\linewidth]{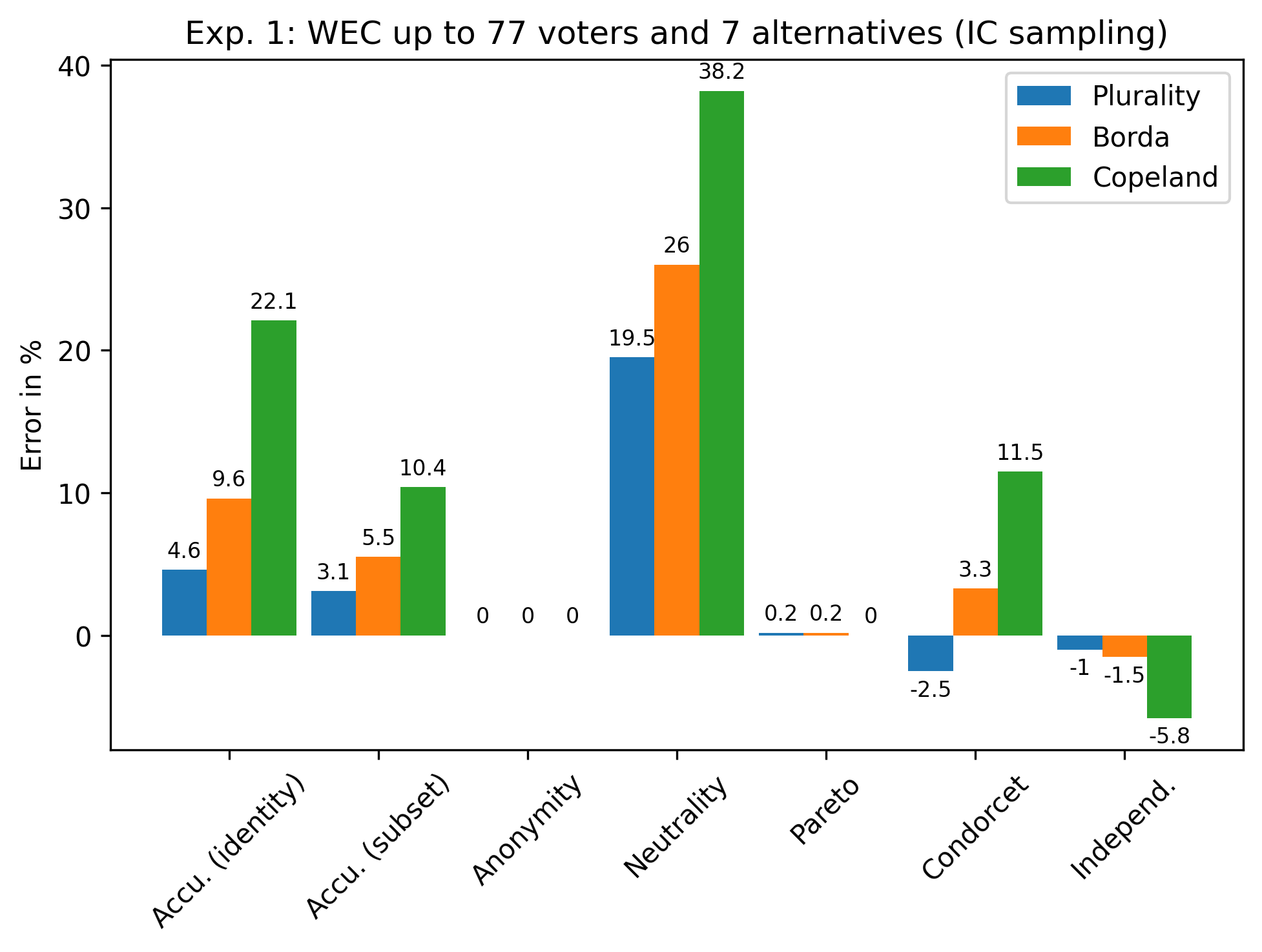}
\Description{Three barplots describing the accuracies and axiom satisfactions of the three architectures.}
\caption{Training the three architectures (MLP, CNN, and WEC) on data from Plurality, Borda, and Copeland (the three bars in each plot) with IC samples and comparing the errors in both accuracy and axiom satisfaction.
The error describes the rule's evaluation minus the model's evaluation.  
Intuitively, $2.8\%$ accuracy error means $2.8\%$ loss in accuracy: the rule by definition is correct so it has $100\%$  accuracy, but the model obtains only $97.2\%$ accuracy; similarly, $-6.5\%$ Condorcet error means $6.5\%$ gain in accuracy: the rule has $73.25\%$ Condorcet satisfaction, but the model obtains $79.75\%$ Condorcet satisfaction.
}
\label{fig: exp1}
\end{figure*}

The relative accuracy and axiom satisfaction when sampling with the IC distribution are given in Figure~\ref{fig: exp1}. (Section~\ref{sec: app_exp1} in the Appendix shows similar results for the other distributions.)
The three architectures do not differ much in accuracy. The best accuracy is achieved for the simple Plurality rule, while the complex Copeland rule decreases accuracy.

Notably, across all voting rules, architectures, and distributions, we see large losses in neutrality despite only low losses in accuracy (e.g., $4.6\%$ relative identity-accuracy loss but  $19.5\%$ relative neutrality loss for the WEC architecture when trained on the Plurality rule). Large anonymity losses are also observed under the MLP and CNN architectures (the WEC is anonymous by design). This is particularly noteworthy since anonymity and neutrality are  $100\%$ satisfied by the given voting rules.
The MLP and CNN models regularly show larger neutrality losses than anonymity losses (with the models trained on Plurality demonstrating the smallest such difference).  

Regarding the other axioms, all models adhere perfectly to Pareto, in accordance with the voting rules on which they are trained. The MLP and WEC models trained on Plurality seem to exhibit relative Condorcet gains, but Condorcet losses are found for the CNN model. Along a similar line, the MLP model trained on Borda obtains relative Condorcet gains, but this is not the case for the CNN and WEC models. Since Copeland always satisfies the Condorcet principle, there is a relative Condorcet loss for all models---yet, it is rather small.
The MLP and WEC models trained on Plurality and Borda, as well as the CNN  model trained on Plurality, satisfy independence to a similar degree as the rules do on which they are trained. All models trained on Copeland exhibit relative independence gains, and the same holds for the CNN model trained on Borda.

\paragraph{Discussion.}

Regarding the learnability of different voting rules, the simplicity of Plurality is probably the reason behind the high accuracy with which all models learn it. However, this simplicity also renders Plurality problematic in other contexts~\citep{laslier2011and}.

The models take a stance on the well-documented tension between anonymity and neutrality.\footnote{No voting rule that always elects a single winner can simultaneously be anonymous and neutral.} They tend to favor outcomes that align more closely with the former than with the latter. This inclination exposes an inherent bias within neural networks when navigating fundamental democratic axioms.

As the architectures are not invariant to permutations of the input data, the severe violations of neutrality (and of anonymity for MLPs and CNNs) are not \emph{a priori} surprising. However, these violations persist even for high accuracy with respect to rules that are perfectly neutral and anonymous.\footnote{
Actually, whether this is surprising depends on which of the following two intuitions one has. The first intuition regards these results as surprisingly bad: Given the high accuracy, we may expect that the neural network should have `gotten the idea' of the voting rule, and hence of its anonymity and neutrality, so the amount of violations is surprising. The second intuition regards the results as surprisingly good: For anonymity and, respectively, neutrality to be satisfied on a given profile, we require the neural network to output the correct answer on 50 permutations of the profile, while accuracy requires being correct only on that very profile, so it is not surprising that the neural network struggles more with anonymity and neutrality. 
But whether surprising or not, the results stay the same: for the specified percentages of considered profiles, we have violations in anonymity and neutrality, i.e., we do \emph{not} have high certainty (at least one failure in 50 checks) that the neural network outputs the desired answer under permutations. 
}

Overall, our experiment on accurate versus principled learning highlights the importance of the \emph{reasons} behind automated decision-making. Outcomes that mimic well-defined voting rules are arguably still unreliable, since they do not come with a guarantee of respecting the principles on which those rules are built.

\subsection{Experiment 2: Learning Principles by Example?}
\label{ssec: experiment 2}

Can we teach neural networks voting-theoretic principles, beyond merely presenting data from various voting rules? A natural approach to integrate expert knowledge in neural networks is data augmentation. In the voting context, this was proposed by \citet{xia2013designing} but has not been tested in practice, to the best of our knowledge. 
We focus on the anonymity and neutrality axioms, since they were violated most, and we also test the effects of data augmentation on the model's accuracy.

Asking if data augmentation helps can be understood in two ways: First, does training with augmented data increase axiom satisfaction without diminishing accuracy? Second, if so, does it do this better than just training with sampled data points? A `yes' to the first question improves data efficiency: we get at least as good a model even when only part of the data is `real' and the rest is augmented. A `yes' to the second question means that we can actually improve our model's performance in the preceding experiment.

\paragraph{Design.}
We test data augmentation in two versions. In the \emph{first version}, we form an initial dataset (i.e., pairs of a sampled profile with corresponding winning set) and we train an architecture on this dataset. Then we continue training it on augmented data points, i.e., data points obtained from the initial data points by renaming alternatives (`neutrality variations') or by renaming voters (`anonymity variations'). For comparison, we make a copy of our model after the initial training and continue training the copy with the same number, but sampled data points (rather than augmented data points). If the model trained on augmented data improves axiom satisfaction without worsening accuracy, we get a `yes' to the first question. If it also is better than the copied model, we also get a `yes' to the second question. Otherwise, any improvement only comes from a mere increase in the quantity of data points and not from their quality.

A potential issue of this version is that, during the continued training with augmented data, the model does not see any further sampled data and hence might lose in accuracy. The \emph{second version} fixes this issue by making sure that each training batch consists of $p$ percent sampled data points with the remaining data points being neutrality variations (resp., anonymity variations) of those sampled data points. 
For different choices of $p$, we then test the models' achieved axiom satisfaction and accuracy. We get a `yes' to the first (resp., second) question if axiom satisfaction and accuracy are not worse (resp., better) for lower values of $p$ when compared to $p = 100\%$ (i.e., only sampled data).

\paragraph{Results.}

\begin{figure*}
\begin{center}
\includegraphics[width=0.43\linewidth]{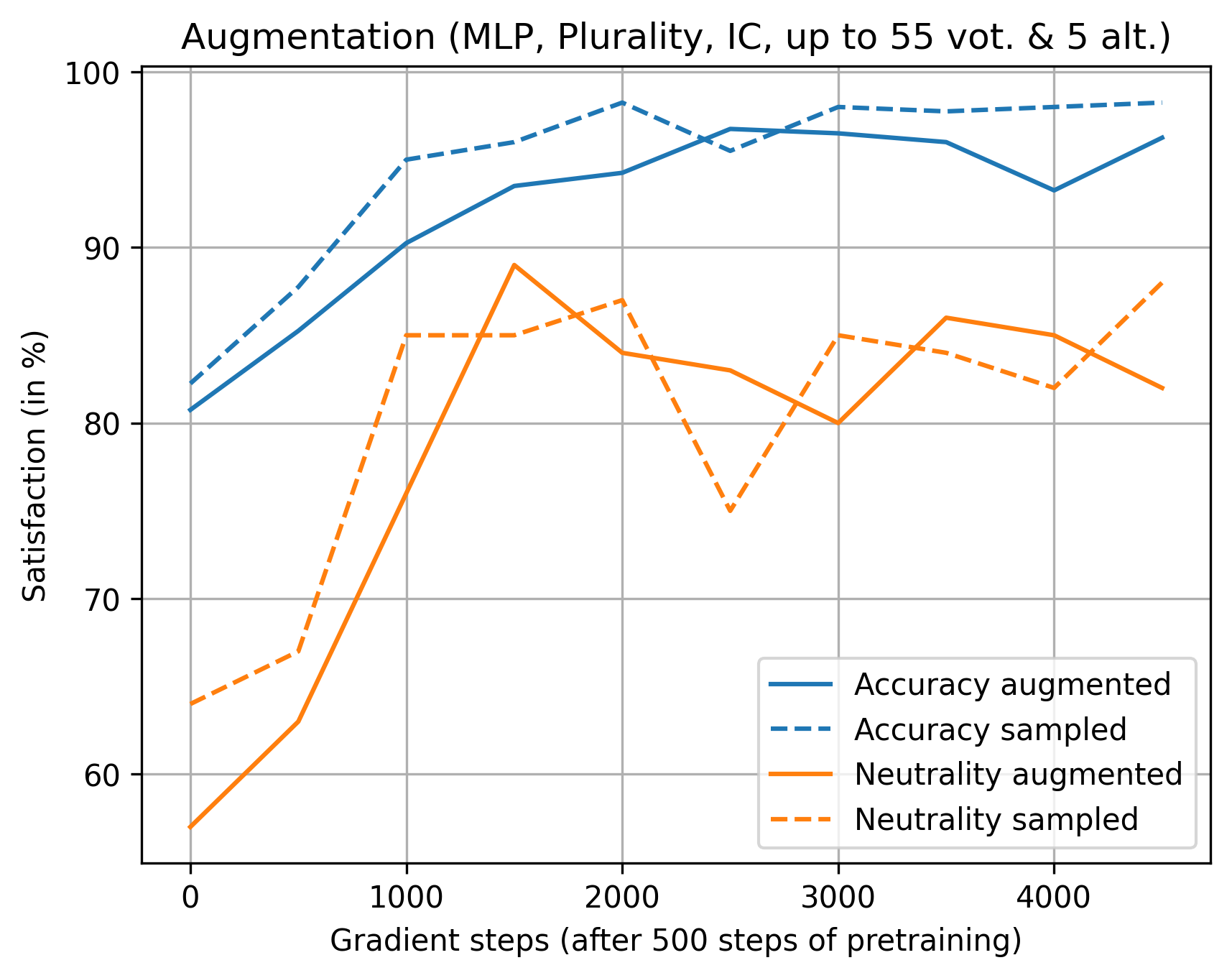}
\includegraphics[width=0.43\linewidth]{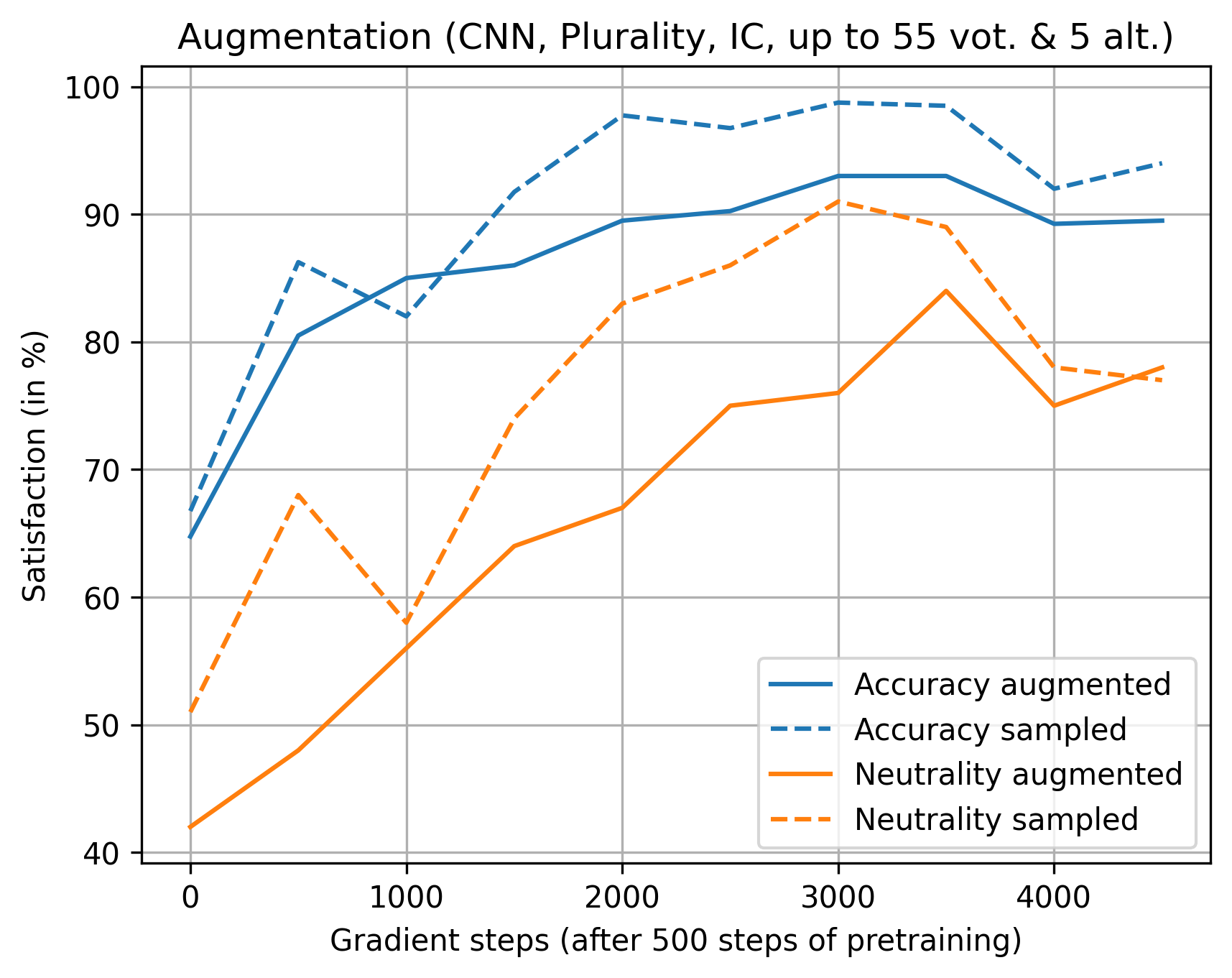}\\
\includegraphics[width=0.43\linewidth]{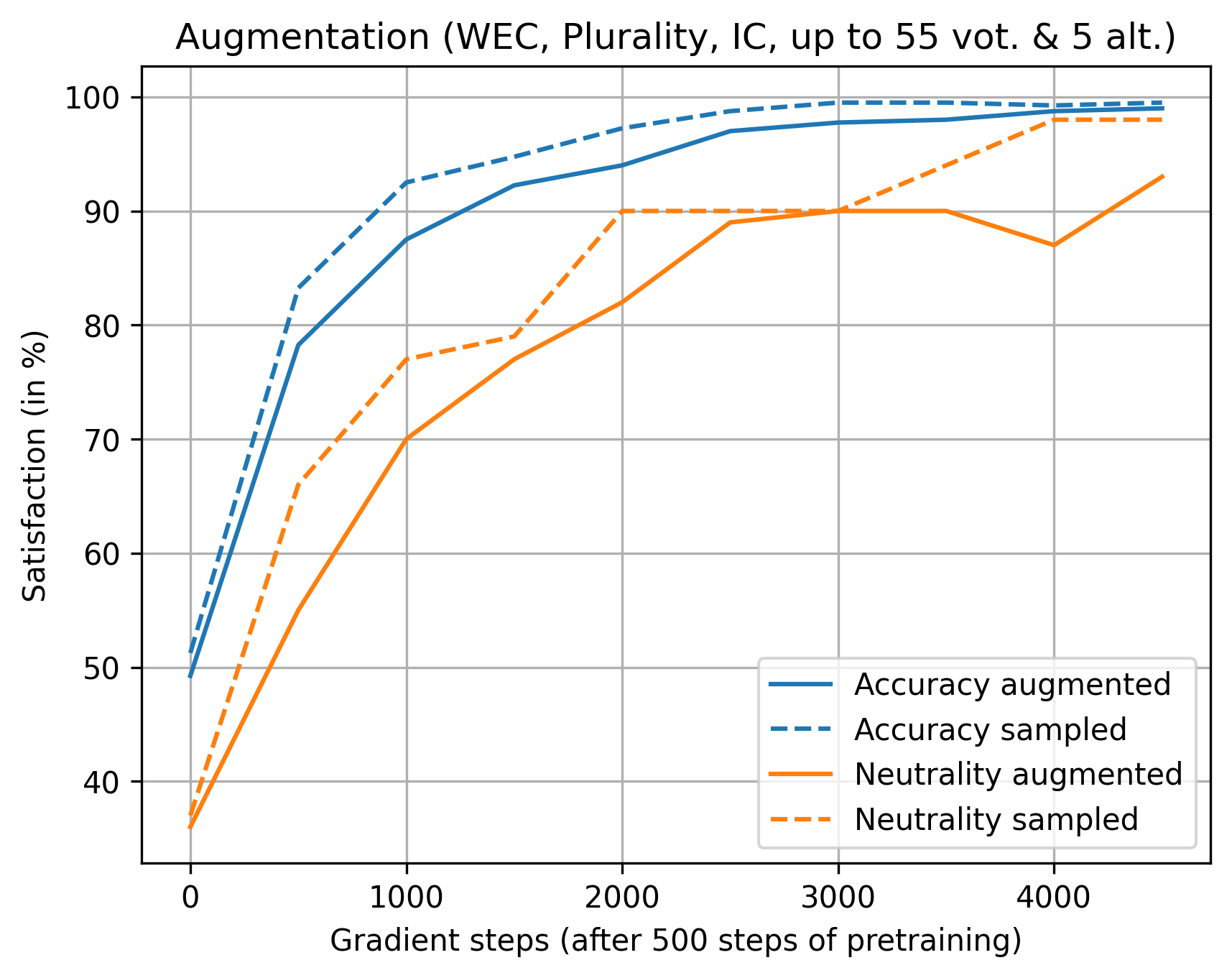}
\includegraphics[width=0.43\linewidth]{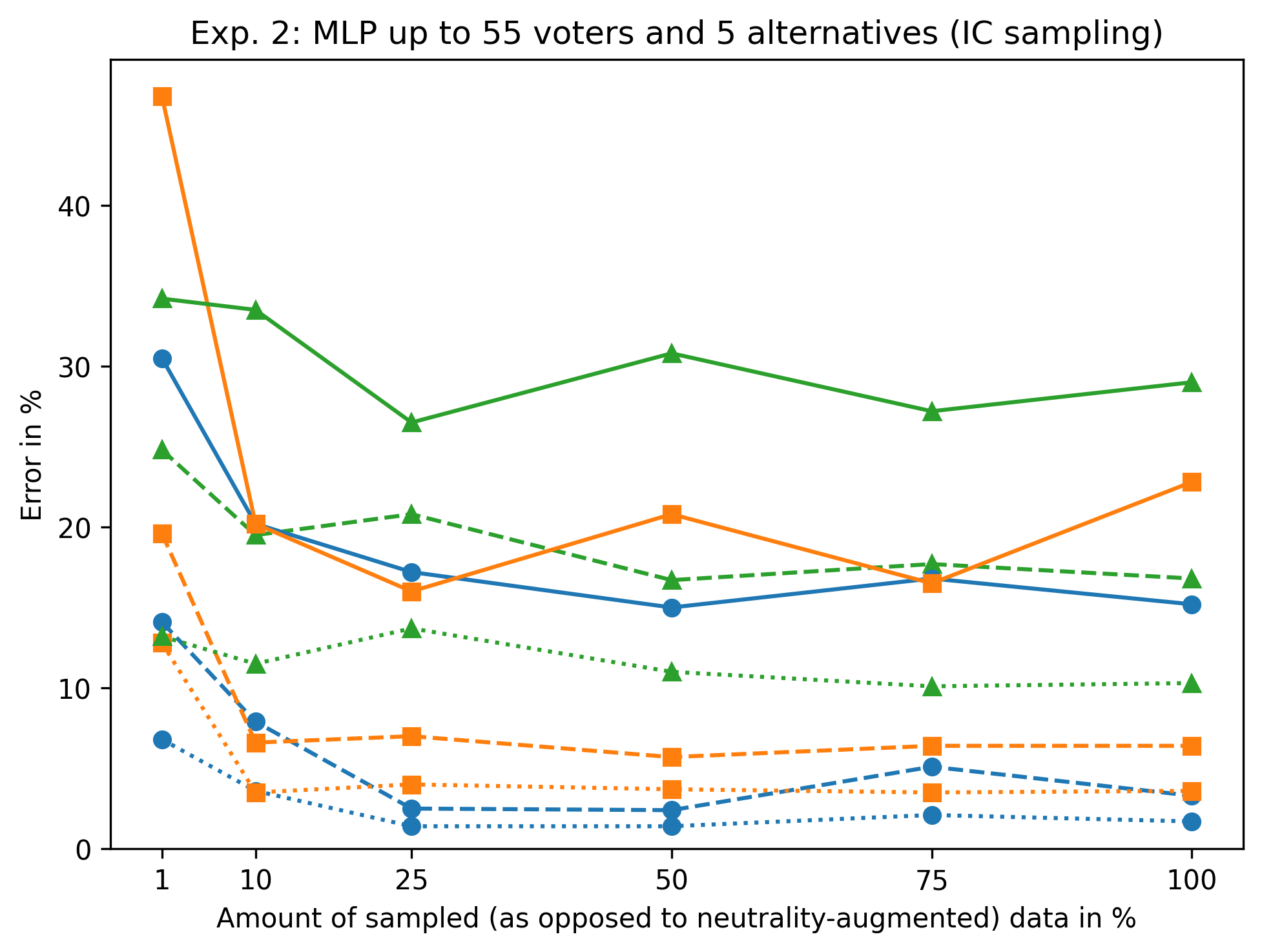}\\
\includegraphics[width=0.43\linewidth]{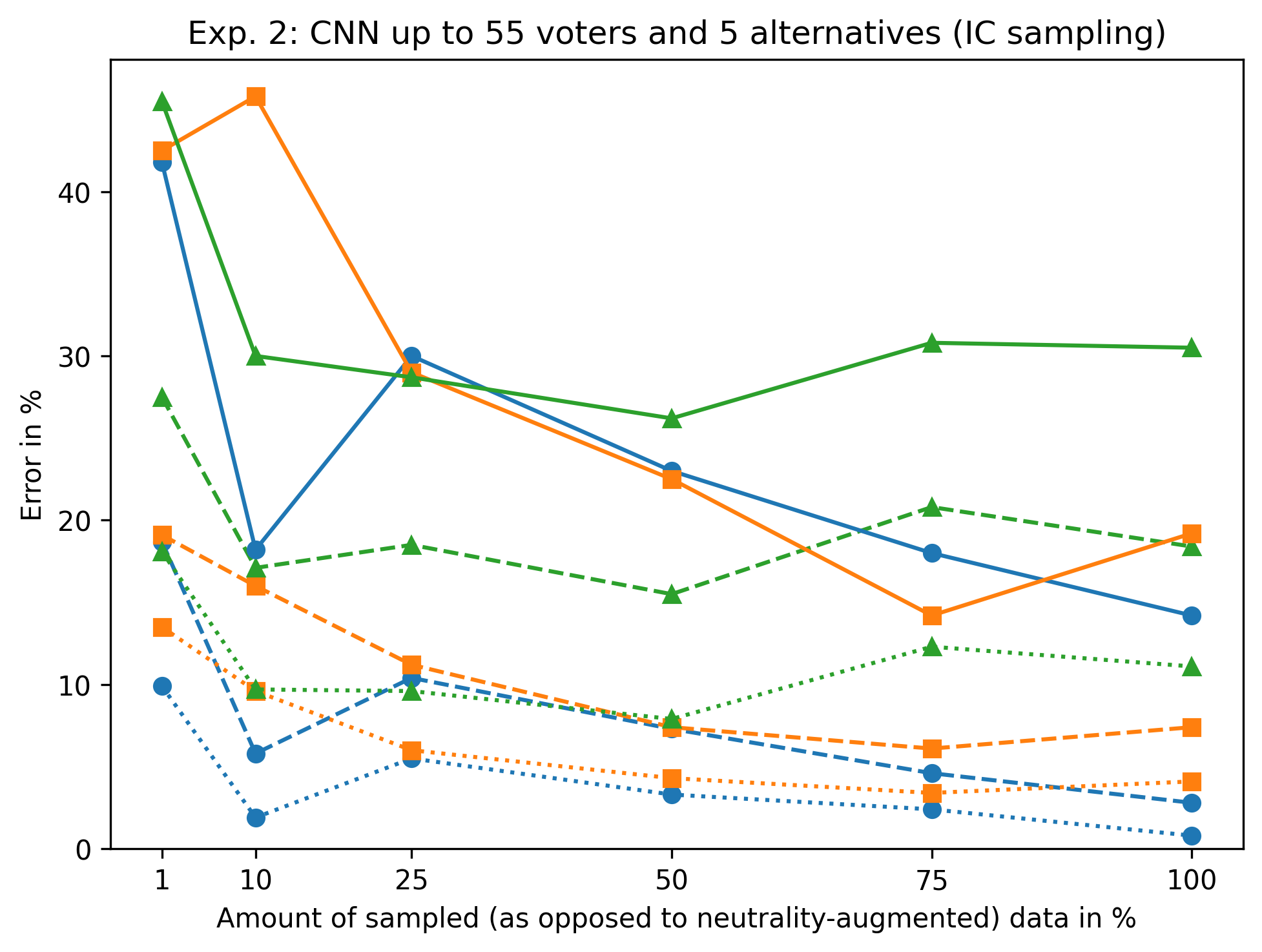}
\includegraphics[width=0.43\linewidth]{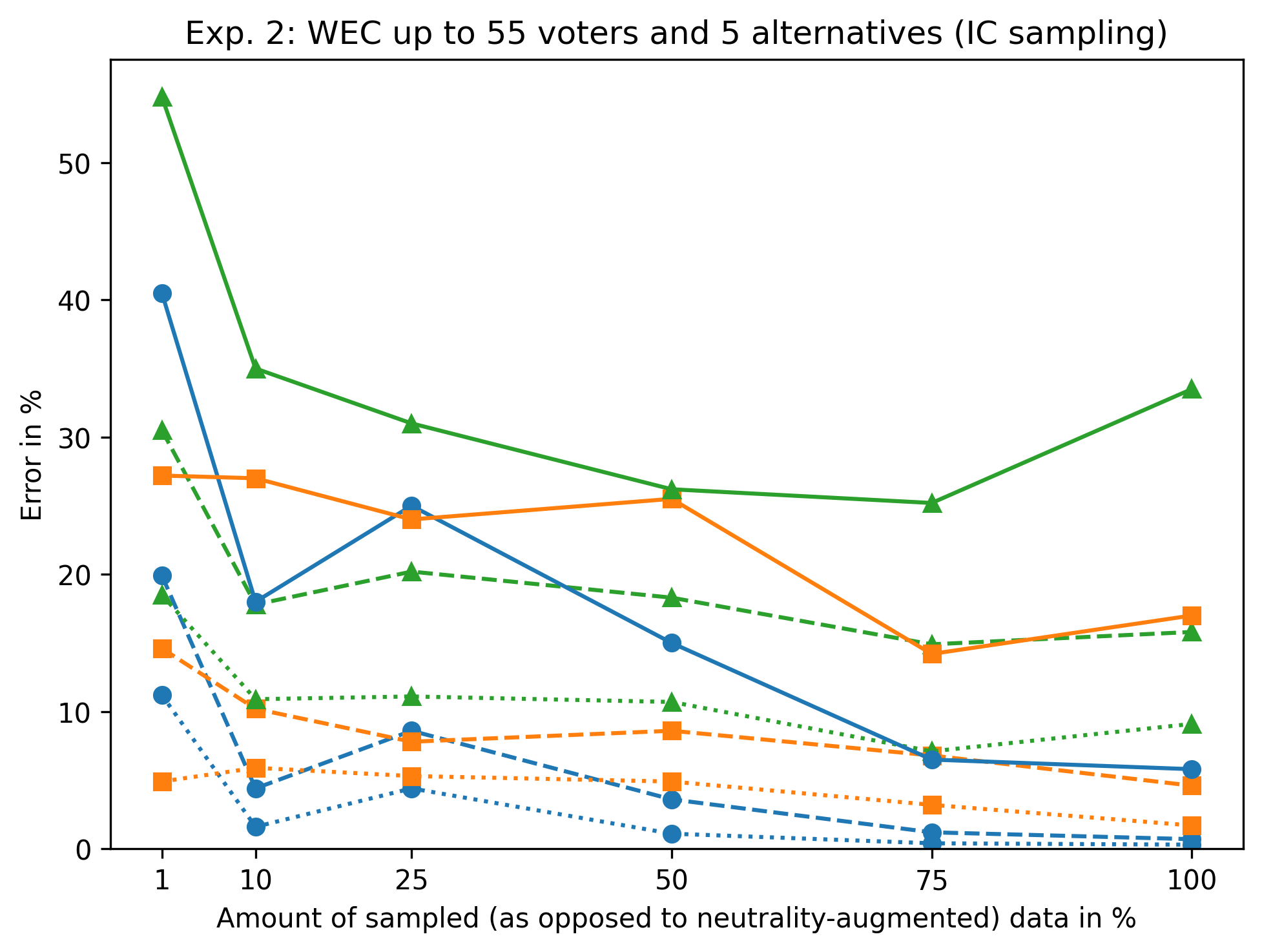}\\
\includegraphics[width=0.7\linewidth]{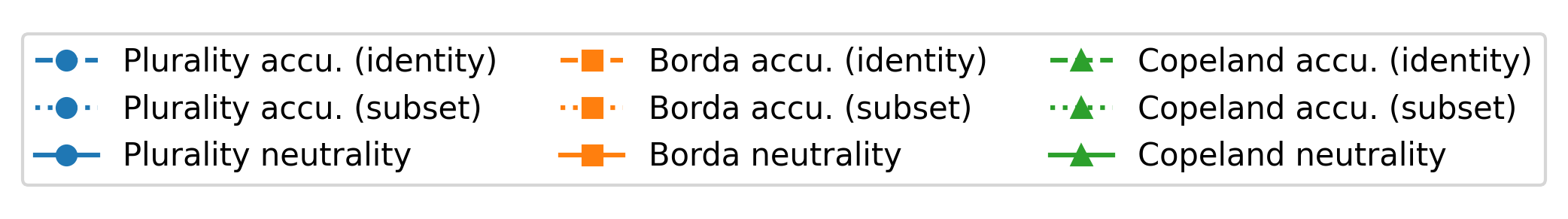}
\Description{Six plots describing the results of the second experiment.}
\caption{Top row and second row on the left: The first version. Pretraining on IC-sampled data from Plurality, continuing with  neutrality variations  and comparing this with identity accuracy to sampled data. 
Rest: The second version. }
\label{fig: exp2}
\end{center}
\end{figure*}

Results for IC-sampling and neutrality augmentation are exhibited in Figure~\ref{fig: exp2}.
(Appendix~\ref{sec: app_exp2} presents results also for other distributions and anonymity augmentation.)
Overall, we find a `yes' to the first question but a `no' to the second: training with augmented data does not hurt accuracy, but it does not reliably improve axiom satisfaction.
In the first version of the experiment depicted at the top row and the left plot of the second row in Figure~\ref{fig: exp2}, augmented data does not seem to be advantageous for neutrality relatively to sampled data (on the contrary, it often seems harmful, e.g., when applying CNN or WEC); in the second version  of the experiment depicted at the bottom row and the right plot of the second row in Figure~\ref{fig: exp2}, the ratio $p$ between sampled and augmented data does not seem to correlate with neutrality satisfaction either.

In more detail, regarding the first version, the conclusion of our experiment~1 is again apparent: even when the models excel in accuracy, their satisfaction of neutrality and anonymity remains consistently below perfect---this does not change when considering augmented data. For the CNN and the WEC, neutrality satisfaction with augmented data is (almost) always below the corresponding one with sampled data, while for the MLP it is mixed. The accuracy of the models is generally not hurt by augmented data. Indeed, it does not seem to vary much between sampled and augmented data. Though, in some cases, data augmentation is detrimental to accuracy: e.g., the identity accuracy of the CNN after 1000 gradient steps (on top of 500 steps of pretraining) is up to 10$\%$ lower with augmented data than with sampled data.

Regarding the second version, when $p < 10\%$, i.e., with almost only augmented data, both accuracy and neutrality satisfaction are unsatisfactory, so data augmentation only becomes relevant for $p \geq 10\%$. Here accuracy is stable: it does not vary by more than $5\%$. In some cases, neutrality is equally stable: for the CNN on all rules and the MLP on Borda (certainly for $p \geq 25\%$, with slightly worse neutrality satisfaction for smaller $p$). In the remaining non-stable cases, the best neutrality satisfaction is achieved for $p = 100\%$, i.e., without augmented data---with only negligible exceptions.\footnote{The only two exceptions are the CNN on Plurality (where neutrality is most satisfied at $p = 75\%$ but to a very similar degree as for $p = 100\%$) and the CNN on Copeland (where neutrality is minimized at $p = 25\%$). Moreover, CNN on Borda and MLP on Copeland have a local---albeit not global---minimum at $p = 25\%$. Thus, while there might be some special cases where neutrality is improved in the highly augmented scenario, this is not enough to consider data augmentation as a successful strategy to improve neutrality satisfaction (which is what we are concerned with here).
}
Thus, neither in the stable nor the unstable cases can we see reliable comparative improvements in neutrality satisfaction with more neutrality augmented data.

\paragraph{Discussion.}
Learning voting-theoretic principles by examples---augmented to the training data---does not seem to work for neural networks: Comparatively more neutrality-augmented or anonymity-augmented data does not necessarily lead to higher neutrality or anonymity satisfaction. However, an advantage of data augmentation is a drastic increase in data efficiency when we only aim for accuracy. For instance, sampling only $10\%$ of the total data set (and using neutrality augmented data for the remaining $90\%$) does not substantially decrease the MLP's or WEC's accuracy in comparison to sampling the whole data set. This is crucial if we use real and not sampled election data, where having access to a vast amount of data points is practically impossible. Even when more data is needed to increase the accuracy of network, we could build an appropriate data set based on a limited amount of real data points and then augment it via the neutrality axiom. This gives us an answer to the two questions raised earlier on: `Yes', training with additional, augmented data points can increase axiom satisfaction, but `no', not better than just training with equally many sampled data points.

\subsection{Experiment 3: Rule Synthesis Guided by Principles?}
\label{ssec: experiment 3}

We saw that neural networks, when trained on data from established voting rules, struggle to vote with principles. This raises the question: can we directly train neural networks to form principled collective decisions, without relying on any pre-existing voting rules? This will be limited by Arrow's Impossibility Theorem~\citep{arrow1951}: a voting rule cannot simultaneously satisfy anonymity, Pareto, and independence. Neural network-based approaches also face this impossibility. However, how close can we get to full axiom satisfaction? We design an optimization task, using custom loss functions, to guide neural networks in learning novel and principled voting rules.

\paragraph{Design.}
We train each one of the three neural network architectures (MLP, CNN, and WEC) on the loss functions defined in Section~\ref{ssec: loss functions} which represent the axioms anonymity, neutrality, Condorcet, Pareto, and independence. Since neural networks could attempt to vacuously satisfy the axioms by proposing no winner, we also consider the ``No-winner'' loss function, which demands the winning sets to be nonempty.
Moreover, by Arrow's Theorem, the axioms cannot be jointly satisfied and will, hence, negatively influence each other. So optimizing for all axioms is not necessarily the best. We pick, for each architecture, a set $\mathcal{O}$ of objectives that we optimize for. For WEC, we choose: no winner, Condorcet, and Pareto. (Appendix~\ref{ssec: app ablation study} establishes in an ablation study the optimality of this choice.) For MLP and CNN, we add: anonymity and independence. 
Then the optimization problem is:
\begin{equation*}
\mathop{\mathrm{argmin}}_w  \sum_{O \in \mathcal{O}} \mathop{{}\mathbb{E}}_{\mathbf{P} \sim D} \big[ L_O ( f_w , \mathbf{P} ) \big],
\end{equation*}
where the loss functions $L_O$ are described in Section~\ref{ssec: loss functions} and $D$ is the chosen distribution of profiles $\mathbf{P}$.
Note that, unlike the previous experiments, this is an unsupervised learning task.

In order to have an architecture that is also neutral by design (not just anonymous by design like the WEC), we design a further decoding function in addition to the one used so far (Section~\ref{ssec: decoding}). This \emph{neutrality-averaged} decoding works as follows~\citep[cf.][]{burka2022voting}. 
Given an input profile, we first generate all alternative-permuted versions of the profile, then compute the logits-predictions of the model on each of those permuted profiles (in one batch), next de-permute the predictions again and average all of them, and finally we turn those average logits into a winning set with the decoding function used so far.

Thus, WEC with neutrality-averaged decoding is anonymous and neutral by design.
For the other architectures, we also test a decoding method that is \emph{neutrality-and-anonymity-averaged}. 
For that, given an input profile, we first randomly generate $12$ alternative-permuted versions of it, and, for each of those, we also randomly generate $10$ voter-permuted versions and, as before, compute the averaged logits and from those the winning set. The numbers are explained as follows: Neutrality-averaging requires, with at most $5$ alternatives, considering at most $5! = 120$ permutations; hence neutrality-and-anonymity-averaging also considers $12 \times 10 = 120$ permutations (checking all $55! \approx 10^{73}$ voter permutations would be infeasible).

\paragraph{Results.}
Table~\ref{tbl: exp 3 IC} shows the axiom satisfaction of different neural networks (bottom) and, for comparison, of several known rules from voting theory (top), all using IC sampling.
(Appendix~\ref{sec: app_exp3} shows the results for the other distributions. See also Figure~\ref{fig: exp3 app loss evolution MLP CNN WEC} in the Appendix for the evolution of the loss during optimization.)

The best neural-network based rule in terms of axiom satisfaction is always the neutrality-averaged WEC, with close contestants the neutrality-averaged CNN and MLP. The neutrality-averaged WEC outperforms the classic Plurality, Borda, and Copeland rules in every single axiom, except for a slight loss on Condorcet. Even when we consider more modern rules in voting theory, the neutrality-averaged WEC is competitive: the existing rule with highest axiom satisfaction is Stable Voting and its edge is marginal, with its average axiom satisfaction being less than $1\%$ higher than that of the neutrality-averaged WEC.\footnote{Table~\ref{tbl: exp 3 continued training} in the Appendix suggests that more gradient steps do not further improve the results.} In fact, when averaging five runs of checking axiom satisfaction (which always involves some stochasticity), the neutrality-averaged WEC even comes out better than the rules: see Table~\ref{tbl: exp 3 averaged training IC} in the Appendix. (This is also true for Euclidean but not for Mallows and Urn.) In any case, the neutrality-averaged WEC has a comparably good axiom satisfaction as the best voting rules known today.

\begin{table*}
\begin{footnotesize}
\begin{center}
\begin{tabular}{lcccccc}
\toprule
                         &  Anon.\  &   Neut.\  &  Condorcet  &  Pareto  &  Indep.\  &  Average  \\
\midrule
 Plurality               &   100   &   100    &    80.2     &   100    &   28.5   &  81.8  \\
 Borda                   &   100   &   100    &    95.5     &   100    &   37.2   &  86.5  \\
 Anti-Plurality          &   100   &   100    &    74.2     &   100    &   24.8   &  79.8  \\
 Copeland                &   100   &   100    &     100     &   100    &    28.0    &  85.6  \\
 Llull                   &   100   &   100    &     100     &   100    &   26.8   &  85.4  \\
 Uncovered Set           &   100   &   100    &     100     &   100    &   27.8   &  85.5  \\
 Top Cycle               &   100   &   100    &     100     &   100    &    29.0    &  85.8  \\
 Banks                   &   100   &   100    &     100     &   100    &   27.8   &  85.5  \\
 Stable Voting           &   100   &   100    &     100     &   100    &    43.0    &  88.6  \\
 Blacks                  &   100   &   100    &     100     &   100    &   35.2   &  87.1  \\
 Instant Runoff TB       &   100   &   100    &    94.8     &   100    &   28.2   &  84.6  \\
 PluralityWRunoff PUT    &   100   &   100    &     95.0      &   100    &   25.5   &  84.1  \\
 Coombs                  &   100   &   100    &    96.2     &   100    &   34.5   &  86.2  \\
 Baldwin                 &   100   &   100    &     100     &   100    &   39.2   &  87.9  \\
 Weak Nanson             &   100   &   100    &     100     &   100    &    40.0    &   88.0   \\
 Kemeny-Young            &   100   &   100    &     100     &   100    &   39.2   &  87.9  \\
\midrule
 MLP p (NW, A, C, P, I)  &  77.8   &   75.8   &    92.5     &   100    &   39.5   &  77.1  \\
 MLP n (NW, A, C, P, I)  &  89.2   &   100    &     95.0      &   100    &   42.2   &  85.3  \\
 MLP na (NW, A, C, P, I) &  89.8   &   86.8   &    95.5     &   100    &   36.5   &  81.7  \\
 CNN p (NW, A, C, P, I)  &  85.2   &   67.2   &     92.0      &   100    &   39.5   &  76.8  \\
 CNN n (NW, A, C, P, I)  &  92.2   &   100    &    94.5     &   100    &    40.0    &  85.4  \\
 CNN na (NW, A, C, P, I) &   86.0    &   86.5   &    94.8     &   100    &    34.0    &  80.2  \\
 WEC p (NW, C, P)        &   100   &   72.5   &    94.2     &   100    &   41.8   &  81.7  \\
 WEC n (NW, C, P)        &   100   &   100    &    96.8     &   100    &   41.2   &  87.6  \\
\bottomrule
\end{tabular}
\end{center}
\end{footnotesize}
\caption{Axiom satisfaction of different rules (top part of the table) and models (bottom part of the table), for IC sampling. Rounded to one decimal. The names of the models are explained as follows: The letters after the architecture type indicate how the voting rule is computed from the model:  
p--plain (i.e., no averaging), 
n--neutrality-averaged, 
na--neutrality-and-anonymity-averaged. The letters in the brackets indicate which axioms the model optimized for during training: 
NW--No winner,
A--Anonymity,
C--Condorcet,
P--Pareto,
I--Independence. 
All models have been trained for 15k gradient steps with batch size 200.}
\label{tbl: exp 3 IC}
\end{table*}

In addition to examining axiom satisfaction, we should also consider how often the examined rules produce the same outcomes: because similar axiom satisfaction does not imply similarity of outcomes.\footnote{For example, the  Blacks and Weak Nanson rules are close in average axiom satisfaction (less than $1\%$ difference), but Table~\ref{tbl: exp 3 similarities IC} shows that more than $8\%$ of the time they propose a different set of winners.} 
Table~\ref{tbl: exp 3 similarities IC} describes similarity in outcome compared to the five closest rules, using IC sampling (again, see Appendix~\ref{sec: app_exp3} for the other distributions). In particular, we see that the rule discovered by the neutrality-averaged WEC model is substantially different from the existing voting rules: it proposes different outcomes than each one of them, according to identity accuracy, at least $9.3\%$ of the time (resp., $10.6\%$ for Mallows, $11.1\%$ for Urn, and $7.8\%$  for Euclidean, see Tables~\ref{tbl: exp 3 similarities Mallows}, \ref{tbl: exp 3 similarities Urn}, and \ref{tbl: exp 3 similarities Euclidean} in the Appendix). In comparison, Stable Voting, which  was found best in Table~\ref{tbl: exp 3 IC}, disagrees with Borda and Copeland $8.9\%$ of the time and with Weak Nanson and Blacks only $6.6\%$ of the time.  
Thus, the discovered rule not only is competitive in axiom satisfaction, it also is novel, i.e., substantially different from existing voting rules. 

Moving from hard to soft accuracy, the neutrality-averaged WEC produces winning sets that are a subset (resp., superset) of those of existing rules at least $95.4\%$ (resp., $95.3\%$) of the time for IC, and similarly for other distributions. This means that even if the results of the model differ from those of existing voting rules, very frequently they do so by only excluding or by only adding certain winning alternatives. This is not unique to our ML model---it is also the case between known voting rules that exhibit high axiomatic satisfaction: for example, the outcome of Stable Voting is a subset (resp., superset) of the outcome of Weak Nanson  $97.9\%$ (resp.,  $94.35\%$) of the time.

\begin{table} 
\begin{footnotesize}
\begin{center}
\begin{tabular}{lcccccc}
\toprule
\emph{Identity accuracy}	&  WEC n  			&  Blacks  			&  Stable Voting	&  Borda  			&  Weak Nanson  	&  Copeland	\\
\midrule
 WEC n         				&   100   			&  91    			&  90.5 			&  89.5   			&  88.4      		&  87.7    	\\
 Blacks        				&   \phantom{91}	&  100    			&  95.71 			&  95.13  			&  91.26     		&  90.57   	\\
 Stable Voting 				&  \phantom{90.5}  	&  \phantom{95.71}	&  100 				&  90.84  			&  93.5      		&  91.67   	\\
 Borda         				&  \phantom{89.5}  	&  \phantom{95.13}	&  \phantom{90.84} 	&  100   			&  86.39     		&  85.7    	\\
 Weak Nanson   				&  \phantom{88.4}  	&  \phantom{91.26}	&  \phantom{93.5}  	&  \phantom{86.39} 	&  100      		&  92.43   	\\
 Copeland      				&  \phantom{87.7}  	&  \phantom{90.57}	&  \phantom{91.67} 	&  \phantom{85.7}  	&  \phantom{92.43} 	&  100     	\\
\bottomrule
\end{tabular}

\medskip
\begin{tabular}{lcccccc}
\toprule
\emph{Subset accuracy}	&  WEC n  &  Blacks  &  Stable Voting  &  Borda  &  Weak Nanson  &  Copeland  \\
\midrule
 WEC n         			&  100    &   93.6   &      92.7       &  93.9   &      93       &    95.3    \\
 Blacks        			&  95.6   &   100    &      96.53      &  97.3   &     95.57     &    98.2    \\
 Stable Voting 			&  95.4   &  97.21   &       100       &  94.51  &     97.9      &   99.59    \\
 Borda         			&  93.8   &  95.13   &      91.66      &   100   &     90.7      &   93.33    \\
 Weak Nanson   			&  92     &  92.69   &      94.35      &  89.99  &      100      &   97.71    \\
 Copeland      			&  90.4   &  91.29   &      91.73      &  88.59  &     94.15     &    100     \\
\bottomrule
\end{tabular}
\end{center}
\end{footnotesize}
\caption{Similarities between the rules. Computed on 10,000 IC-sampled profiles. For example, in the top table (`identity accuracy'), the entry $90.5$ in row `WEC n' and column `Stable Voting' means that, in $90.5\%$ of the sampled profiles, Stable Voting outputs the identical winning set as the neutrality-averaged WEC. Hence the values below the diagonal are symmetric and thus omitted. In the bottom table (`subset accuracy'), the entries show when the rule in the row outputs a winning set that is a subset of the rule in the column. So the entry $92.7$ in row `WEC n' and column `Stable Voting' means that, in $92.7\%$ of the sampled profiles, the winning set outputted by the neutrality-averaged WEC is a subset of the winning set outputted by Stable Voting. This table is not symmetric, because the entry $95.4$ in row `Stable Voting' and column `WEC n' means that, in $95.4\%$ of the sampled profiles, the winning set outputted by Stable Voting is a subset of the model's winning set (equivalently, the model's winning set is a superset of the rule's winning set).
}
\label{tbl: exp 3 similarities IC}
\end{table}

To illustrate the difference between the discovered and the existing rules, Figure~\ref{fig: exp 3 differing profile IC} shows an example of a profile where the winning set provided by the neutrality-averaged WEC is different to all the winning sets provided by the considered existing voting rules. The choice of the WEC also has intuitive plausibility: it chooses alternative $a$ which, among the eight voters, is three times the most preferred option and two times the second-most preferred option.  The sigmoids indicate that alternative $b$ was also a close competitor for the winning set, and would indeed win under many of the known rules from voting theory.

\floatstyle{boxed}
\restylefloat{figure}
\begin{figure}
\centering
\begin{footnotesize}
\begin{tabular}{cccccccc}
\toprule
  1 &  2  &  3  &  4  &  5  &  6  &  7  &  8  \\
\midrule
$a$ & $e$ & $d$ & $a$ & $e$ & $b$ & $e$ & $a$ \\
$b$ & $b$ & $b$ & $c$ & $b$ & $a$ & $a$ & $b$ \\
$e$ & $d$ & $c$ & $b$ & $c$ & $e$ & $c$ & $d$ \\
$d$ & $a$ & $e$ & $e$ & $a$ & $c$ & $d$ & $e$ \\
$c$ & $c$ & $a$ & $d$ & $d$ & $d$ & $b$ & $c$ \\
\bottomrule
\end{tabular} \\
\smallskip
\begin{tabular}{rl}
$\{ a \}$ & neutrality-averaged WEC, with sigmoids (rounded)
$a$:.51, 
$b$:.49, 
$c$:.31, 
$d$:.32, 
$e$:.43  
\\  
$\{ b \}$ & Blacks, Stable Voting, Borda, Weak Nanson, Copeland\\
$\{ a, e \}$ & Plurality, PluralityWRunoff PUT\\
$\{ e \}$ & Instant Runoff TB, Anti-Plurality\\
$\{ a,b \}$ & Llull, Uncovered Set, Banks, Coombs, Baldwin, and Kemeny-Young\\
$\{ a, b, e \}$ & Top Cycle\\
\end{tabular}
\end{footnotesize}
\Description{A profile on which the `WEC n' model disagress with existing voting rules.}
\caption{Profile where the `WEC n' model disagrees with existing voting rules. The winning sets for each rule are mentioned below the table.}
\label{fig: exp 3 differing profile IC}
\end{figure}
\floatstyle{plain}
\restylefloat{figure}

\paragraph{Discussion.}
The reason why the WEC outperforms the other two architectures is that, because it is anonymous by design, it is enough to use the neutrality-averaged decoding to get a model that is anonymous and neutral. Since the MLP and CNN are not anonymous, they need neutrality-and-anonymity-averaged decoding to become anonymous and neutral by design. This, however, needs infeasibly many permutations, so it can only be approximated via sampling permutations. Here, however, the tension between the axioms of anonymity and neutrality resurfaces: sampled neutrality-and-anonymity-averaging can result in negative interference with the other axioms yielding worse performance than just neutrality-averaging (e.g., the CNN rules in Table~\ref{tbl: exp 3 IC}).
Mere neutrality-averaging also influences the satisfaction of the other axioms, but in this case not in a negative way.\footnote{We did not use neutrality-averaging in the previous experiments because it would not directly correspond to the binary cross entropy loss and the interference with the other axioms blurs the axiomatic evaluation of the neural network.}

Moreover, for the WEC just three optimization objectives were enough to obtain the above competitive results. Since the MLP and CNN are not anonymous by design, they needed to optimize for anonymity and independence as well. The WEC interestingly had enough \emph{implicit} inductive bias toward satisfying independence---again highlighting non-trivial interference of the axioms and the network architecture.

The neural networks beat the classic voting rules in terms of axiom satisfaction while being comparable to the best voting rules known today. This may be taken to suggest that existing rules may already be close to optimal axiom satisfaction. In other words, they are in the (approximate) \emph{Pareto front} of axiom satisfaction. At the same time, even if the novel rules derived from axiom optimization inherit the opacity of neural networks, they assure high adherence to key normative principles in collective decisions. Since these newly discovered rules were substantially different from existing rules, they extend the boundaries of what is so far explored in voting theory.

\section{Discussion}
\label{sec: discussion}

With our axiomatic deep voting framework, we investigated the space of all voting rules by fruitfully combing voting theory and machine learning.
The neural network explores the space and the voting-theoretic axioms evaluate the network, thus guiding the exploration. 
The universal approximation theorems~\citep{Hornik1989, Cybenko1989} ensure that the neural networks are dense in the space of all voting rules, so all areas of that space can be explored with axiomatic deep voting. 
Arrow's Impossibility Theorem~\citep{arrow1951} establishes insurmountable divisions of that space: e.g., the area of rules satisfying anonymity and Pareto does not intersect the area of rules satisfying independence.

The importance of our results for AI is twofold. First, the axiomatic evaluation offers another cautionary tale that accuracy is not everything: Neural networks can have high accuracy (descriptively good) without following the right reasons (normatively bad).
Second, this changes, however, when we move from the supervised setting of learning rules from examples to the unsupervised setting of directly optimizing axiom satisfaction. We were able to do this by translating the voting-theoretic axioms into corresponding loss functions. 
Having a way to optimize the axioms is important because the axioms can be seen as \emph{mathematical formalizations} of important normative notions in modern machine learning. For example:
\begin{itemize}
\item
\emph{Bias}: anonymity says that the neural network is not biased towards particular individuals.
\item
\emph{Fairness}: neutrality demands that the neural network treats all alternatives equally. 
\item
\emph{Value-alignment}: the Pareto principle requires that if all individuals value one alternative more than another, then the neural network aligns with this; and similarly for the Condorcet principle.
\item
\emph{Interpretability}: 
independence provides a sense of `compositionality' when interpreting the network---to understand its choice for two given alternatives, we can ignore all other alternatives.
\end{itemize}
Hence, our axiomatic optimization provides a way of improving the neural network---in a mathematically precise sense---regarding bias, fairness, value-alignment, and interpretability.

Moreover, qua interdisciplinary project, our results are also relevant for voting theory. Axiomatic deep voting offers a new tool for the field's central goal of exploring the space of all voting rules. While existing voting rules are crafted by human insight, we could find---in a completely automated process---novel voting rules that are comparable in terms of axiom satisfaction to the best rules known today. This provides a promising starting point for an analytic exploration of new axiom-optimal voting rules and the influence the axioms exert on each other.

\paragraph{Limitations.}
We tested a wide range of standard neural network architectures. However, future work could also investigate further architectures like Set Transformers, Graph Isomorphism Networks, or Deep Sets (which were used by~\citet{anil2021learning}) and, more generally, the transformer architecture (as a refinement of our word embedding architecture). 
We also covered the most important voting-theoretic axioms, but yet more can be considered, e.g., monotonicitiy and transitivity (the latter then requires architectures that are not transitive by design).
Finally, the large number of permutations causes a high statistical variance in testing the satisfaction of the independence axiom.

\paragraph{Future work.}
First, more options in generating the dataset can be explored. For example, we can consider the \emph{extrapolation} task in which the model has to find a general rule after only observing the rule on a small part of the input space, namely the profiles where some given voting rules agree or satisfy a given axiom. Or we can consider the \emph{interpolation} task in which the model sees data of different rules and has to find a compromise between their outputs.

Second, we can implement further social choice theory frameworks. Since we already output logits corresponding to the alternatives, we could, instead of winning sets, also consider preference rankings or welfare functions. It would also be interesting to consider judgment aggregation, which includes reasoning about logical implications between the alternatives.

Third, it seems promising to bridge notions of explainability in voting theory \citep{CaillouxEndrissAAMAS2016, NardiEtAlAAMAS2022,BoixelEtAlAAAI2022} and notions of explainability in AI \citep{Adadi2018}. In particular, is it possible to extract out a symbolic representation (e.g., in logic programming) of the rule that the model learned?

Fourth, studying the voting-theoretic concept of manipulatability via neural networks \citep{holliday2024learning} can be further connected to machine learning notions like \emph{adversarial attacks} \citep{Goodfellow2014} or \emph{performativity} \citep{Perdomo2020}.

Fifth, from the point of view of \emph{geometric deep learning},\footnote{See the work of Bronstein et al., 2021. Geometric Deep Learning: Grids, Groups, Graphs, Geodesics,
and Gauges. arXiv:2104.13478.} axioms represent \emph{symmetries} that the neural networks should learn. For example, anonymity says that the neural network should be invariant under the group action of the voter-permutation group on profiles; and neutrality says that the neural network should be equivariant under the group action of the alternative-permutation group on profiles and winning sets, respectively. It seems worth exploring this connection to geometric deep learning.

\section{Conclusion} 
\label{sec: conclusion}

We introduced the axiomatic deep voting framework to study how neural networks aggregate preferences. We found that neural networks \emph{do not} learn to vote with principles, despite achieving high accuracy, when trained on data from existing voting rules---even when augmented with axiom-specific data. However, they \emph{do} learn to vote with principles when they directly optimize for axiom satisfaction, which we achieved by translating axioms into custom loss functions.
The axiomatic deep voting framework promises fruitful further investigation both in voting theory (new ways of exploring the space of voting rules) and AI (a mathematically precise testing ground for normative notions like bias and value-alignment).

\begin{acks}
For very helpful comments and discussions, we would like to thank Ben Armstrong, Balder ten Cate, Timo Freiesleben, Ronald de Haan, Thomas Icard, Alina Leidinger, Christian List, Ignacio Ojea, as well as the audiences at the `Workshop on Learning and Logic 2025' at the University of Amsterdam and the `Social Choice for AI Ethics and Safety 2025' workshop at AAMAS in Detroit. We are especially grateful to the JAIR editor Alessandro Farinelli and two anonymous reviewers.
\end{acks}

\bibliographystyle{ACM-Reference-Format}
\bibliography{literature}


\begin{thebibliography}{55}


\ifx \showCODEN    \undefined \def \showCODEN     #1{\unskip}     \fi
\ifx \showDOI      \undefined \def \showDOI       #1{#1}\fi
\ifx \showISBNx    \undefined \def \showISBNx     #1{\unskip}     \fi
\ifx \showISBNxiii \undefined \def \showISBNxiii  #1{\unskip}     \fi
\ifx \showISSN     \undefined \def \showISSN      #1{\unskip}     \fi
\ifx \showLCCN     \undefined \def \showLCCN      #1{\unskip}     \fi
\ifx \shownote     \undefined \def \shownote      #1{#1}          \fi
\ifx \showarticletitle \undefined \def \showarticletitle #1{#1}   \fi
\ifx \showURL      \undefined \def \showURL       {\relax}        \fi
\providecommand\bibfield[2]{#2}
\providecommand\bibinfo[2]{#2}
\providecommand\natexlab[1]{#1}
\providecommand\showeprint[2][]{arXiv:#2}

\bibitem[Adadi and Berrada(2018)]%
        {Adadi2018}
\bibfield{author}{\bibinfo{person}{A. Adadi} {and} \bibinfo{person}{M.
  Berrada}.} \bibinfo{year}{2018}\natexlab{}.
\newblock \showarticletitle{Peeking Inside the Black-Box: {A} Survey on
  Explainable Artificial Intelligence ({XAI})}.
\newblock \bibinfo{journal}{\emph{IEEE Access}}  \bibinfo{volume}{6}
  (\bibinfo{year}{2018}), \bibinfo{pages}{52138--52160}.
\newblock


\bibitem[Anil and Bao(2021)]%
        {anil2021learning}
\bibfield{author}{\bibinfo{person}{Cem Anil} {and} \bibinfo{person}{Xuchan
  Bao}.} \bibinfo{year}{2021}\natexlab{}.
\newblock \showarticletitle{Learning to Elect}.
\newblock \bibinfo{journal}{\emph{Advances in Neural Information Processing
  Systems}}  \bibinfo{volume}{34} (\bibinfo{year}{2021}),
  \bibinfo{pages}{8006--8017}.
\newblock


\bibitem[Armstrong and Larson(2019)]%
        {armstrong2019machine}
\bibfield{author}{\bibinfo{person}{Ben Armstrong} {and} \bibinfo{person}{Kate
  Larson}.} \bibinfo{year}{2019}\natexlab{}.
\newblock \showarticletitle{Machine Learning to Strengthen Democracy}. In
  \bibinfo{booktitle}{\emph{NeurIPS Joint Workshop on AI for Social Good}}.
\newblock


\bibitem[Arrow(1951)]%
        {arrow1951}
\bibfield{author}{\bibinfo{person}{Kenneth~J Arrow}.}
  \bibinfo{year}{1951}\natexlab{}.
\newblock \bibinfo{booktitle}{\emph{Social Choice and Individual Values}}.
\newblock \bibinfo{publisher}{John Wiley $\&$ Sons}.
\newblock


\bibitem[Bender and Koller(2020)]%
        {Bender2020}
\bibfield{author}{\bibinfo{person}{Emily~M. Bender} {and}
  \bibinfo{person}{Alexander Koller}.} \bibinfo{year}{2020}\natexlab{}.
\newblock \showarticletitle{Climbing towards {NLU}: {On} Meaning, Form, and
  Understanding in the Age of Data}. In \bibinfo{booktitle}{\emph{Proceedings
  of the 58th Annual Meeting of the Association for Computational
  Linguistics}}. \bibinfo{pages}{5185--5198}.
\newblock


\bibitem[Boehmer et~al\mbox{.}(2021)]%
        {boehmer2021putting}
\bibfield{author}{\bibinfo{person}{Niclas Boehmer}, \bibinfo{person}{Robert
  Bredereck}, \bibinfo{person}{Piotr Faliszewski}, \bibinfo{person}{Rolf
  Niedermeier}, {and} \bibinfo{person}{Stanis{\l}aw Szufa}.}
  \bibinfo{year}{2021}\natexlab{}.
\newblock \showarticletitle{Putting a Compass on the Map of Elections}. In
  \bibinfo{booktitle}{\emph{Proceedings of the 30th International Joint
  Conference on Artificial Intelligence (IJCAI)}}.
\newblock


\bibitem[Boehmer et~al\mbox{.}(2023)]%
        {boehmer2023properties}
\bibfield{author}{\bibinfo{person}{Niclas Boehmer}, \bibinfo{person}{Piotr
  Faliszewski}, {and} \bibinfo{person}{Sonja Kraiczy}.}
  \bibinfo{year}{2023}\natexlab{}.
\newblock \showarticletitle{Properties of the Mallows Model Depending on the
  Number of Alternatives: {A} Warning for an Experimentalist}. In
  \bibinfo{booktitle}{\emph{Proceedings of the 40th International Conference on
  Machine Learning (ICML)}}.
\newblock


\bibitem[Boixel et~al\mbox{.}(2022)]%
        {BoixelEtAlAAAI2022}
\bibfield{author}{\bibinfo{person}{Arthur Boixel}, \bibinfo{person}{Ulle
  Endriss}, {and} \bibinfo{person}{Ronald de Haan}.}
  \bibinfo{year}{2022}\natexlab{}.
\newblock \showarticletitle{A Calculus for Computing Structured Justifications
  for Election Outcomes}. In \bibinfo{booktitle}{\emph{Proceedings of the 36th
  AAAI Conference on Artificial Intelligence (AAAI)}}.
\newblock


\bibitem[Brandt et~al\mbox{.}(2016)]%
        {HBCOMSOC2016}
\bibfield{editor}{\bibinfo{person}{Felix Brandt}, \bibinfo{person}{Vincent
  Conitzer}, \bibinfo{person}{Ulle Endriss}, \bibinfo{person}{J\'er\^ome Lang},
  {and} \bibinfo{person}{Ariel~D. Procaccia}} (Eds.).
  \bibinfo{year}{2016}\natexlab{}.
\newblock \bibinfo{booktitle}{\emph{Handbook of Computational Social Choice}}.
\newblock \bibinfo{publisher}{Cambridge University Press}.
\newblock


\bibitem[Burka et~al\mbox{.}(2022)]%
        {burka2022voting}
\bibfield{author}{\bibinfo{person}{D{\'a}vid Burka}, \bibinfo{person}{Clemens
  Puppe}, \bibinfo{person}{L{\'a}szl{\'o} Szepesv{\'a}ry}, {and}
  \bibinfo{person}{Attila Tasn{\'a}di}.} \bibinfo{year}{2022}\natexlab{}.
\newblock \showarticletitle{Voting: {A} Machine Learning Approach}.
\newblock \bibinfo{journal}{\emph{European Journal of Operational Research}}
  \bibinfo{volume}{299}, \bibinfo{number}{3} (\bibinfo{year}{2022}),
  \bibinfo{pages}{1003--1017}.
\newblock


\bibitem[Cailloux and Endriss(2016)]%
        {CaillouxEndrissAAMAS2016}
\bibfield{author}{\bibinfo{person}{Olivier Cailloux} {and}
  \bibinfo{person}{Ulle Endriss}.} \bibinfo{year}{2016}\natexlab{}.
\newblock \showarticletitle{Arguing about Voting Rules}. In
  \bibinfo{booktitle}{\emph{Proceedings of the 15th International Conference on
  Autonomous Agents and Multiagent Systems (AAMAS)}}.
\newblock


\bibitem[Caragiannis and Micha(2017)]%
        {caragiannis2017learning}
\bibfield{author}{\bibinfo{person}{Ioannis Caragiannis} {and}
  \bibinfo{person}{Evi Micha}.} \bibinfo{year}{2017}\natexlab{}.
\newblock \showarticletitle{Learning a Ground Truth Ranking Using Noisy
  Approval Votes}. In \bibinfo{booktitle}{\emph{Proceedings of the 26th
  International Joint Conference on Artificial Intelligence (IJCAI)}}.
\newblock


\bibitem[Conitzer et~al\mbox{.}(2024)]%
        {conitzerposition}
\bibfield{author}{\bibinfo{person}{Vincent Conitzer}, \bibinfo{person}{Rachel
  Freedman}, \bibinfo{person}{Jobst Heitzig}, \bibinfo{person}{Wesley
  Holliday}, \bibinfo{person}{Bob Jacobs}, \bibinfo{person}{Nathan Lambert},
  \bibinfo{person}{Milan Moss{\'e}}, \bibinfo{person}{Eric Pacuit},
  \bibinfo{person}{Stuart Russell}, \bibinfo{person}{Hailey Schoelkopf},
  \bibinfo{person}{Emanuel Tewolde}, {and} \bibinfo{person}{William Zwicker}.}
  \bibinfo{year}{2024}\natexlab{}.
\newblock \showarticletitle{Position: {S}ocial Choice Should Guide AI Alignment
  in Dealing with Diverse Human Feedback}. In
  \bibinfo{booktitle}{\emph{Proceedings of the 41st International Conference on
  Machine Learning (ICML)}}.
\newblock


\bibitem[Cybenko(1989)]%
        {Cybenko1989}
\bibfield{author}{\bibinfo{person}{G. Cybenko}.}
  \bibinfo{year}{1989}\natexlab{}.
\newblock \showarticletitle{Approximations by Superpositions of Sigmoidal
  Functions}.
\newblock \bibinfo{journal}{\emph{Mathematics of Control, Signals, and
  Systems}} \bibinfo{volume}{2}, \bibinfo{number}{4} (\bibinfo{year}{1989}),
  \bibinfo{pages}{303--314}.
\newblock


\bibitem[Dai and Fleisig(2024)]%
        {dai2024mapping}
\bibfield{author}{\bibinfo{person}{Jessica Dai} {and} \bibinfo{person}{Eve
  Fleisig}.} \bibinfo{year}{2024}\natexlab{}.
\newblock \showarticletitle{Mapping Social Choice Theory to {RLHF}}. In
  \bibinfo{booktitle}{\emph{ICLR Workshop on Reliable and Responsible
  Foundation Models}}.
\newblock


\bibitem[Dougherty and Heckelman(2020)]%
        {dougherty2020probability}
\bibfield{author}{\bibinfo{person}{Keith~L Dougherty} {and}
  \bibinfo{person}{Jac~C Heckelman}.} \bibinfo{year}{2020}\natexlab{}.
\newblock \showarticletitle{The Probability of Violating {A}rrow’s
  Conditions}.
\newblock \bibinfo{journal}{\emph{European Journal of Political Economy}}
  \bibinfo{volume}{65} (\bibinfo{year}{2020}), \bibinfo{pages}{101936}.
\newblock


\bibitem[Eggenberger and P{\'o}lya(1923)]%
        {eggenberger1923statistik}
\bibfield{author}{\bibinfo{person}{Florian Eggenberger} {and}
  \bibinfo{person}{George P{\'o}lya}.} \bibinfo{year}{1923}\natexlab{}.
\newblock \showarticletitle{{\"U}ber die Statistik Verketteter Vorg{\"a}nge}.
\newblock \bibinfo{journal}{\emph{ZAMM-Journal of Applied Mathematics and
  Mechanics/Zeitschrift f{\"u}r Angewandte Mathematik und Mechanik}}
  \bibinfo{volume}{3}, \bibinfo{number}{4} (\bibinfo{year}{1923}),
  \bibinfo{pages}{279--289}.
\newblock


\bibitem[Favardin and Lepelley(2006)]%
        {favardin2006some}
\bibfield{author}{\bibinfo{person}{Pierre Favardin} {and}
  \bibinfo{person}{Dominique Lepelley}.} \bibinfo{year}{2006}\natexlab{}.
\newblock \showarticletitle{Some Further Results on the Manipulability of
  Social Choice Rules}.
\newblock \bibinfo{journal}{\emph{Social Choice and Welfare}}
  (\bibinfo{year}{2006}), \bibinfo{pages}{485--509}.
\newblock


\bibitem[Favardin et~al\mbox{.}(2002)]%
        {favardin2002borda}
\bibfield{author}{\bibinfo{person}{Pierre Favardin}, \bibinfo{person}{Dominique
  Lepelley}, {and} \bibinfo{person}{J{\'e}r{\^o}me Serais}.}
  \bibinfo{year}{2002}\natexlab{}.
\newblock \showarticletitle{{B}orda rule, {C}opeland Method and Strategic
  Manipulation}.
\newblock \bibinfo{journal}{\emph{Review of Economic Design}}
  \bibinfo{volume}{7} (\bibinfo{year}{2002}), \bibinfo{pages}{213--228}.
\newblock


\bibitem[Fishburn and Gehrlein(1982)]%
        {fishburn1982majority}
\bibfield{author}{\bibinfo{person}{Peter~C Fishburn} {and}
  \bibinfo{person}{William~V Gehrlein}.} \bibinfo{year}{1982}\natexlab{}.
\newblock \showarticletitle{Majority Efficiencies for Simple Voting Procedures:
  {S}ummary and Interpretation}.
\newblock \bibinfo{journal}{\emph{Theory and Decision}} \bibinfo{volume}{14},
  \bibinfo{number}{2} (\bibinfo{year}{1982}), \bibinfo{pages}{141--153}.
\newblock


\bibitem[Goodfellow et~al\mbox{.}(2016)]%
        {Goodfellow2016}
\bibfield{author}{\bibinfo{person}{Ian Goodfellow}, \bibinfo{person}{Yoshua
  Bengio}, {and} \bibinfo{person}{Aaron Courville}.}
  \bibinfo{year}{2016}\natexlab{}.
\newblock \bibinfo{booktitle}{\emph{Deep Learning}}.
\newblock \bibinfo{publisher}{The MIT Press}, \bibinfo{address}{Cambridge,
  Massachusetts}.
\newblock


\bibitem[Goodfellow et~al\mbox{.}(2015)]%
        {Goodfellow2014}
\bibfield{author}{\bibinfo{person}{Ian~J. Goodfellow},
  \bibinfo{person}{Jonathon Shlens}, {and} \bibinfo{person}{Christian
  Szegedy}.} \bibinfo{year}{2015}\natexlab{}.
\newblock \showarticletitle{Explaining and Harnessing Adversarial Examples}. In
  \bibinfo{booktitle}{\emph{International Conference on Learning
  Representations (ICLR)}}.
\newblock


\bibitem[Gudi\~{n}o Rosero et~al\mbox{.}(2024)]%
        {gudino2024llms}
\bibfield{author}{\bibinfo{person}{Jairo Gudi\~{n}o Rosero},
  \bibinfo{person}{Umberto Grandi}, {and} \bibinfo{person}{C\'{e}sar~A.
  Hidalgo}.} \bibinfo{year}{2024}\natexlab{}.
\newblock \showarticletitle{Large Language Models ({LLM}s) as Agents for
  Augmented Democracy}.
\newblock \bibinfo{journal}{\emph{Philosophical Transactions A}}
  \bibinfo{volume}{382}, \bibinfo{number}{2285} (\bibinfo{year}{2024}),
  \bibinfo{pages}{20240100}.
\newblock


\bibitem[Hatzivelkos(2018)]%
        {hatzivelkos2018borda}
\bibfield{author}{\bibinfo{person}{Aleksandar Hatzivelkos}.}
  \bibinfo{year}{2018}\natexlab{}.
\newblock \showarticletitle{{B}orda and {P}lurality Comparison with Regard to
  Compromise as a Sorites Paradox}.
\newblock \bibinfo{journal}{\emph{Interdisciplinary Description of Complex
  Systems: INDECS}} \bibinfo{volume}{16}, \bibinfo{number}{3B}
  (\bibinfo{year}{2018}), \bibinfo{pages}{465--484}.
\newblock


\bibitem[Holliday and Pacuit(2023a)]%
        {holliday2023split}
\bibfield{author}{\bibinfo{person}{Wesley Holliday} {and} \bibinfo{person}{Eric
  Pacuit}.} \bibinfo{year}{2023}\natexlab{a}.
\newblock \showarticletitle{Split Cycle: {A} New {C}ondorcet-Consistent Voting
  Method Independent of Clones and Immune to Spoilers}.
\newblock \bibinfo{journal}{\emph{Public Choice}} \bibinfo{volume}{197},
  \bibinfo{number}{1} (\bibinfo{year}{2023}), \bibinfo{pages}{1--62}.
\newblock


\bibitem[Holliday and Pacuit(2023b)]%
        {holliday2023stable}
\bibfield{author}{\bibinfo{person}{Wesley Holliday} {and} \bibinfo{person}{Eric
  Pacuit}.} \bibinfo{year}{2023}\natexlab{b}.
\newblock \showarticletitle{Stable Voting}.
\newblock \bibinfo{journal}{\emph{Constitutional Political Economy}}
  \bibinfo{volume}{34}, \bibinfo{number}{3} (\bibinfo{year}{2023}),
  \bibinfo{pages}{421--433}.
\newblock


\bibitem[Holliday et~al\mbox{.}(2025)]%
        {holliday2024learning}
\bibfield{author}{\bibinfo{person}{Wesley~H. Holliday},
  \bibinfo{person}{Alexander Kristoffersen}, {and} \bibinfo{person}{Eric
  Pacuit}.} \bibinfo{year}{2025}\natexlab{}.
\newblock \showarticletitle{Learning to Manipulate under Limited Information}.
  In \bibinfo{booktitle}{\emph{Proceedings of the 39th AAAI Conference on
  Artificial Intelligence (AAAI)}}.
\newblock


\bibitem[Holliday and Pacuit(2025)]%
        {holliday2025pref_voting}
\bibfield{author}{\bibinfo{person}{Wesley~H Holliday} {and}
  \bibinfo{person}{Eric Pacuit}.} \bibinfo{year}{2025}\natexlab{}.
\newblock \showarticletitle{pref\_voting: {T}he Preferential Voting Tools
  package for Python}.
\newblock \bibinfo{journal}{\emph{Journal of Open Source Software}}
  \bibinfo{volume}{10}, \bibinfo{number}{105} (\bibinfo{year}{2025}),
  \bibinfo{pages}{7020}.
\newblock


\bibitem[Hornik et~al\mbox{.}(1989)]%
        {Hornik1989}
\bibfield{author}{\bibinfo{person}{Kurt Hornik}, \bibinfo{person}{Maxwell
  Stinchcombe}, {and} \bibinfo{person}{Halbert White}.}
  \bibinfo{year}{1989}\natexlab{}.
\newblock \showarticletitle{Multilayer Feedforward Networks are Universal
  Approximators}.
\newblock \bibinfo{journal}{\emph{Neural Networks}} \bibinfo{volume}{2},
  \bibinfo{number}{5} (\bibinfo{year}{1989}), \bibinfo{pages}{359--366}.
\newblock


\bibitem[Koster et~al\mbox{.}(2022)]%
        {koster2022human}
\bibfield{author}{\bibinfo{person}{Raphael Koster}, \bibinfo{person}{Jan
  Balaguer}, \bibinfo{person}{Andrea Tacchetti}, \bibinfo{person}{Ari
  Weinstein}, \bibinfo{person}{Tina Zhu}, \bibinfo{person}{Oliver Hauser},
  \bibinfo{person}{Duncan Williams}, \bibinfo{person}{Lucy
  Campbell-Gillingham}, \bibinfo{person}{Phoebe Thacker}, {and}
  \bibinfo{person}{Matthew Botvinick}.} \bibinfo{year}{2022}\natexlab{}.
\newblock \showarticletitle{Human-centred Mechanism Design with {D}emocratic
  {AI}}.
\newblock \bibinfo{journal}{\emph{Nature Human Behaviour}} \bibinfo{volume}{6},
  \bibinfo{number}{10} (\bibinfo{year}{2022}), \bibinfo{pages}{1398--1407}.
\newblock


\bibitem[Kujawska et~al\mbox{.}(2020)]%
        {kujawska2020predicting}
\bibfield{author}{\bibinfo{person}{Hanna Kujawska}, \bibinfo{person}{Marija
  Slavkovik}, {and} \bibinfo{person}{Jan-Joachim R{\"u}ckmann}.}
  \bibinfo{year}{2020}\natexlab{}.
\newblock \showarticletitle{Predicting the Winners of {B}orda, {K}emeny, and
  {D}odgson Elections with Supervised Machine Learning}. In
  \bibinfo{booktitle}{\emph{EUMAS Multi-Agent Systems and Agreement
  Technologies Workshop}}. \bibinfo{pages}{440--458}.
\newblock


\bibitem[Laslier(2011)]%
        {laslier2011and}
\bibfield{author}{\bibinfo{person}{Jean-Fran{\c{c}}ois Laslier}.}
  \bibinfo{year}{2011}\natexlab{}.
\newblock \showarticletitle{And the Loser is… Plurality Voting}.
\newblock \bibinfo{journal}{\emph{Electoral Systems}} (\bibinfo{year}{2011}),
  \bibinfo{pages}{327--351}.
\newblock


\bibitem[Lee et~al\mbox{.}(2014)]%
        {lee2014crowdsourcing}
\bibfield{author}{\bibinfo{person}{David Lee}, \bibinfo{person}{Ashish Goel},
  \bibinfo{person}{Tanja Aitamurto}, {and} \bibinfo{person}{Helene Landemore}.}
  \bibinfo{year}{2014}\natexlab{}.
\newblock \showarticletitle{Crowdsourcing for Participatory Democracies:
  {E}fficient Elicitation of Social Choice Functions}. In
  \bibinfo{booktitle}{\emph{Proceedings of the 2nd AAAI Conference on Human
  Computation and Crowdsourcing}}.
\newblock


\bibitem[List(2011)]%
        {List2011}
\bibfield{author}{\bibinfo{person}{Christian List}.}
  \bibinfo{year}{2011}\natexlab{}.
\newblock \showarticletitle{The Logical Space of Democracy}.
\newblock \bibinfo{journal}{\emph{Philosophy \& Public Affairs}}
  \bibinfo{volume}{39}, \bibinfo{number}{3} (\bibinfo{year}{2011}),
  \bibinfo{pages}{262--297}.
\newblock


\bibitem[List(2022)]%
        {List2022}
\bibfield{author}{\bibinfo{person}{Christian List}.}
  \bibinfo{year}{2022}\natexlab{}.
\newblock \showarticletitle{{Social Choice Theory}}.
\newblock In \bibinfo{booktitle}{\emph{The {Stanford} Encyclopedia of
  Philosophy} (\bibinfo{edition}{{W}inter 2022} ed.)},
  \bibfield{editor}{\bibinfo{person}{Edward~N. Zalta} {and}
  \bibinfo{person}{Uri Nodelman}} (Eds.). \bibinfo{publisher}{Metaphysics
  Research Lab, Stanford University}.
\newblock


\bibitem[Loshchilov and Hutter(2017)]%
        {loshchilov2017}
\bibfield{author}{\bibinfo{person}{Ilya Loshchilov} {and}
  \bibinfo{person}{Frank Hutter}.} \bibinfo{year}{2017}\natexlab{}.
\newblock \showarticletitle{{SGDR}: {S}tochastic Gradient Descent with Warm
  Restarts}. In \bibinfo{booktitle}{\emph{International Conference on Learning
  Representations (ICLR)}}.
\newblock


\bibitem[Loshchilov and Hutter(2019)]%
        {Loshchilov2019}
\bibfield{author}{\bibinfo{person}{Ilya Loshchilov} {and}
  \bibinfo{person}{Frank Hutter}.} \bibinfo{year}{2019}\natexlab{}.
\newblock \showarticletitle{Decoupled Weight Decay Regularization}. In
  \bibinfo{booktitle}{\emph{International Conference on Learning
  Representations (ICLR)}}.
\newblock


\bibitem[Mallows(1957)]%
        {mallows1957non}
\bibfield{author}{\bibinfo{person}{Colin~L Mallows}.}
  \bibinfo{year}{1957}\natexlab{}.
\newblock \showarticletitle{Non-Null Ranking Models}.
\newblock \bibinfo{journal}{\emph{Biometrika}} \bibinfo{volume}{44},
  \bibinfo{number}{1/2} (\bibinfo{year}{1957}), \bibinfo{pages}{114--130}.
\newblock


\bibitem[Merrill(1984)]%
        {merrill1984comparison}
\bibfield{author}{\bibinfo{person}{Samuel Merrill}.}
  \bibinfo{year}{1984}\natexlab{}.
\newblock \showarticletitle{A Comparison of Efficiency of Multicandidate
  Electoral Systems}.
\newblock \bibinfo{journal}{\emph{American Journal of Political Science}}
  (\bibinfo{year}{1984}), \bibinfo{pages}{23--48}.
\newblock


\bibitem[Mikolov et~al\mbox{.}(2013)]%
        {Mikolov2013}
\bibfield{author}{\bibinfo{person}{Tomas Mikolov}, \bibinfo{person}{Ilya
  Sutskever}, \bibinfo{person}{Kai Chen}, \bibinfo{person}{Greg~S Corrado},
  {and} \bibinfo{person}{Jeff Dean}.} \bibinfo{year}{2013}\natexlab{}.
\newblock \showarticletitle{Distributed Representations of Words and Phrases
  and their Compositionality}.
\newblock \bibinfo{journal}{\emph{Advances in Neural Information Processing
  Systems}}  \bibinfo{volume}{26} (\bibinfo{year}{2013}).
\newblock


\bibitem[Mohsin et~al\mbox{.}(2022)]%
        {mohsin2022learning}
\bibfield{author}{\bibinfo{person}{Farhad Mohsin}, \bibinfo{person}{Ao Liu},
  \bibinfo{person}{Pin-Yu Chen}, \bibinfo{person}{Francesca Rossi}, {and}
  \bibinfo{person}{Lirong Xia}.} \bibinfo{year}{2022}\natexlab{}.
\newblock \showarticletitle{Learning to Design Fair and Private Voting Rules}.
\newblock \bibinfo{journal}{\emph{Journal of Artificial Intelligence Research}}
   \bibinfo{volume}{75} (\bibinfo{year}{2022}), \bibinfo{pages}{1139--1176}.
\newblock


\bibitem[Nardi et~al\mbox{.}(2022)]%
        {NardiEtAlAAMAS2022}
\bibfield{author}{\bibinfo{person}{Oliviero Nardi}, \bibinfo{person}{Arthur
  Boixel}, {and} \bibinfo{person}{Ulle Endriss}.}
  \bibinfo{year}{2022}\natexlab{}.
\newblock \showarticletitle{A Graph-Based Algorithm for the Automated
  Justification of Collective Decisions}. In
  \bibinfo{booktitle}{\emph{Proceedings of the 21st International Conference on
  Autonomous Agents and Multiagent Systems (AAMAS)}}.
\newblock


\bibitem[Nitzan(1985)]%
        {nitzan1985vulnerability}
\bibfield{author}{\bibinfo{person}{Shmuel Nitzan}.}
  \bibinfo{year}{1985}\natexlab{}.
\newblock \showarticletitle{The Vulnerability of Point-Voting Schemes to
  Preference Variation and Strategic Manipulation}.
\newblock \bibinfo{journal}{\emph{Public choice}}  \bibinfo{volume}{47}
  (\bibinfo{year}{1985}), \bibinfo{pages}{349--370}.
\newblock


\bibitem[Noothigattu et~al\mbox{.}(2018)]%
        {noothigattu2018voting}
\bibfield{author}{\bibinfo{person}{Ritesh Noothigattu},
  \bibinfo{person}{Snehalkumar Gaikwad}, \bibinfo{person}{Edmond Awad},
  \bibinfo{person}{Sohan Dsouza}, \bibinfo{person}{Iyad Rahwan},
  \bibinfo{person}{Pradeep Ravikumar}, {and} \bibinfo{person}{Ariel
  Procaccia}.} \bibinfo{year}{2018}\natexlab{}.
\newblock \showarticletitle{A Voting-Based System for Ethical Decision Making}.
  In \bibinfo{booktitle}{\emph{Proceedings of the 32nd AAAI Conference on
  Artificial Intelligence (AAAI)}}.
\newblock


\bibitem[Nurmi(1988)]%
        {nurmi1988discrepancies}
\bibfield{author}{\bibinfo{person}{Hannu Nurmi}.}
  \bibinfo{year}{1988}\natexlab{}.
\newblock \showarticletitle{Discrepancies in the Outcomes Resulting from
  Different Voting Schemes}.
\newblock \bibinfo{journal}{\emph{Theory and Decision}}  \bibinfo{volume}{25}
  (\bibinfo{year}{1988}), \bibinfo{pages}{193--208}.
\newblock


\bibitem[Perdomo et~al\mbox{.}(2020)]%
        {Perdomo2020}
\bibfield{author}{\bibinfo{person}{Juan Perdomo}, \bibinfo{person}{Tijana
  Zrnic}, \bibinfo{person}{Celestine Mendler-D{\"u}nner}, {and}
  \bibinfo{person}{Moritz Hardt}.} \bibinfo{year}{2020}\natexlab{}.
\newblock \showarticletitle{Performative Prediction}. In
  \bibinfo{booktitle}{\emph{Proceedings of the 37th International Conference on
  Machine Learning (ICML)}}.
\newblock


\bibitem[Powers(2007)]%
        {powers2007number}
\bibfield{author}{\bibinfo{person}{Robert~C Powers}.}
  \bibinfo{year}{2007}\natexlab{}.
\newblock \showarticletitle{The Number of Times an Anonymous Rule Violates
  Independence in the 3$\times$ 3 Case}.
\newblock \bibinfo{journal}{\emph{Social Choice and Welfare}}
  \bibinfo{volume}{28}, \bibinfo{number}{3} (\bibinfo{year}{2007}),
  \bibinfo{pages}{363--373}.
\newblock


\bibitem[Procaccia et~al\mbox{.}(2009)]%
        {procaccia2009learnability}
\bibfield{author}{\bibinfo{person}{Ariel~D Procaccia}, \bibinfo{person}{Aviv
  Zohar}, \bibinfo{person}{Yoni Peleg}, {and} \bibinfo{person}{Jeffrey~S
  Rosenschein}.} \bibinfo{year}{2009}\natexlab{}.
\newblock \showarticletitle{The Learnability of Voting Rules}.
\newblock \bibinfo{journal}{\emph{Artificial Intelligence}}
  \bibinfo{volume}{173}, \bibinfo{number}{12-13} (\bibinfo{year}{2009}),
  \bibinfo{pages}{1133--1149}.
\newblock


\bibitem[Rafailov et~al\mbox{.}(2024)]%
        {Rafailov2024}
\bibfield{author}{\bibinfo{person}{Rafael Rafailov}, \bibinfo{person}{Archit
  Sharma}, \bibinfo{person}{Eric Mitchell}, \bibinfo{person}{Christopher~D
  Manning}, \bibinfo{person}{Stefano Ermon}, {and} \bibinfo{person}{Chelsea
  Finn}.} \bibinfo{year}{2024}\natexlab{}.
\newblock \showarticletitle{Direct Preference Optimization: {Y}our Language
  Model is Secretly a Reward Model}.
\newblock \bibinfo{journal}{\emph{Advances in Neural Information Processing
  Systems}}  \bibinfo{volume}{36} (\bibinfo{year}{2024}).
\newblock


\bibitem[Terzopoulou(2023)]%
        {terzopoulou2023}
\bibfield{author}{\bibinfo{person}{Zoi Terzopoulou}.}
  \bibinfo{year}{2023}\natexlab{}.
\newblock \showarticletitle{Voting with Limited Energy: {A} Study of
  {P}lurality and {B}orda}. In \bibinfo{booktitle}{\emph{Proceedings of the
  22nd International Conference on Autonomous Agents and Multiagent Systems
  (AAMAS)}}.
\newblock


\bibitem[Thomson(2001)]%
        {Thomson2001}
\bibfield{author}{\bibinfo{person}{William Thomson}.}
  \bibinfo{year}{2001}\natexlab{}.
\newblock \showarticletitle{On the Axiomatic Method and its Recent Applications
  to Game Theory and Resource Allocation}.
\newblock \bibinfo{journal}{\emph{Social Choice and Welfare}}
  \bibinfo{volume}{18}, \bibinfo{number}{2} (\bibinfo{year}{2001}),
  \bibinfo{pages}{327--386}.
\newblock


\bibitem[Xia(2013)]%
        {xia2013designing}
\bibfield{author}{\bibinfo{person}{Lirong Xia}.}
  \bibinfo{year}{2013}\natexlab{}.
\newblock \showarticletitle{Designing Social Choice Mechanisms Using Machine
  Learning}. In \bibinfo{booktitle}{\emph{Proceedings of the 12th International
  Conference on Autonomous Agents and Multiagent Systems (AAMAS)}}.
\newblock


\bibitem[Xu et~al\mbox{.}(2018)]%
        {Xu2018}
\bibfield{author}{\bibinfo{person}{Jingyi Xu}, \bibinfo{person}{Zilu Zhang},
  \bibinfo{person}{Tal Friedman}, \bibinfo{person}{Yitao Liang}, {and}
  \bibinfo{person}{Guy Broeck}.} \bibinfo{year}{2018}\natexlab{}.
\newblock \showarticletitle{A Semantic Loss Function for Deep Learning with
  Symbolic Knowledge}. In \bibinfo{booktitle}{\emph{International conference on
  machine learning (ICML)}}.
\newblock


\bibitem[Yang et~al\mbox{.}(2024)]%
        {yang2024llm}
\bibfield{author}{\bibinfo{person}{Joshua Yang}, \bibinfo{person}{Damian
  Dailisan}, \bibinfo{person}{Marcin Korecki}, \bibinfo{person}{Carina
  Hausladen}, {and} \bibinfo{person}{Dirk Helbing}.}
  \bibinfo{year}{2024}\natexlab{}.
\newblock \showarticletitle{LLM Voting: {H}uman Choices and {AI} Collective
  Decision-Making}. In \bibinfo{booktitle}{\emph{Proceedings of the AAAI/ACM
  Conference on AI, Ethics, and Society}}.
\newblock


\bibitem[Zwicker(2016)]%
        {zwicker2016introduction}
\bibfield{author}{\bibinfo{person}{William~S Zwicker}.}
  \bibinfo{year}{2016}\natexlab{}.
\newblock \showarticletitle{Introduction to the Theory of Voting}.
\newblock In \bibinfo{booktitle}{\emph{Handbook of Computational Social
  Choice}}, \bibfield{editor}{\bibinfo{person}{Felix Brandt},
  \bibinfo{person}{Vincent Conitzer}, \bibinfo{person}{Ulle Endriss},
  \bibinfo{person}{J{\'e}r{\^o}me Lang}, {and} \bibinfo{person}{Ariel~D.
  Procaccia}} (Eds.). \bibinfo{publisher}{Cambridge University Press}.
\newblock


\end{thebibliography}

\newpage
\appendix

\section{Hyperparameter Tuning}
\label{sec: app_hyperparamter}

In this section, we investigate different hyperparameter choices for our models to motivate the choices we make in the main text.

\subsection{Model Sizes}

\begin{figure*}[b]
\centering
\includegraphics[width=0.49\linewidth]{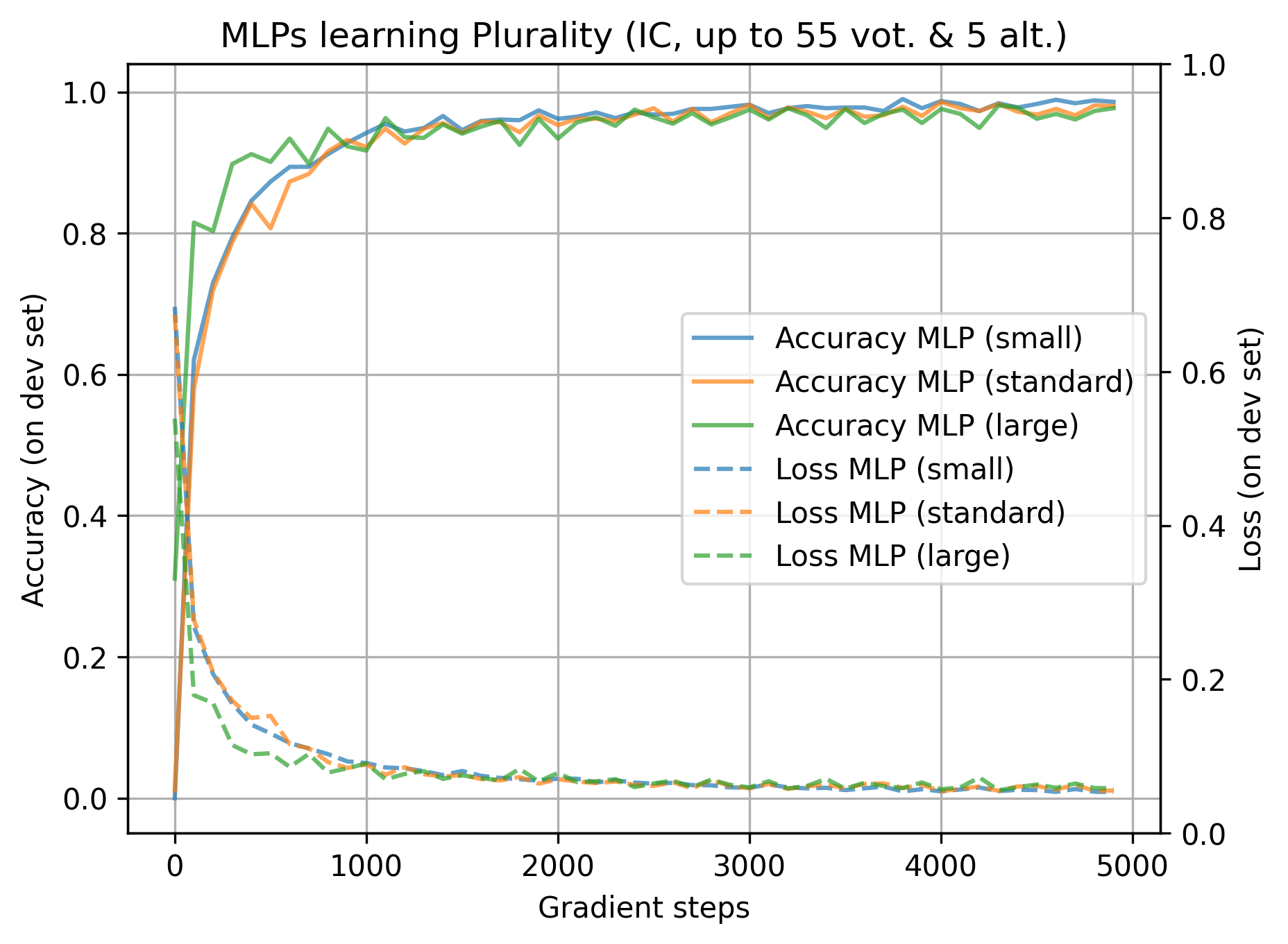}
\includegraphics[width=0.49\linewidth]{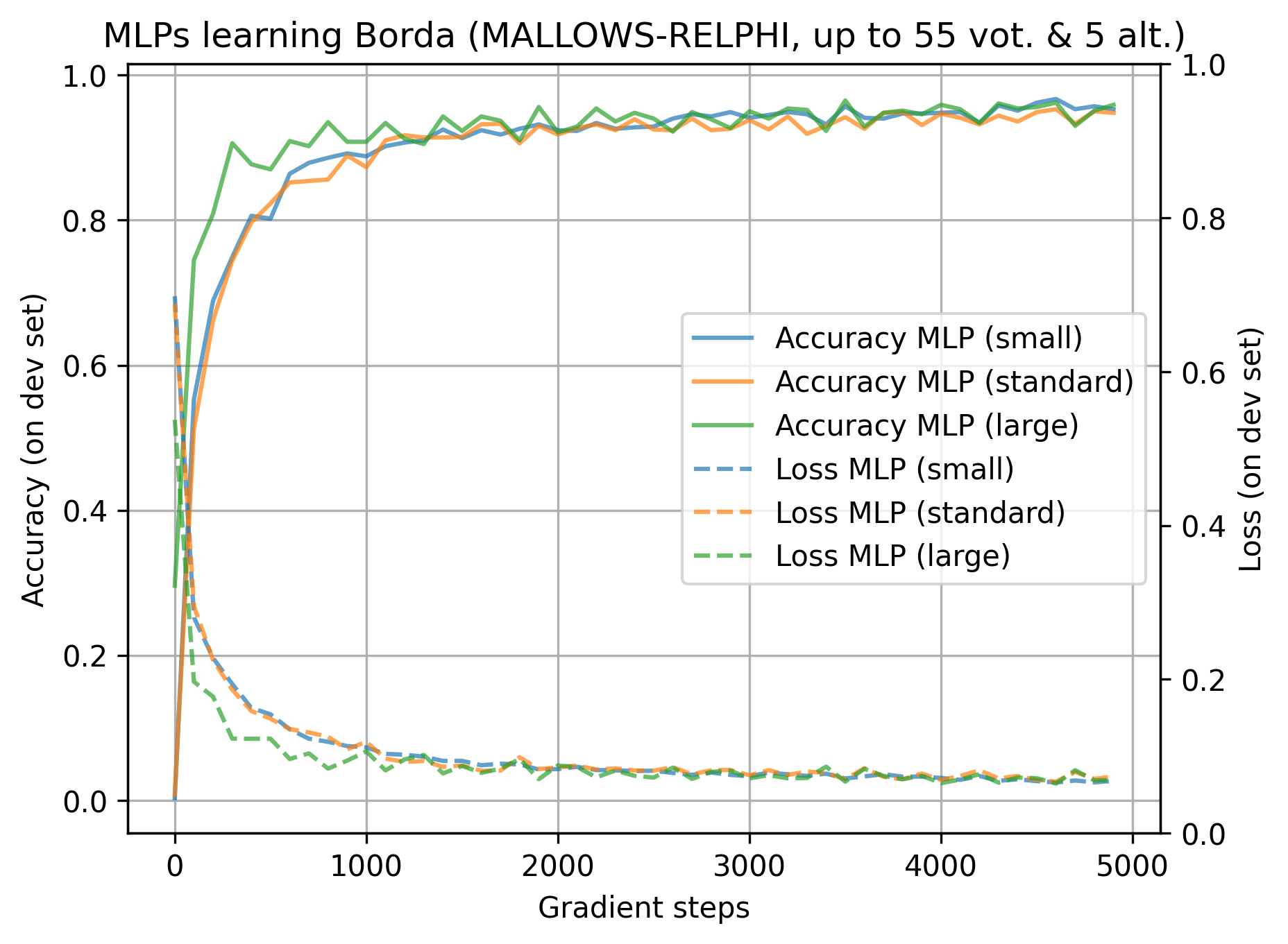}
\Description{Two plots describing the learning performance of different sized MLPs.}
\caption{Different sized MLPs and their learning performance.
The `small' MLP has two hidden layers with 128 neurons each,
the `standard' MLP has four hidden layers with 128 neurons each (our and the literature's choice), and
the `large' MLP has six hidden layers with 128 neurons each plus layer norm.}
\label{fig: exp1 appendix grid search MLP}
\end{figure*}

First, regarding the MLPs, Figure~\ref{fig: exp1 appendix grid search MLP} tests the performance of different sizes. As mentioned in Section~\ref{ssec: hyperparamters}, we use the same sizes for our MLPs as~\citet{anil2021learning}: four hidden layers with 128 neurons each. To test this choice, we compare it to a smaller MLP with only two hidden layers with 128 neurons each, and to a larger MLP with six hidden layers with 128 neurons each plus layer norm. 
Figure~\ref{fig: exp1 appendix grid search MLP} shows that all three MLPs achieve very similar accuracy. The larger MLP learns a bit more quickly than the other two, but it also has a higher variance in achieved accuracy. Hence the larger MLP does not yield a performance improvement. The results suggest that a smaller MLP might work, too, but, for continuity with the literature on the topic, when then choose their model sizes. 

Second, regarding the CNNs and the WECs, the choice for the MLP size also dictates their sizes: in order to have a comparable capacity, they should have a roughly similar number of parameters. 
Indeed, with our choices of numbers and kinds of layers for the CNN and the WEC, we get, as described in Section~\ref{ssec: hyperparamters}, models that are roughly comparable in size.

\subsection{CNN Kernels and Pre-processing}
\label{ssec: app CNN kernel size}

\begin{figure*}
\begin{footnotesize}
\centering
\begin{minipage}{0.45\linewidth}
\centering
[$(5,1)$, $(1,5)$, $32$, no KT]
\includegraphics[width=\linewidth]{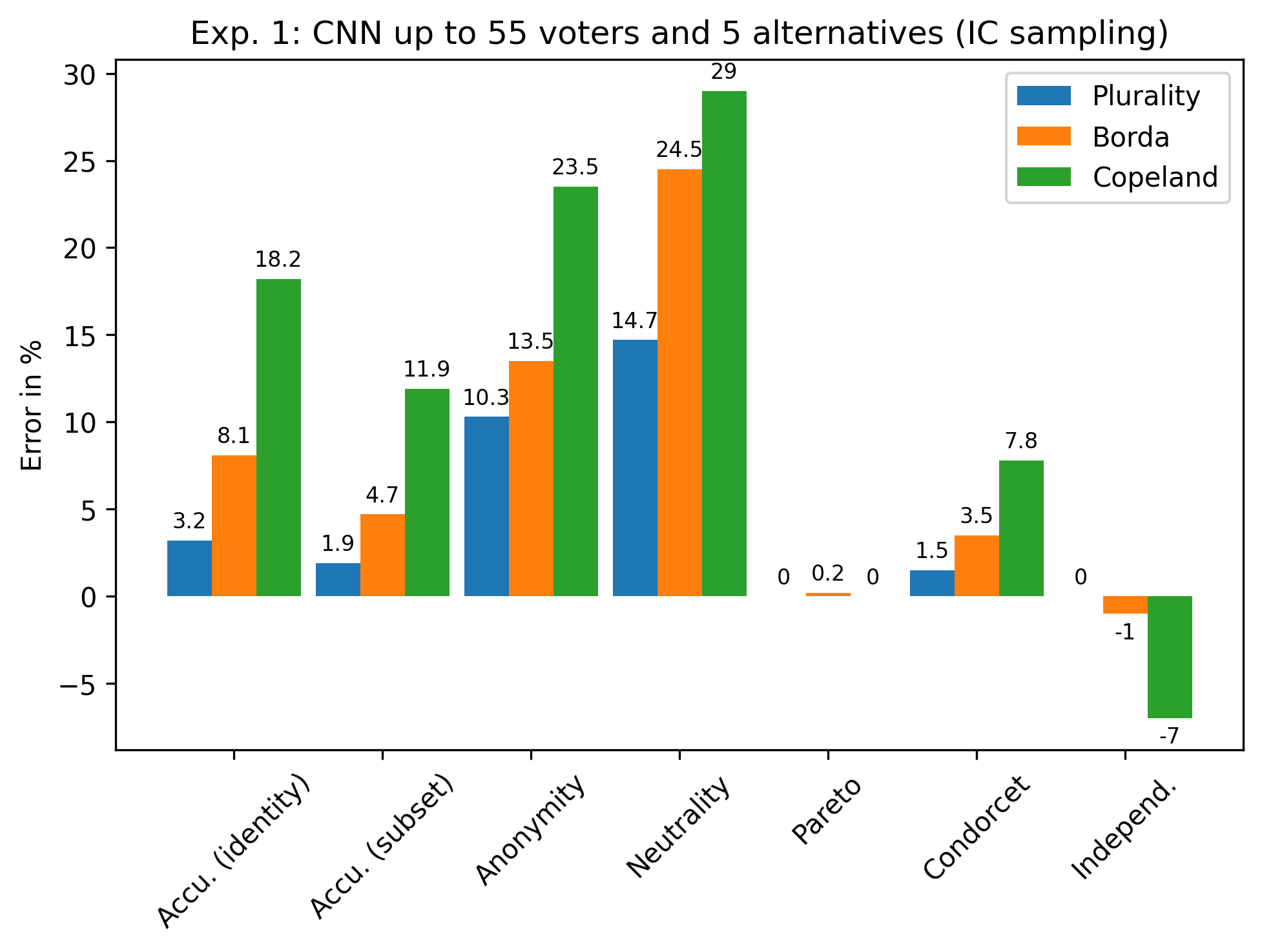}
\end{minipage}
\begin{minipage}{0.45\linewidth}
\centering
[$(3,3)$, $(3,3)$, $32$, no KT]
\includegraphics[width=\linewidth]{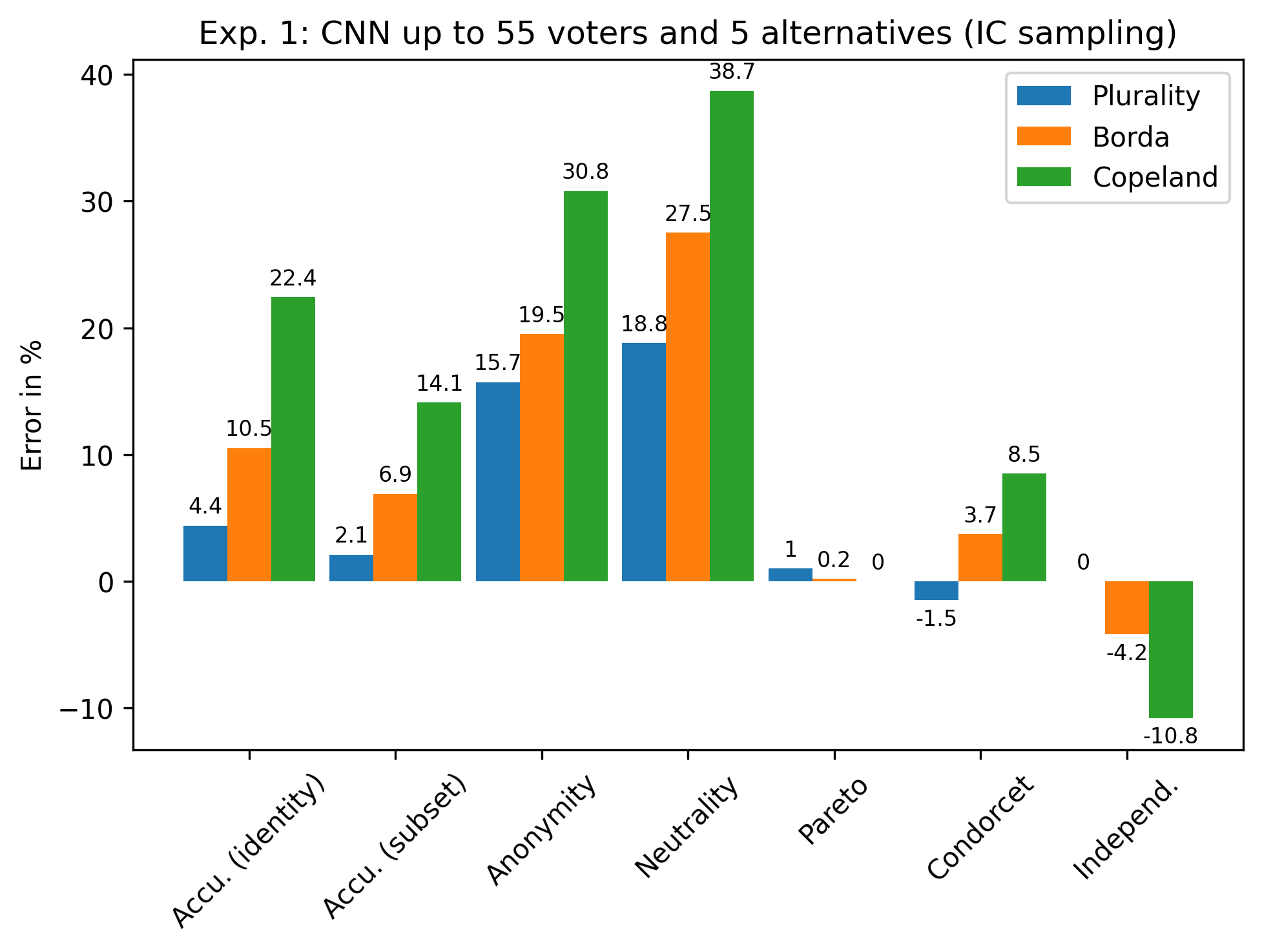}
\end{minipage}
\\
\begin{minipage}{0.45\linewidth}
\centering
[$(5,1)$, $(1,5)$, $32$, KT global]
\includegraphics[width=\linewidth]{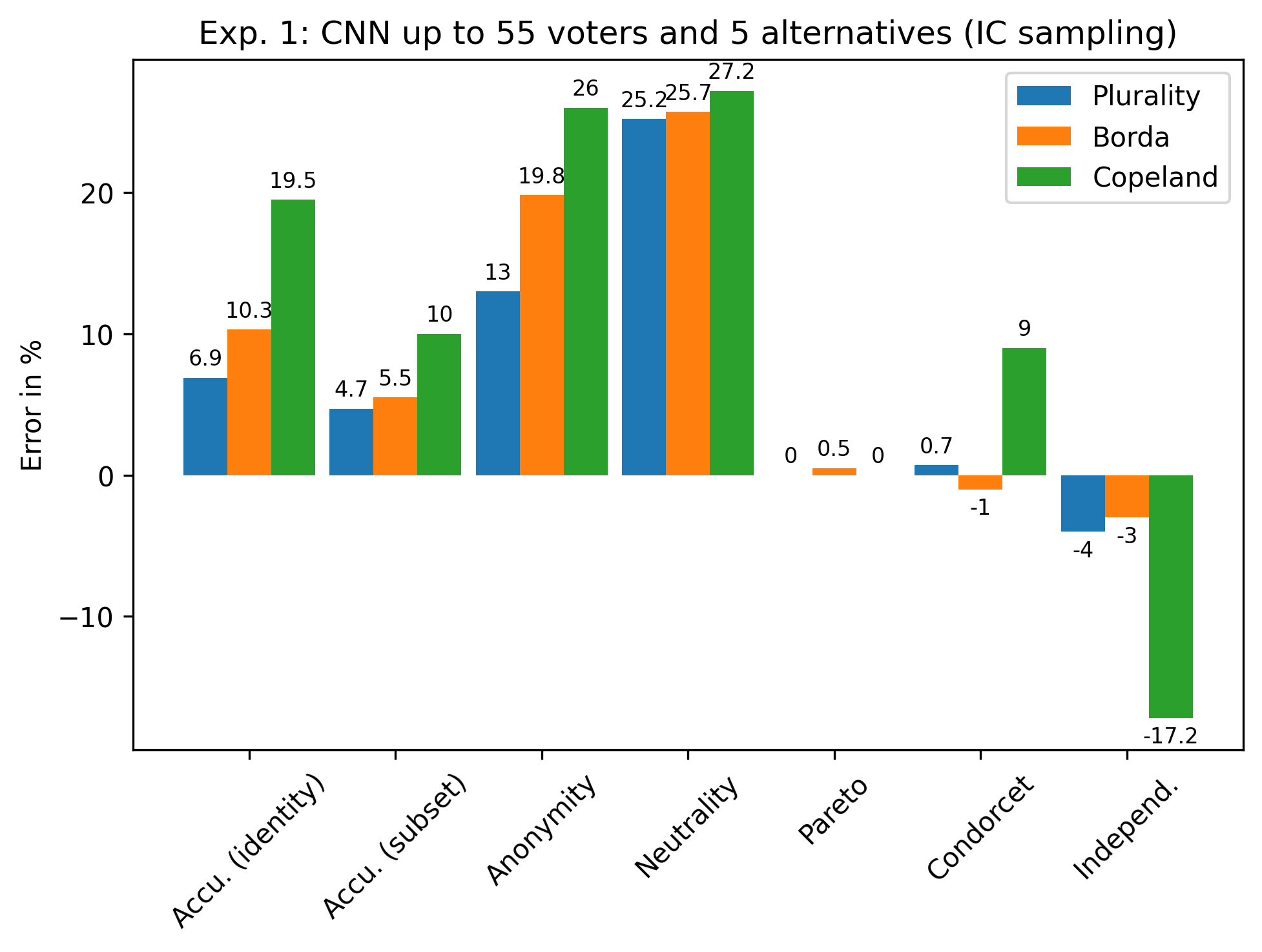}
\end{minipage}
\begin{minipage}{0.45\linewidth}
\centering
[$(3,3)$, $(3,3)$, $32$, KT global]
\includegraphics[width=\linewidth]{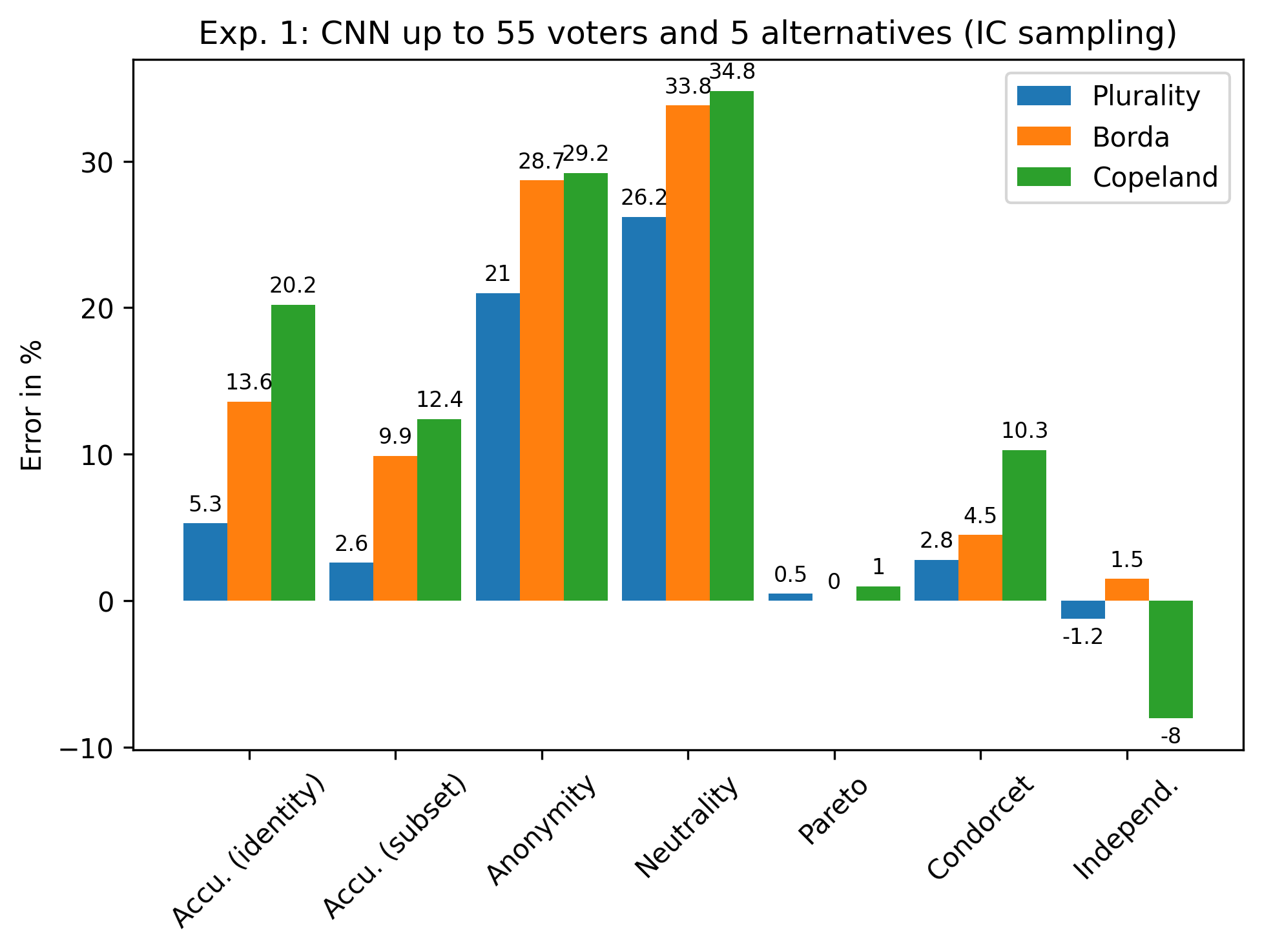}
\end{minipage}
\\
\begin{minipage}{0.45\linewidth}
\centering
[$(5,1)$, $(1,5)$, $32$, KT local]
\includegraphics[width=\linewidth]{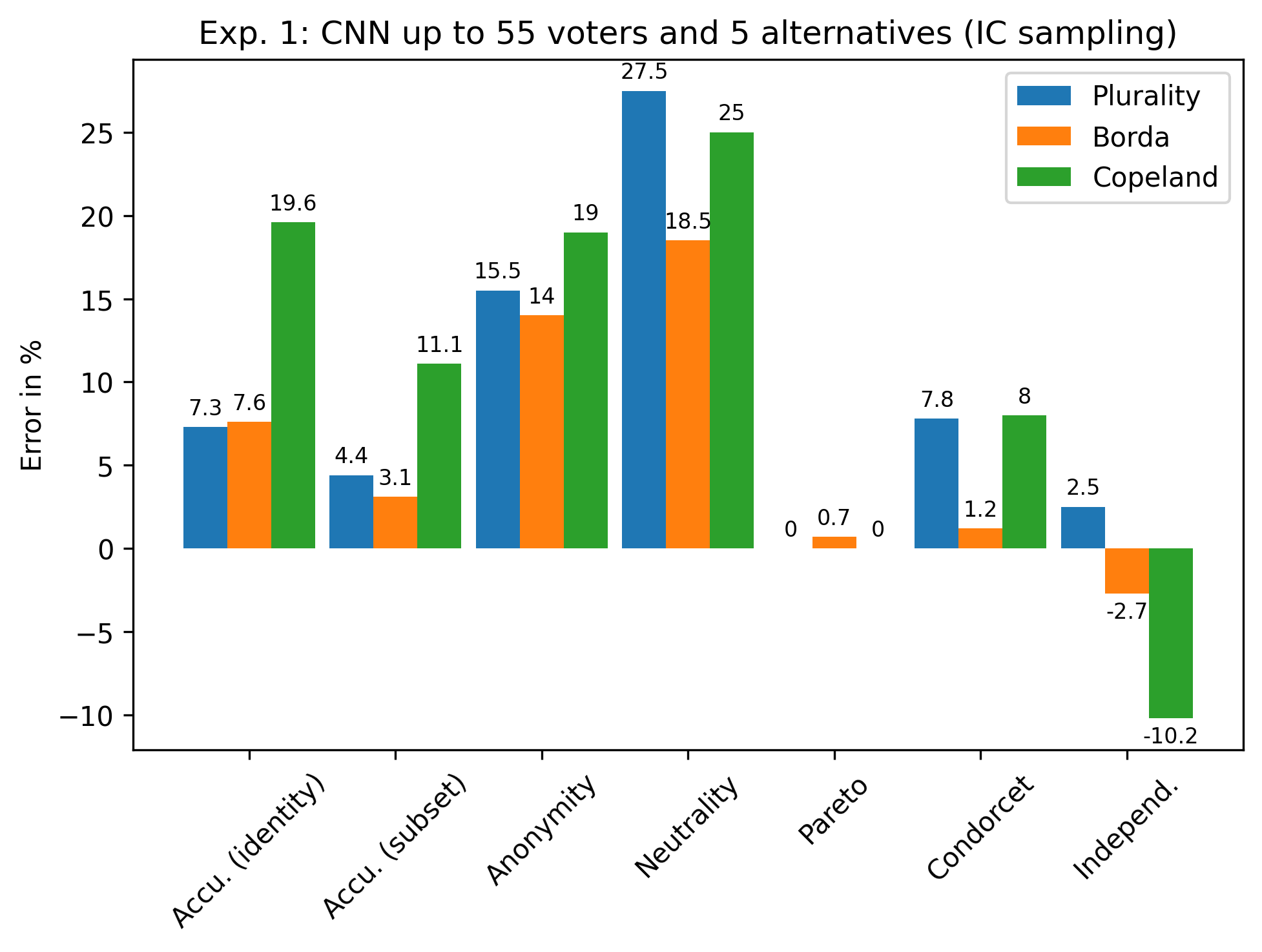}
\end{minipage}
\begin{minipage}{0.45\linewidth}
\centering
[$(3,3)$, $(3,3)$, $32$, KT local]
\includegraphics[width=\linewidth]{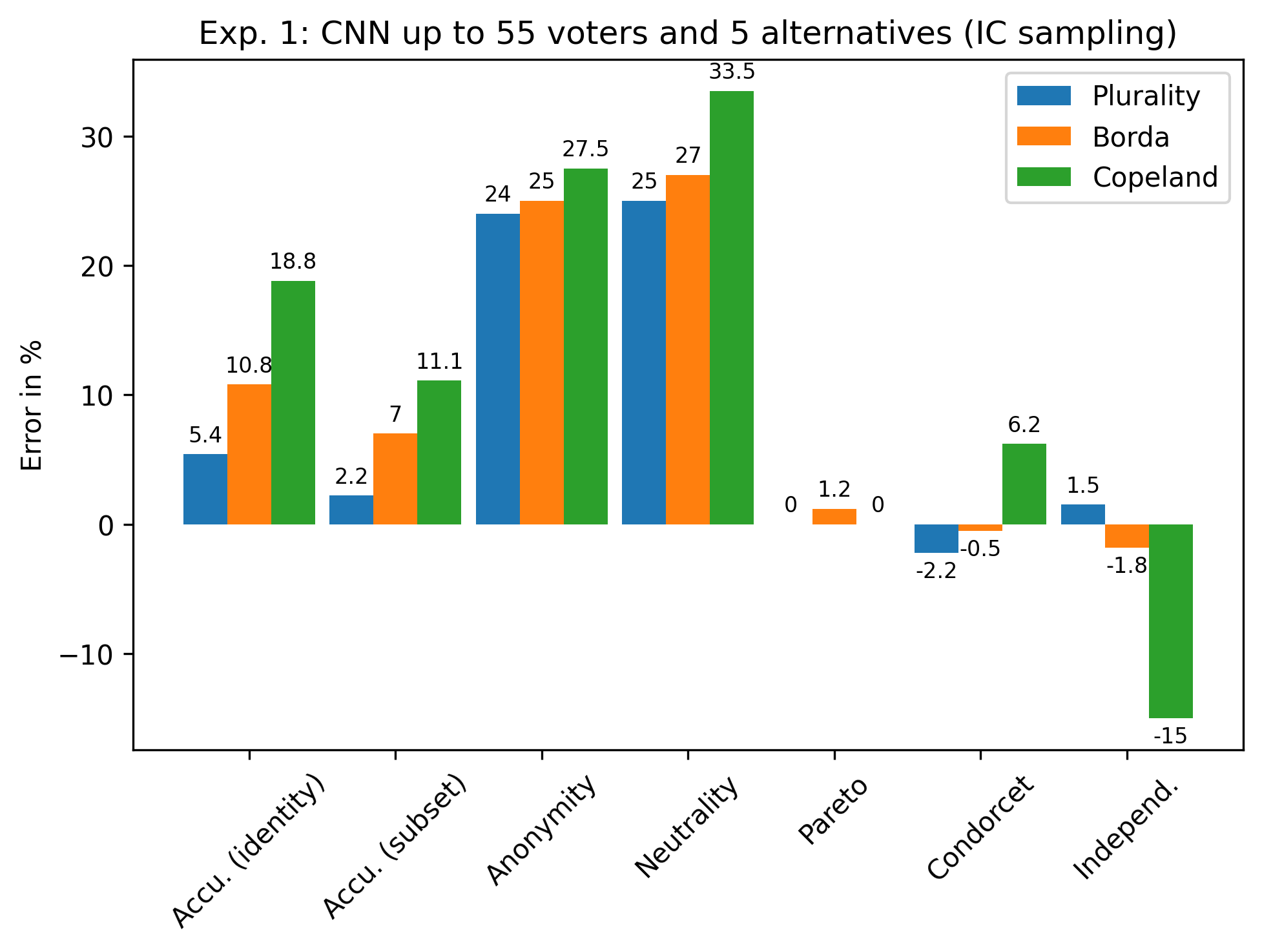}
\end{minipage}
\end{footnotesize}
\Description{Six barplots describing the performance of CNNs with different choices of hyperparameters.}
\caption{Different hyperparameters of CNNs and their performance. The description [$(5,1)$, $(1,5)$, $32$, no KT] means that, for this CNN, the first kernel has size $(5,1)$, the second kernel has size $(1,5)$, the number of channels is 32, and the input is not Kendall Tau preprocessed. Similarly for the other descriptions.
(The top left plot is our standard CNN setting and repeated from Figure~\ref{fig: exp1 appendix part 1} for convenience.) 
}
\label{fig: exp1 appendix different CNNs}
\end{figure*}

Figure~\ref{fig: exp1 appendix different CNNs} investigates different choices for the hyperparameters of the CNN architecture. In Section~\ref{ssec: hyperparamters}, we described our choice: the two convolution layers have kernel sizes $(5,1)$ and $(1,5)$, respectively. Thus, the first kernel can pick up local patterns in the rankings of the voters, while the second kernel can pick up local patterns among the $i$-th preferred alternatives of the voters. 

In image processing, `quadratic' kernel sizes---e.g., $(3,3)$---are more common, to pick up correlations of pixels with their surrounding pixels. In the voting setting, at least conceptually speaking, a quadratic surrounding does not make too much sense: Why should there be important `diagonal' correlations, say between the $i$-th preferred alternative of voter $k$ and the $i+2$-th preferred alternative of voter $k + 3$, especially if the voters should be permutable? On the contrary, vertical and horizontal correlations are important: Vertically, the first kernel captures patterns of correlation between a given alternative in a voter's ranking and more or less preferred alternatives in that ranking; horizontally, the second kernel captures patterns of correlation between the $i$-th preferred alternative of a voter and the $i$-th preferred alternative of other voters. 

Figure~\ref{fig: exp1 appendix different CNNs} shows that our choice of kernel size (top left) indeed achieves overall better results than a quadratic choice of kernel size (top right). Only for Condorcet and Independence, the quadratic choice is slightly better for some rules. 

One might wonder, if one could still leverage diagonal correlations by first reordering the rankings of the voters in a profile so that `similar' rankings are next to each other, before feeding the profile into the CNN. With such a pre-processing of the input, a quadratic kernel could be used since diagonal comparisons now make sense in such a similarity reordered profile.
A standard way to formalize this notion of similarity is via Kendall Tau distance (as defined in footnote~\ref{ftn: definition Kendall Tau distance}). 
We consider two versions of reordering a given profile $\boldsymbol{P} = (P_1, \ldots, P_n)$:
\begin{itemize}
\item
\emph{Global}:
Compute, for $k = 2, \ldots, n$, the Kendall Tau distance $d_k$ between $P_1$ and $P_k$. Then reorder the profile starting with $P_1$ followed by the other rankings with ascending $d_k$. (In case of a tie, pick the ranking with minimal index first.)

\item
\emph{Local}:
The reordered profile $\boldsymbol{P}' = (P_1', \ldots , P_n')$ is computed recursively. Start with $P_1' := P_1$. Given $P_k'$, we determine $P_{k+1}'$ as follows. Go through the rankings that have not been picked yet (i.e., $\{P_1, \ldots , P_n \} \setminus \{P_1', \ldots , P_k' \}$) and compute their Kendall Tau distance to $P_k'$. Then $P_{k+1}'$ is the ranking among these with the smallest Kendall Tau distance. (Again, tie-break via the indices.) 
\end{itemize}
Figure~\ref{fig: exp1 appendix different CNNs} shows that adding either the global or the local version of Kendall Tau pre-processing overall does not reliably help the performance compared to our chosen setting (neither for our choice of kernel size nor for the quadratic choice). In some cases we do observe an improvement, as for example in the independence axiom satisfaction when learning the Copeland rule---however, this always comes with an additional loss, either in the accuracy or in the satisfaction of other axioms such as anonymity and neutrality.

\subsection{Cross Validation}

Finally, we corroborate our choice of hyperparameters by establishing their robust learning capabilities via cross validation in Table~\ref{tbl: exp1 appendix cross validation}.
For this, we IC-sample a fixed dataset of $100,000$ data points. We split the dataset into $10$ folds (each of size $10,000$). Looping over $k = 0, \ldots , 9$, we take fold $k$ as the test set and train the model on the data in the other $9$ folds for $8$ epochs. We record the achieved accuracy and loss (both on the training and the test set). 
Table~\ref{tbl: exp1 appendix cross validation} shows that, for all architectures with their chosen hyperparameters, we always get a high accuracy with little variance. This corroborates the robust learning capabilities of our architectures.

\begin{table*}
\begin{footnotesize}
\begin{center}
MLP

\begin{tabular}{lcccc}
\toprule
 Testing fold number   &  Train loss  &  Train accuracy (in \%)  &  Test loss  &  Test accuracy (in \%)  \\
\midrule
 0                     &    0.013     &          97.7           &    0.031    &          95.1          \\
 1                     &    0.013     &          97.7           &    0.033    &          94.8          \\
 2                     &    0.012     &          97.9           &    0.032    &          95.4          \\
 3                     &    0.009     &          98.5           &    0.031    &          95.4          \\
 4                     &    0.018     &          96.9           &    0.038    &          94.4          \\
 5                     &    0.018     &          96.9           &    0.04     &          94.1          \\
 6                     &    0.025     &          95.8           &    0.042    &          93.7          \\
 7                     &    0.011     &          97.9           &    0.03     &          95.1          \\
 8                     &    0.013     &          97.7           &    0.029    &          95.9          \\
 9                     &    0.008     &          98.7           &    0.025    &           96           \\
\midrule
 Avg.                  &    0.014     &          97.6           &    0.033    &           95           \\
 Std. dev.             &    0.005     &           0.8           &    0.005    &          0.7           \\
\bottomrule
\end{tabular}

\medskip

CNN

\begin{tabular}{lcccc}
\toprule
 Testing fold number   &  Train loss  &  Train accuracy (in \%)  &  Test loss  &  Test accuracy (in \%)  \\
\midrule
 0                     &    0.021     &          96.5           &    0.023    &          96.1          \\
 1                     &    0.015     &          97.2           &    0.019    &          96.8          \\
 2                     &    0.009     &          98.7           &    0.011    &          98.3          \\
 3                     &    0.012     &           98            &    0.015    &          97.3          \\
 4                     &    0.016     &          97.3           &    0.019    &          96.9          \\
 5                     &     0.01     &          98.5           &    0.011    &           98           \\
 6                     &    0.009     &          98.4           &    0.012    &           98           \\
 7                     &     0.02     &          96.3           &    0.02     &          96.4          \\
 8                     &    0.014     &          97.3           &    0.019    &          96.6          \\
 9                     &    0.024     &          95.8           &    0.029    &          95.4          \\
\midrule
 Avg.                  &    0.015     &          97.4           &    0.018    &           97           \\
 Std. dev.             &    0.005     &           0.9           &    0.005    &          0.9           \\
\bottomrule
\end{tabular}

\medskip

WEC

\begin{tabular}{lcccc}
\toprule
 Testing fold number   &  Train loss  &  Train accuracy (in \%)  &  Test loss  &  Test accuracy (in \%)  \\
\midrule
 0                     &    0.005     &          99.4           &    0.006    &          99.5          \\
 1                     &    0.005     &          99.7           &    0.005    &          99.8          \\
 2                     &    0.006     &          99.6           &    0.006    &          99.5          \\
 3                     &    0.007     &          99.3           &    0.008    &          99.1          \\
 4                     &    0.035     &          96.3           &    0.037    &          95.8          \\
 5                     &    0.004     &          99.8           &    0.004    &          99.9          \\
 6                     &    0.012     &          98.6           &    0.014    &          98.5          \\
 7                     &     0.01     &           99            &    0.011    &          98.7          \\
 8                     &     0.01     &          98.5           &    0.009    &          98.6          \\
 9                     &    0.011     &          98.4           &    0.011    &          98.3          \\
\midrule
 Avg.                  &     0.01     &          98.8           &    0.011    &          98.8          \\
 Std. dev.             &    0.009     &            1            &    0.009    &          1.1           \\
\bottomrule
\end{tabular}
\end{center}
\end{footnotesize}
\caption{Cross validation of the three architectures on a dataset with $100,000$ data points IC-sampled from the Plurality rule with up to 55 voters and 5 alternatives. Training is for 8 epochs with a batch size of 200 (hence $8 \times \frac{90,000}{200} = 3,600$ gradient steps).}
\label{tbl: exp1 appendix cross validation}
\end{table*}

\section{Experiment 1} 
\label{sec: app_exp1}

\begin{figure*}
\centering
\includegraphics[width=0.49\linewidth]{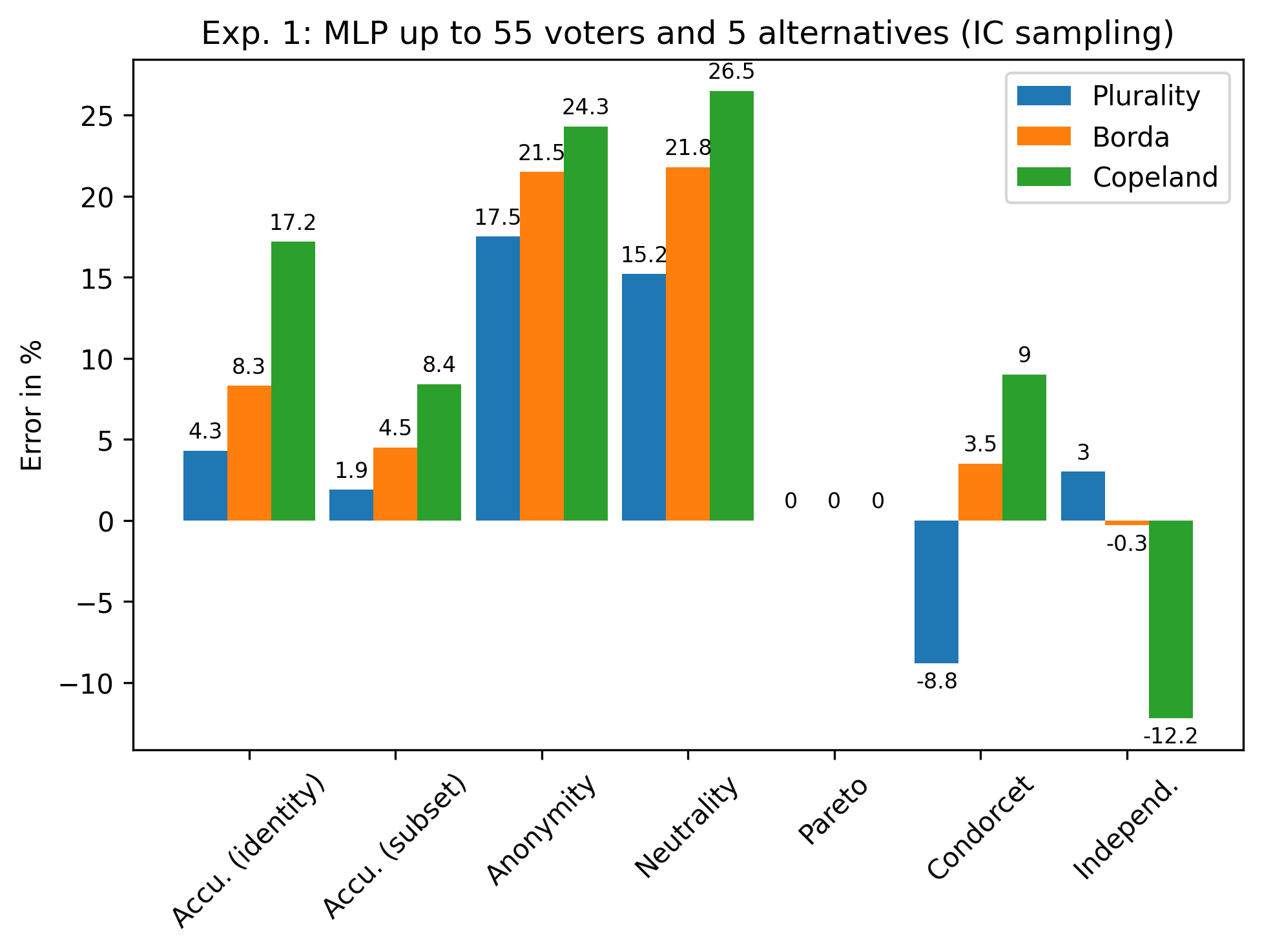}
\includegraphics[width=0.49\linewidth]{plot_2024-06-26_01-25-11_CNN_55_5_IC.png}
\\
\includegraphics[width=0.49\linewidth]{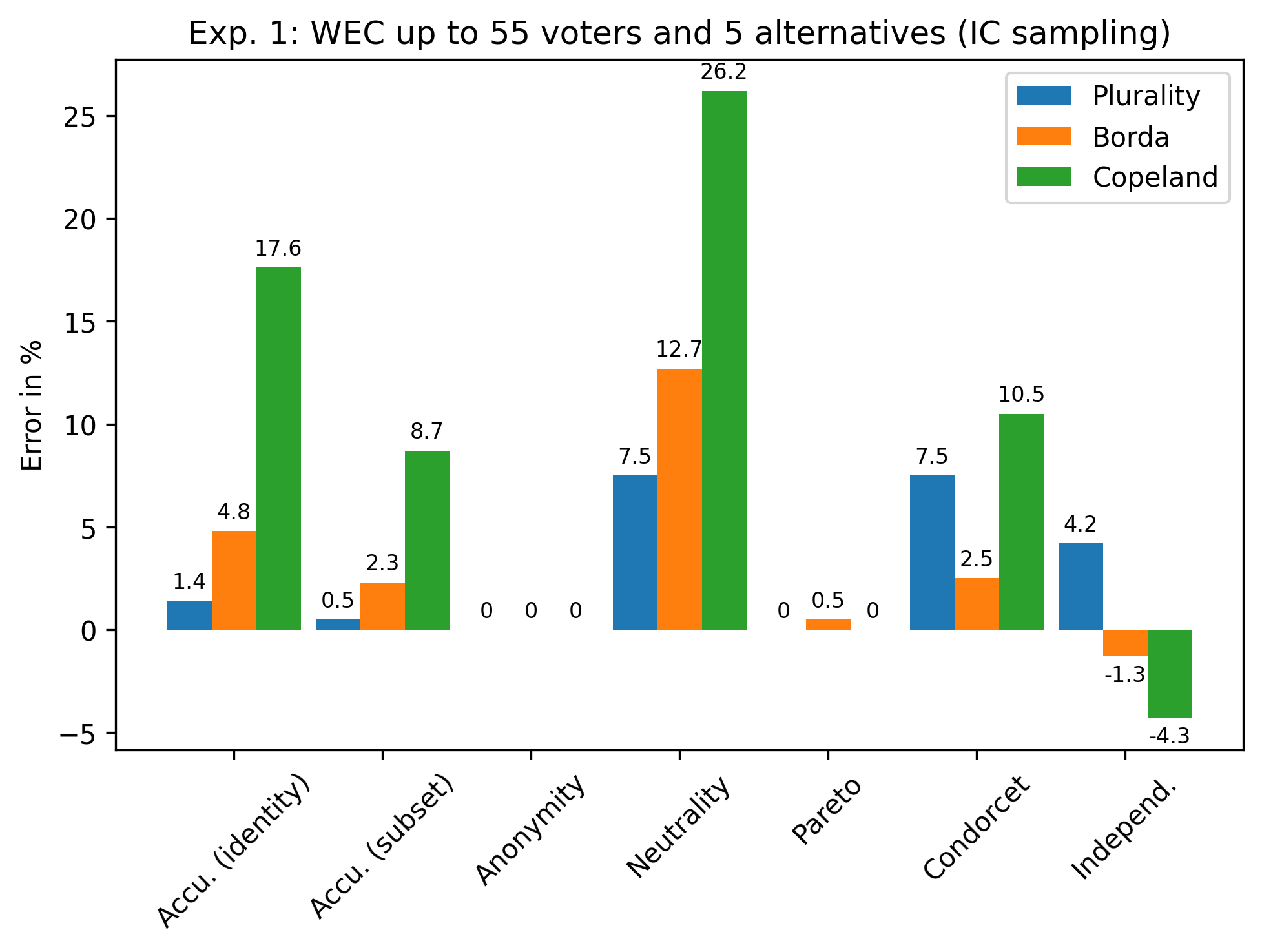}
\includegraphics[width=0.49\linewidth]{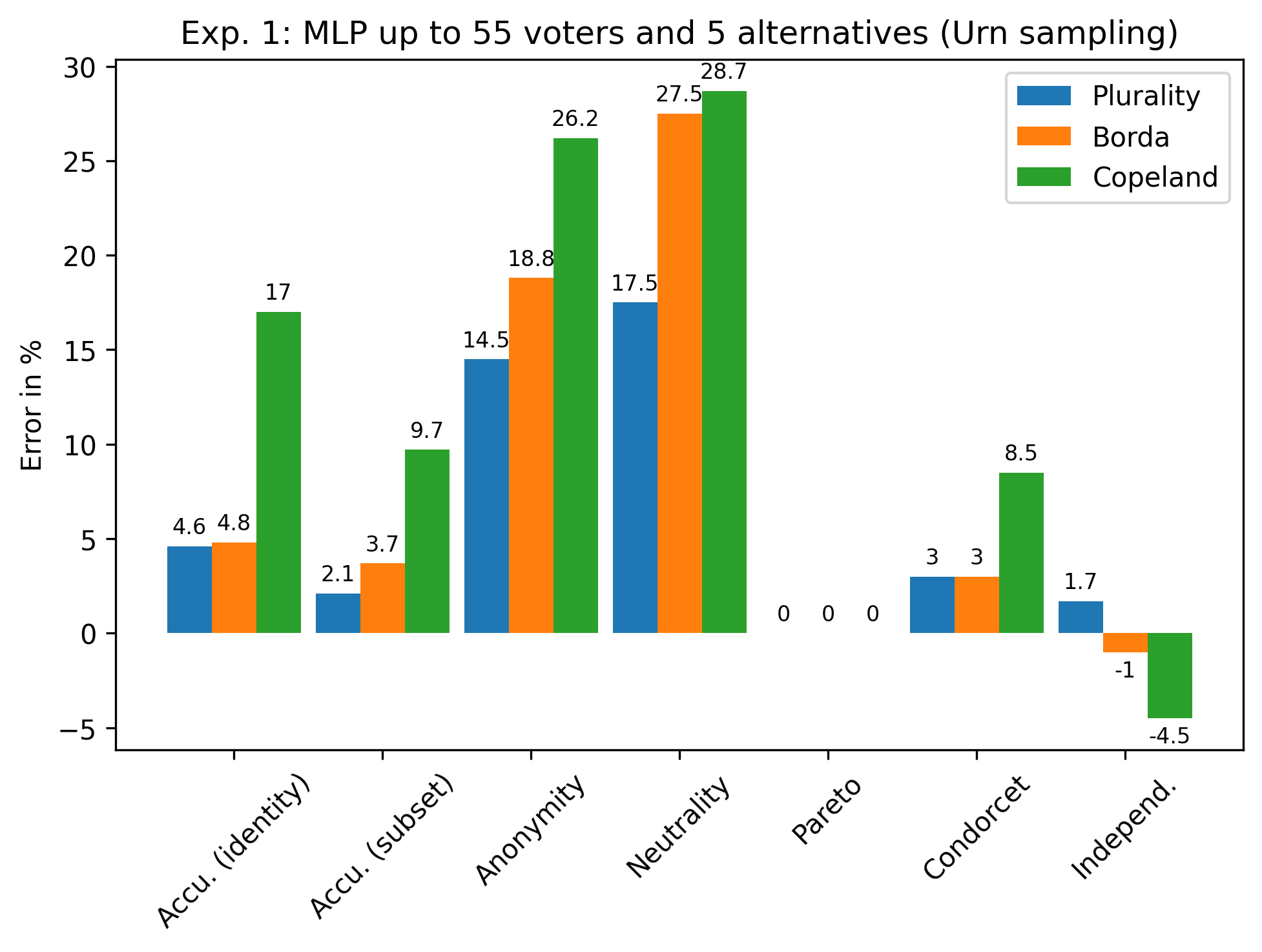}
\\
\includegraphics[width=0.49\linewidth]{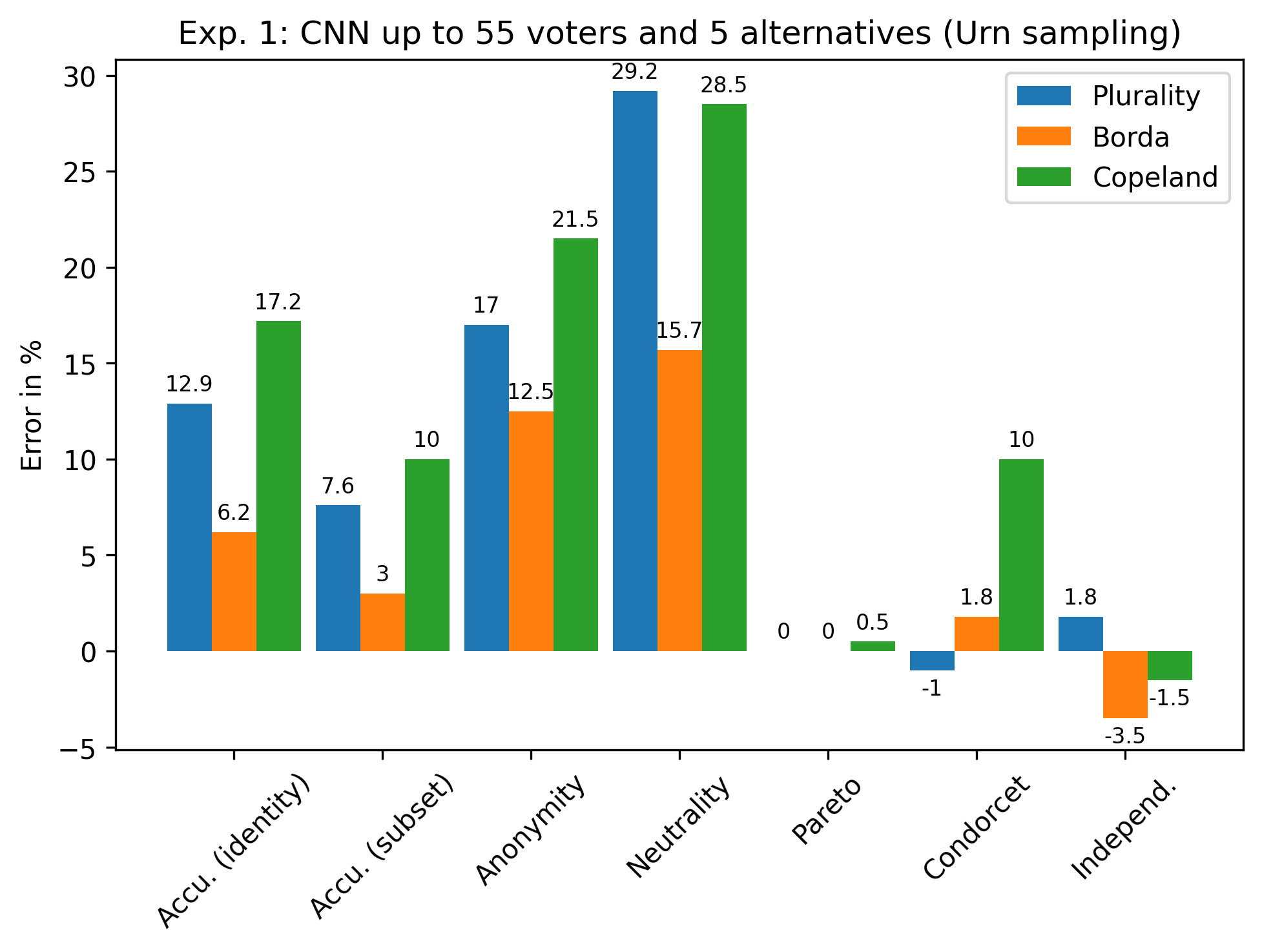}
\includegraphics[width=0.49\linewidth]{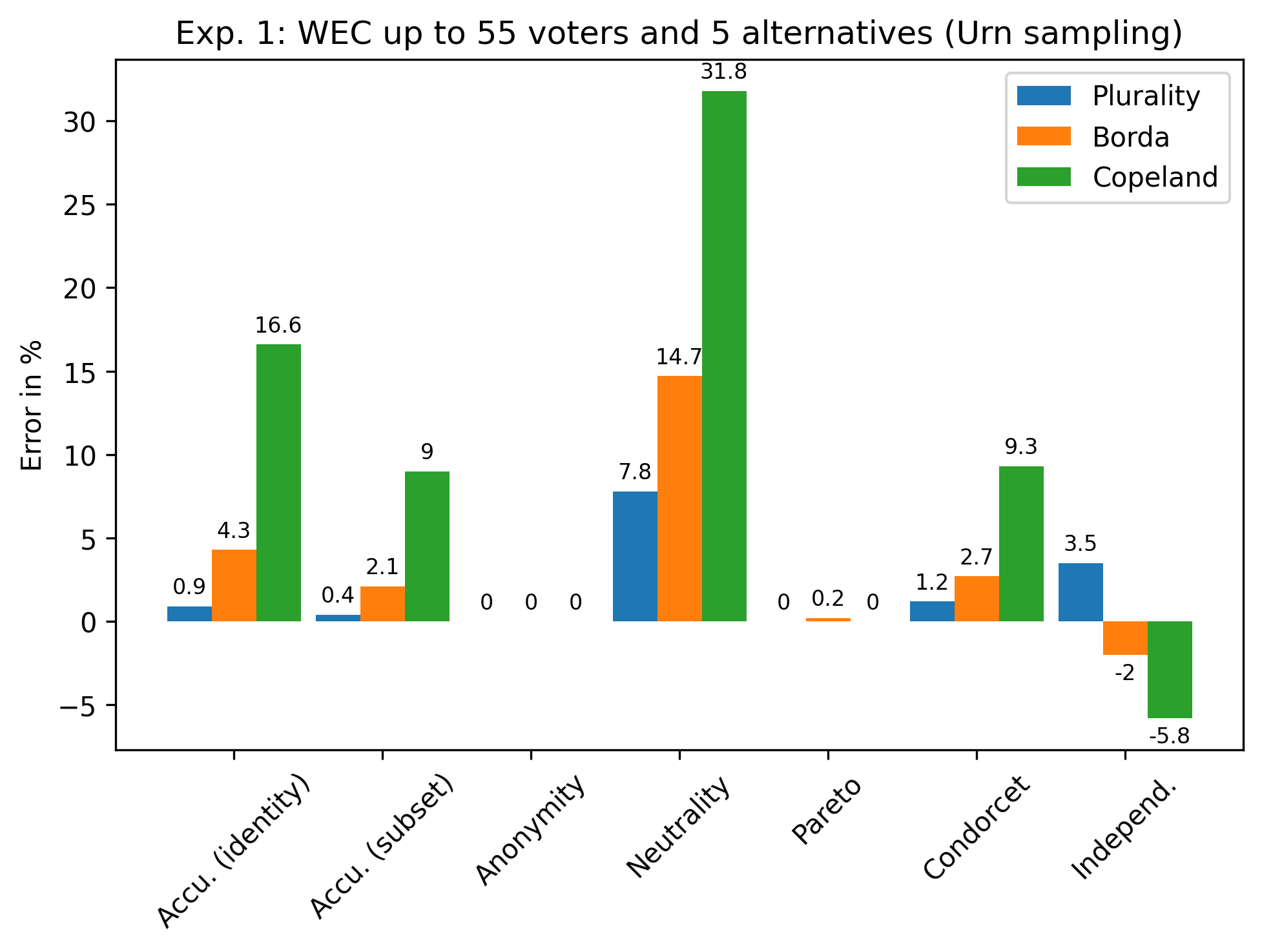}
\Description{Six barplots describing further results of experiment~1 (part~1).}
\caption{Part~1 of more settings of experiment~1 (Section~\ref{ssec: experiment 1}). Varying architectures, rules, and sampling, while comparing the errors in both accuracy and axiom satisfaction.}
\label{fig: exp1 appendix part 1}
\end{figure*}

\begin{figure*}
\centering
\includegraphics[width=0.49\linewidth]{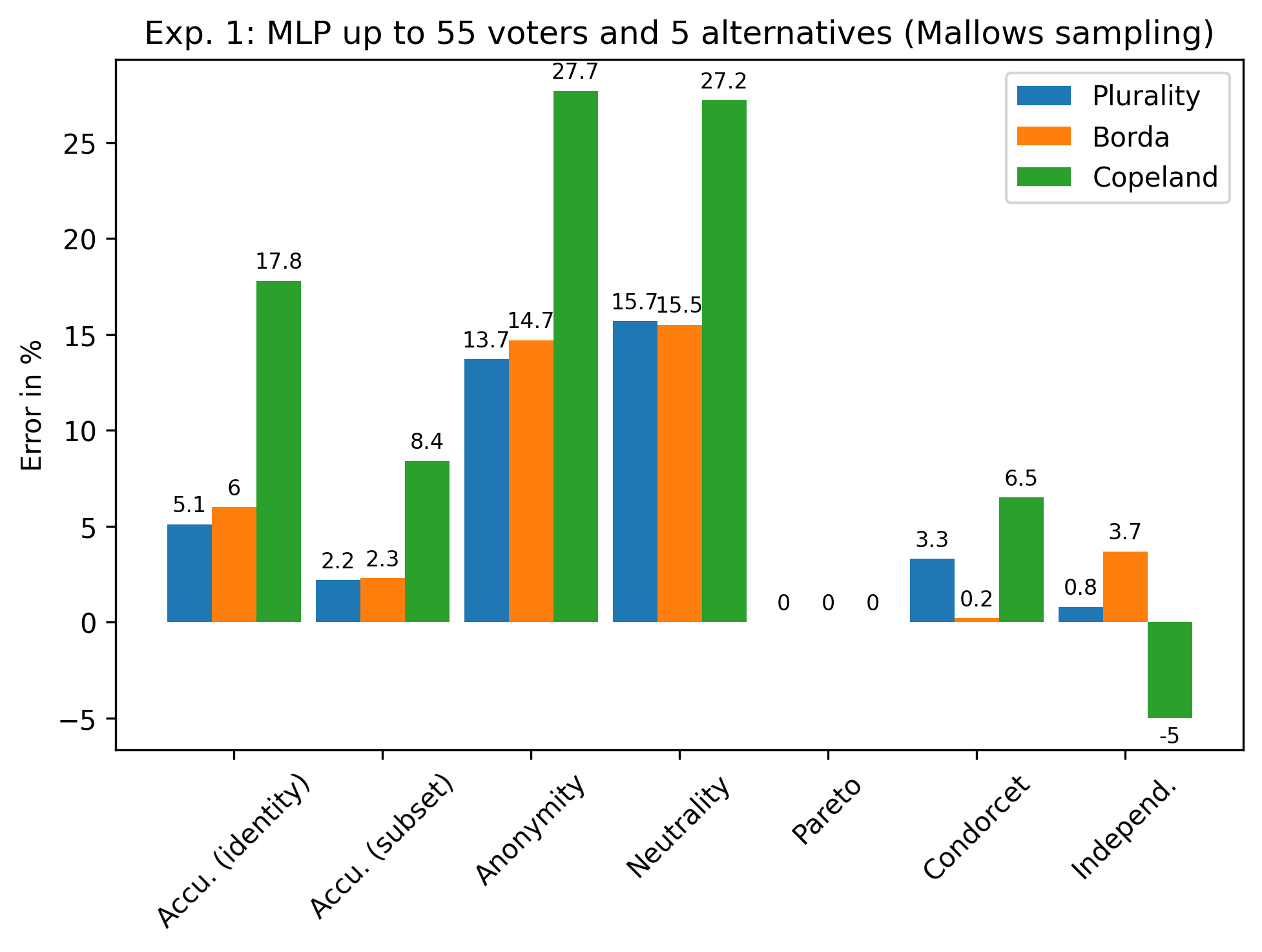}
\includegraphics[width=0.49\linewidth]{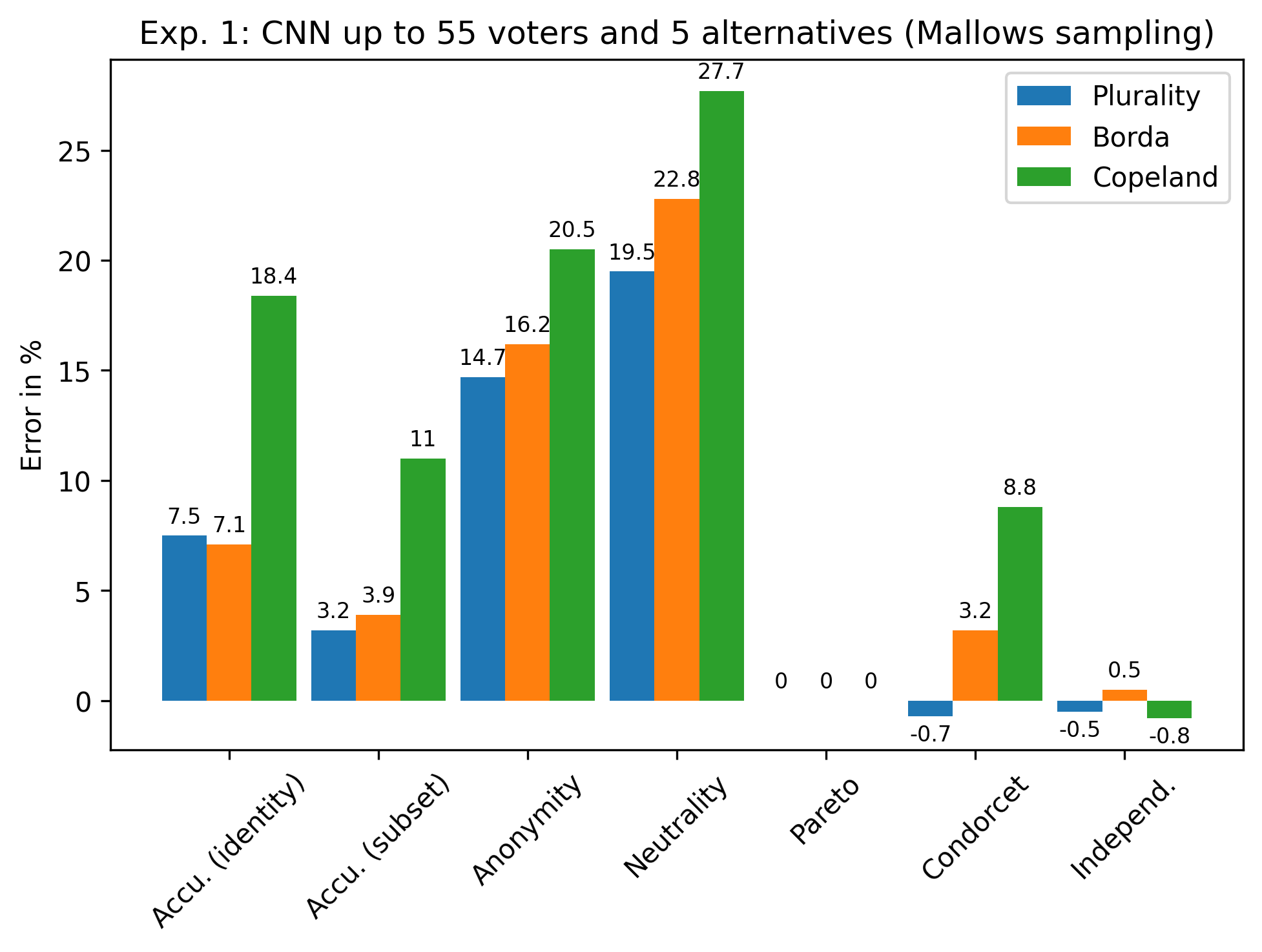}
\\
\includegraphics[width=0.49\linewidth]{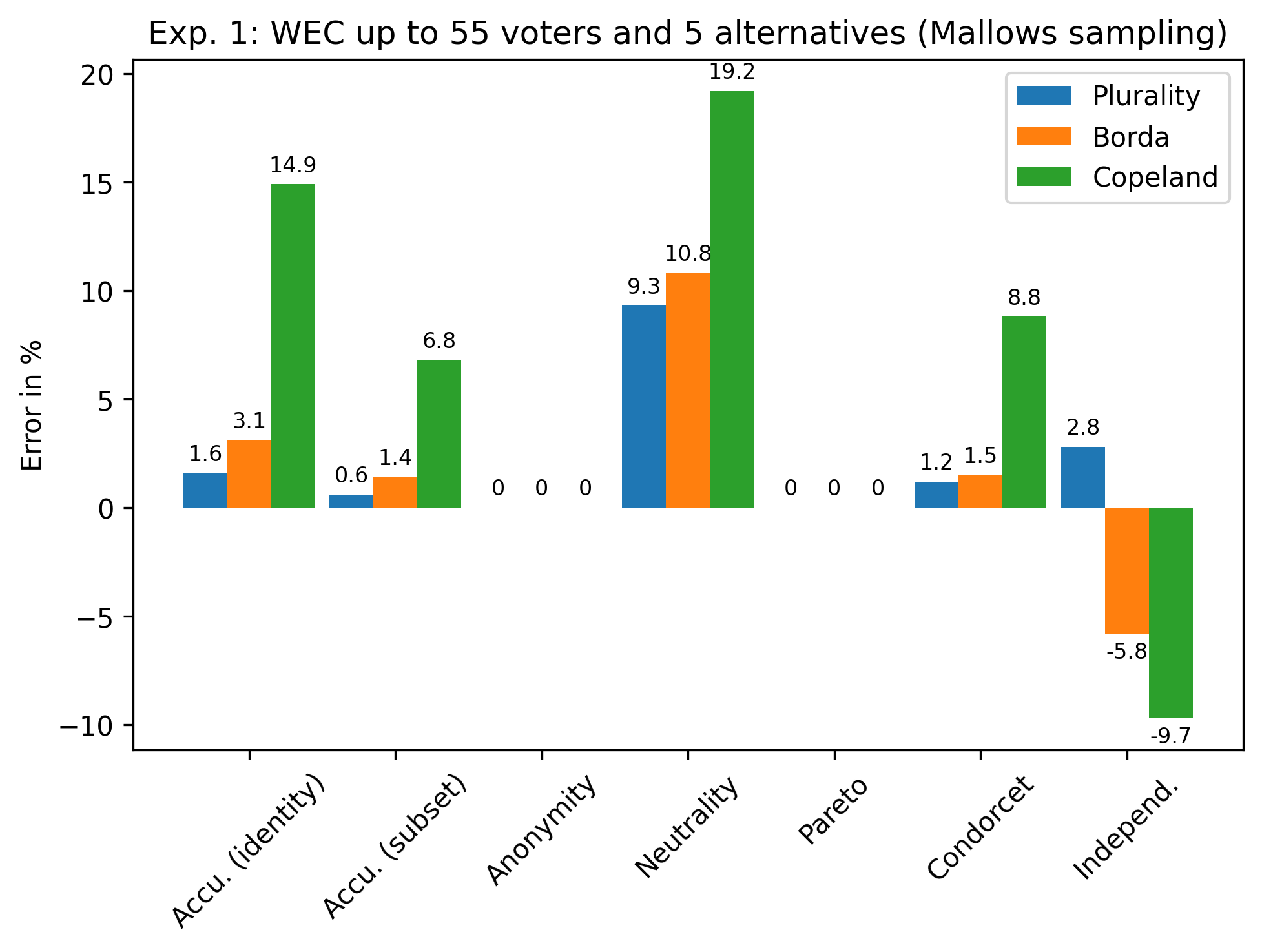}
\includegraphics[width=0.49\linewidth]{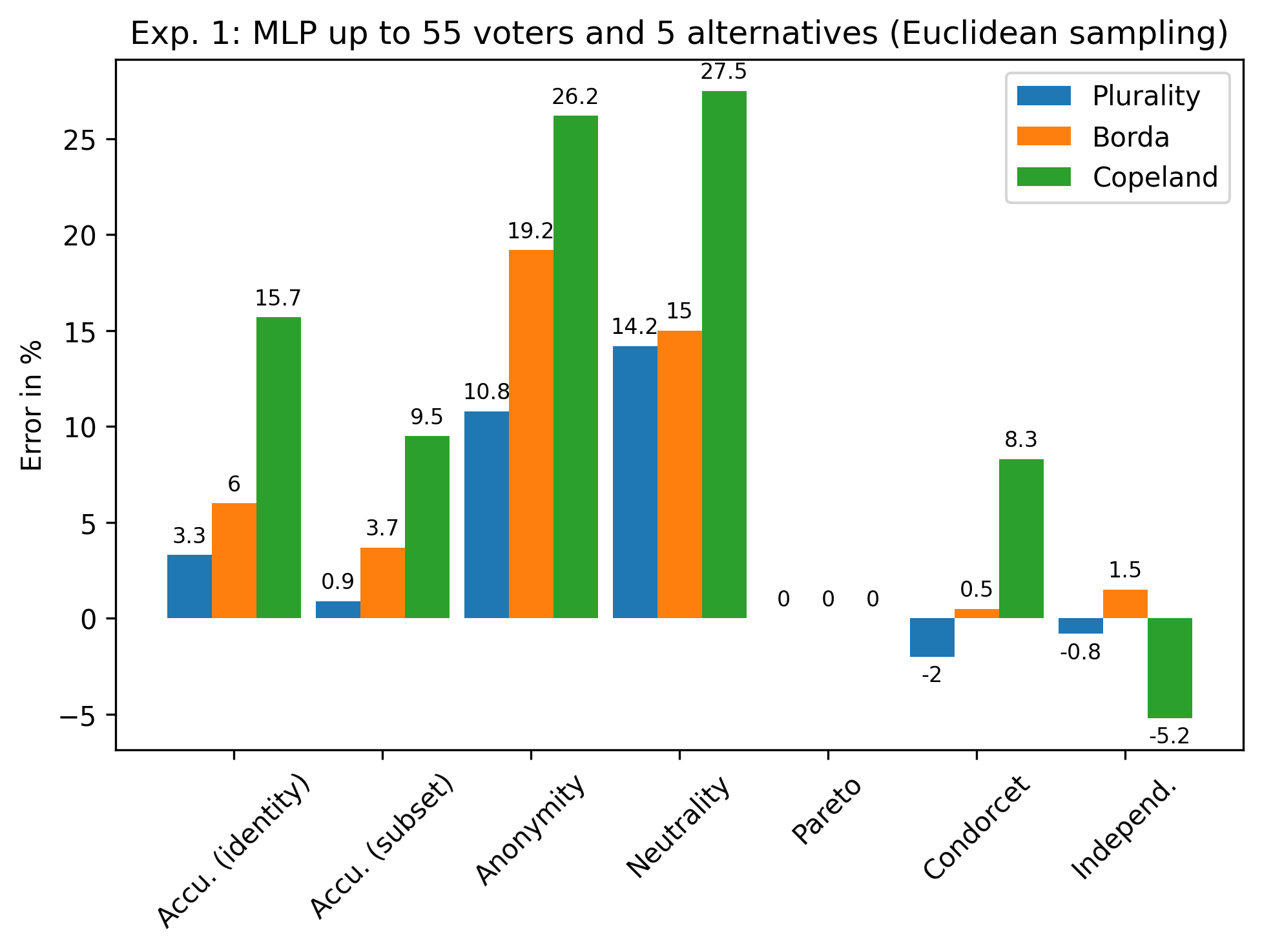}
\\
\includegraphics[width=0.49\linewidth]{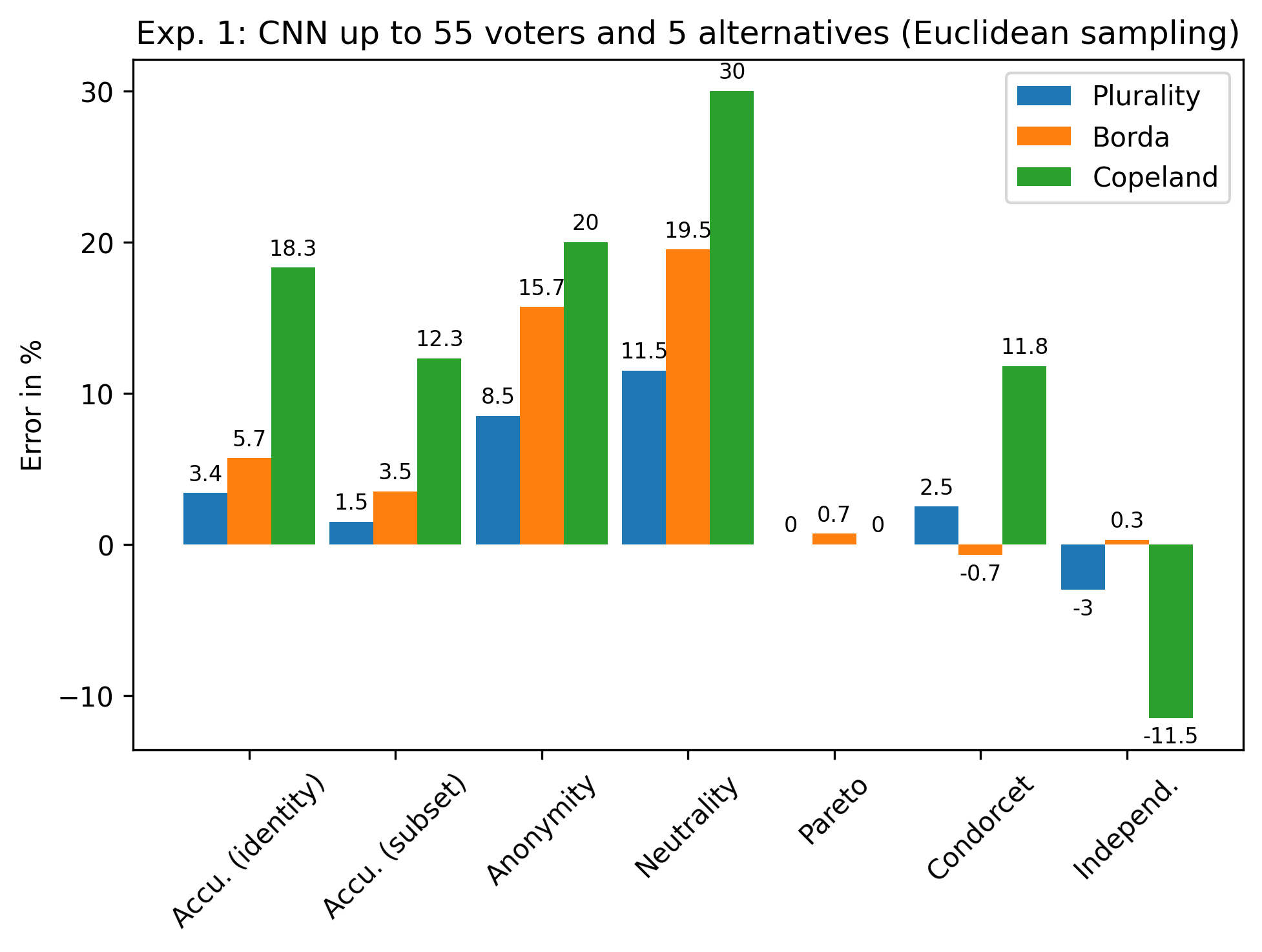}
\includegraphics[width=0.49\linewidth]{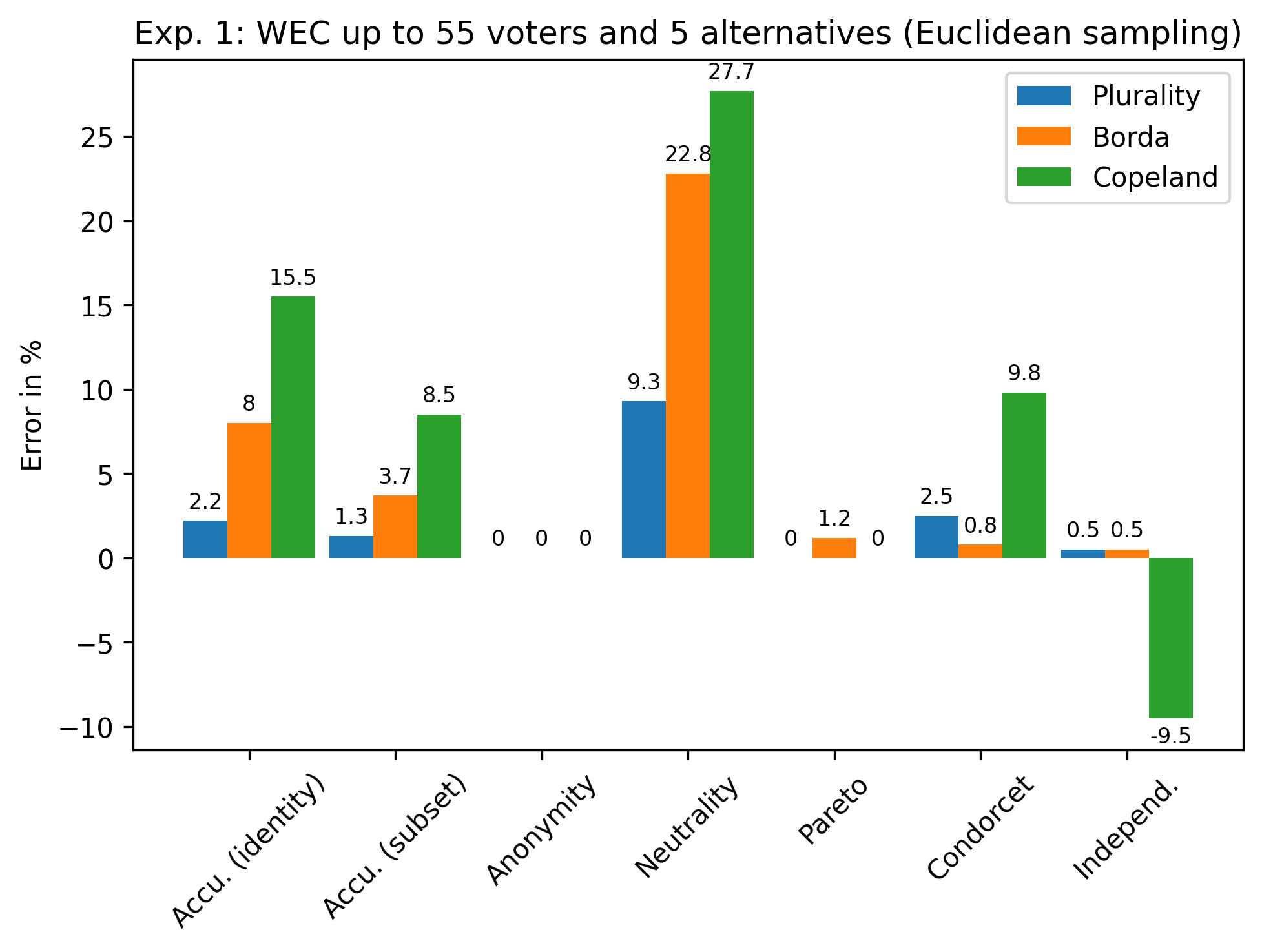}
\Description{Six barplots describing further results of experiment~1 (part~2).}
\caption{Part~2 of more settings of experiment~1 (Section~\ref{ssec: experiment 1}). Varying architectures, rules, and sampling, while comparing the errors in both accuracy and axiom satisfaction.}
\label{fig: exp1 appendix part 2}
\end{figure*}

We run the `correct for the right reasons' experiment from Section~\ref{ssec: experiment 1} in more settings, reported in Figure~\ref{fig: exp1 appendix part 1} (part~1) and Figure~\ref{fig: exp1 appendix part 2} (part~2).

\section{Experiment 2} 
\label{sec: app_exp2}

\begin{figure*}
\begin{center}
\includegraphics[width=0.45\linewidth]{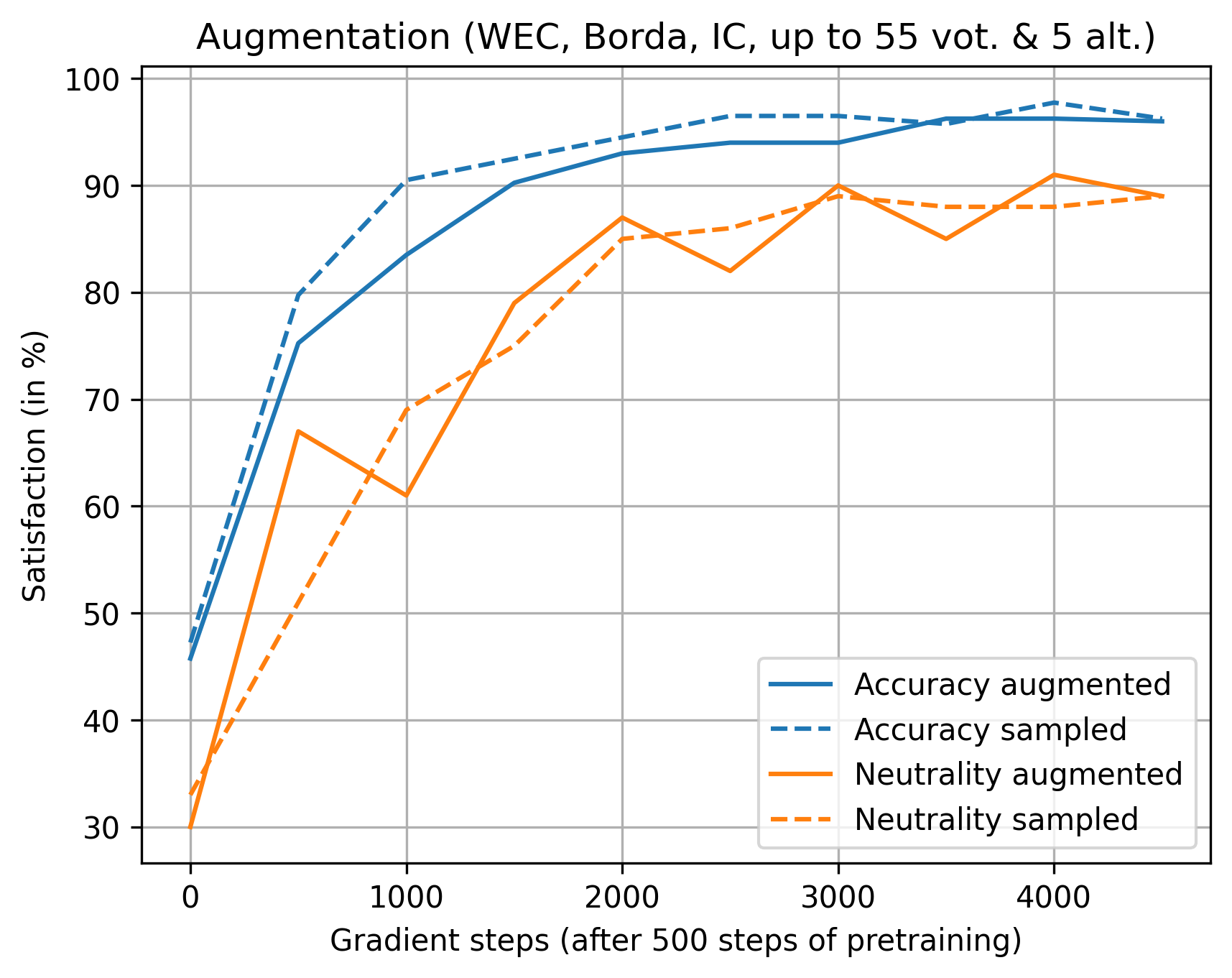}
\includegraphics[width=0.45\linewidth]{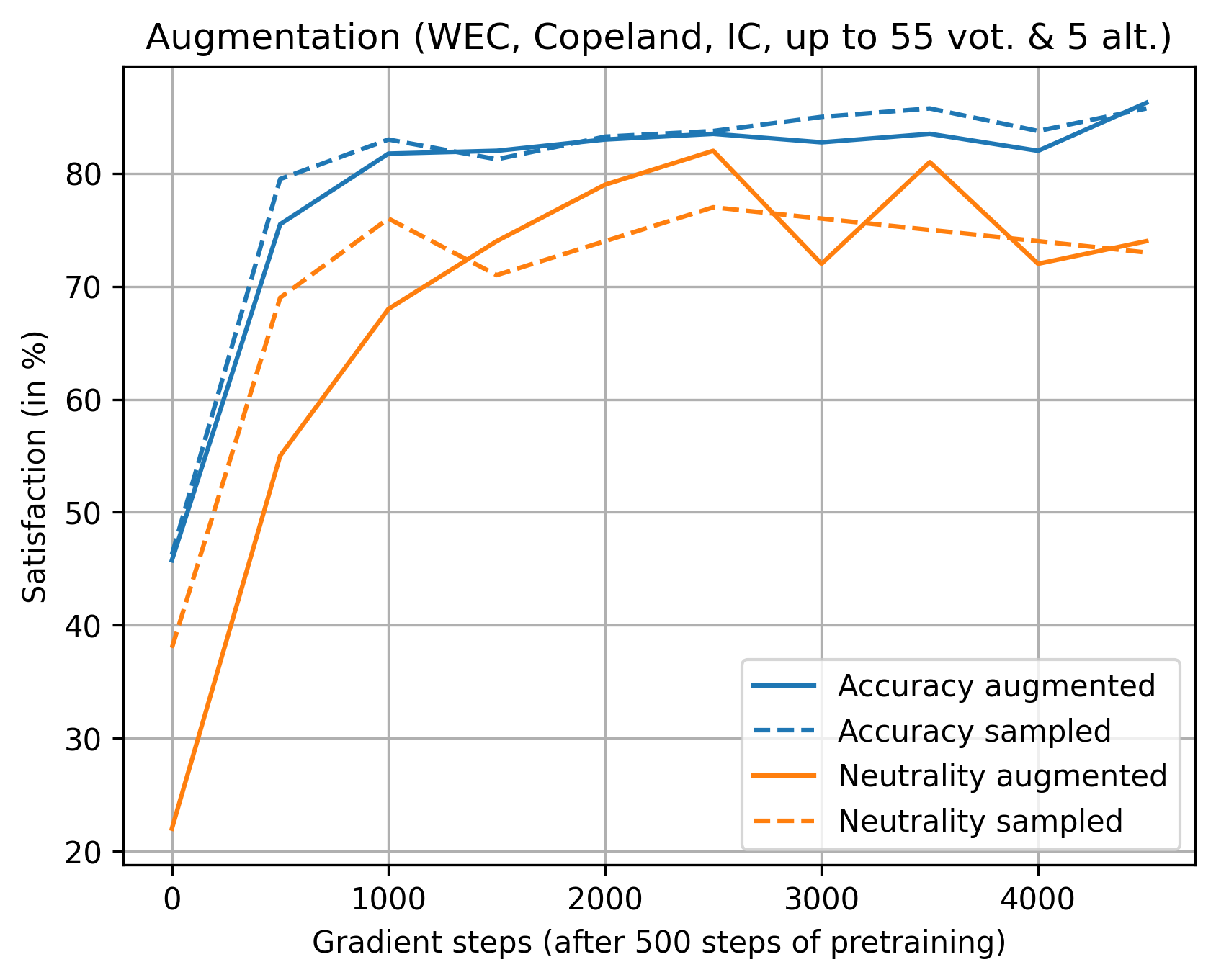}\\
\includegraphics[width=0.45\linewidth]{plot_2025-03-19_19-27-29_aug_WEC_Plurality_IC.png}
\includegraphics[width=0.45\linewidth]{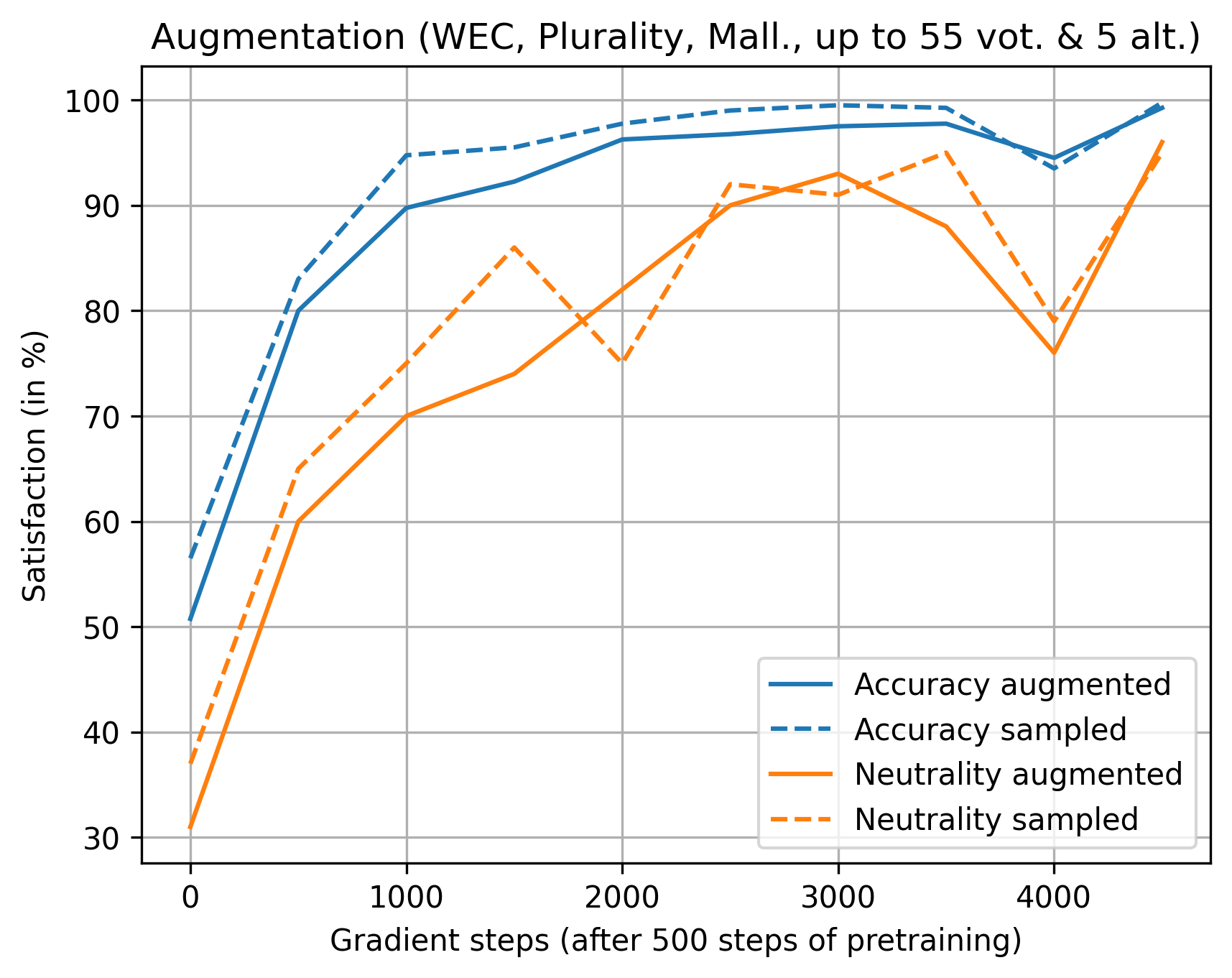}\\
\includegraphics[width=0.45\linewidth]{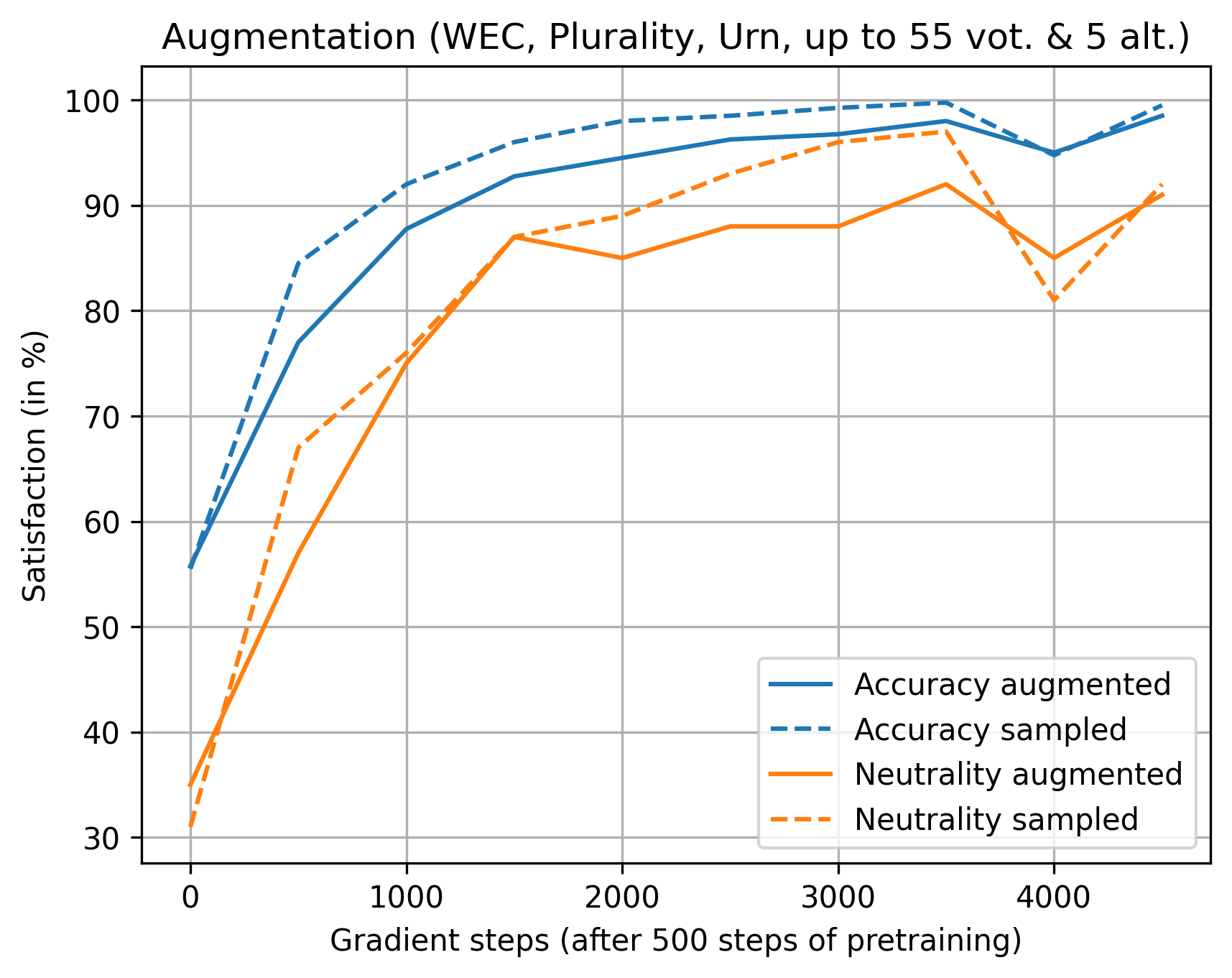}
\includegraphics[width=0.45\linewidth]{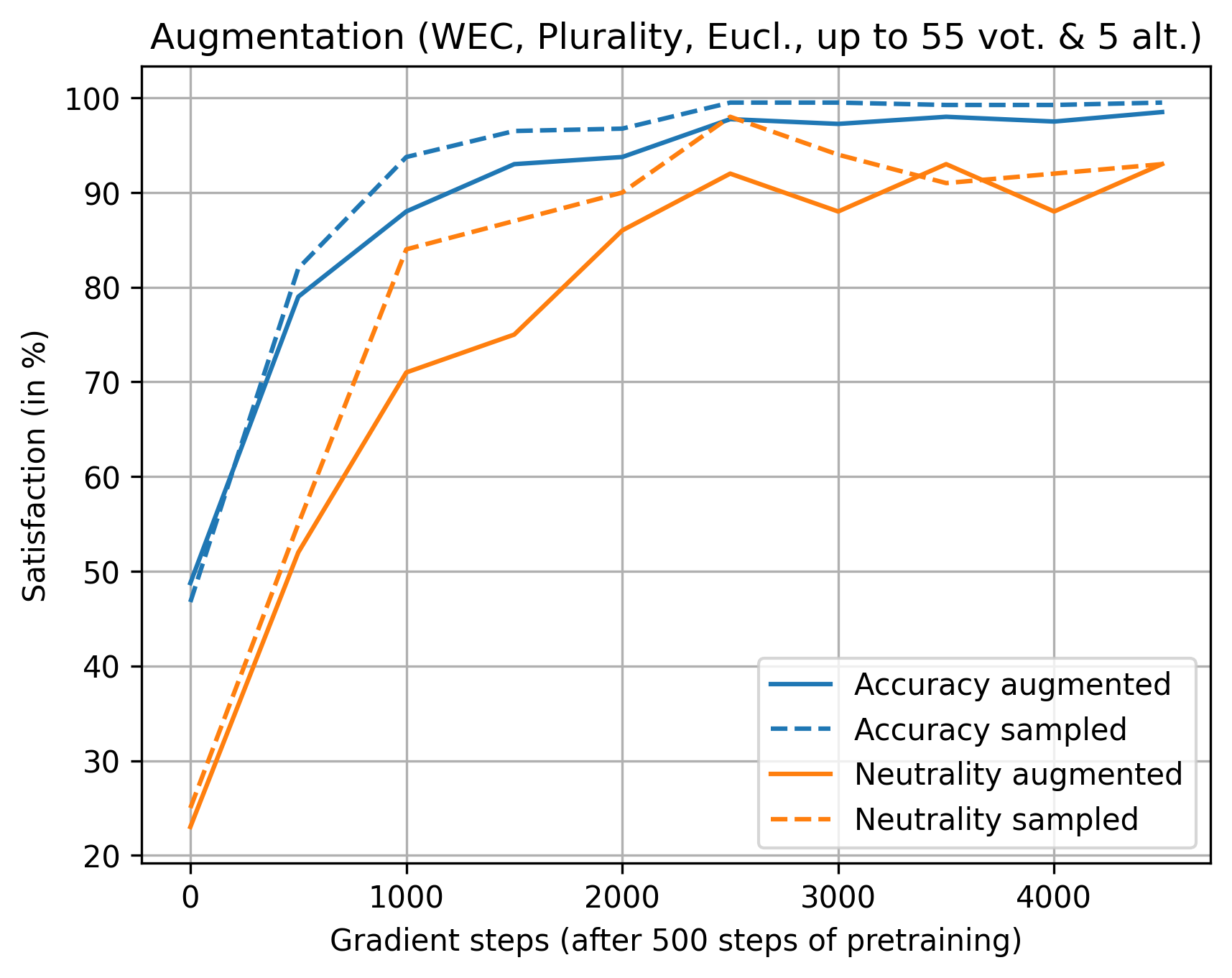}
\Description{Six plots describing further results of experiment~2, in its first version.}
\caption{The first version of experiment~2 (Section~\ref{ssec: experiment 2}) with further architectures, rules, and distributions.}
\label{fig: exp2 app initial}
\end{center}
\end{figure*}

\begin{figure*}
\begin{center}
\includegraphics[width=0.49\linewidth]{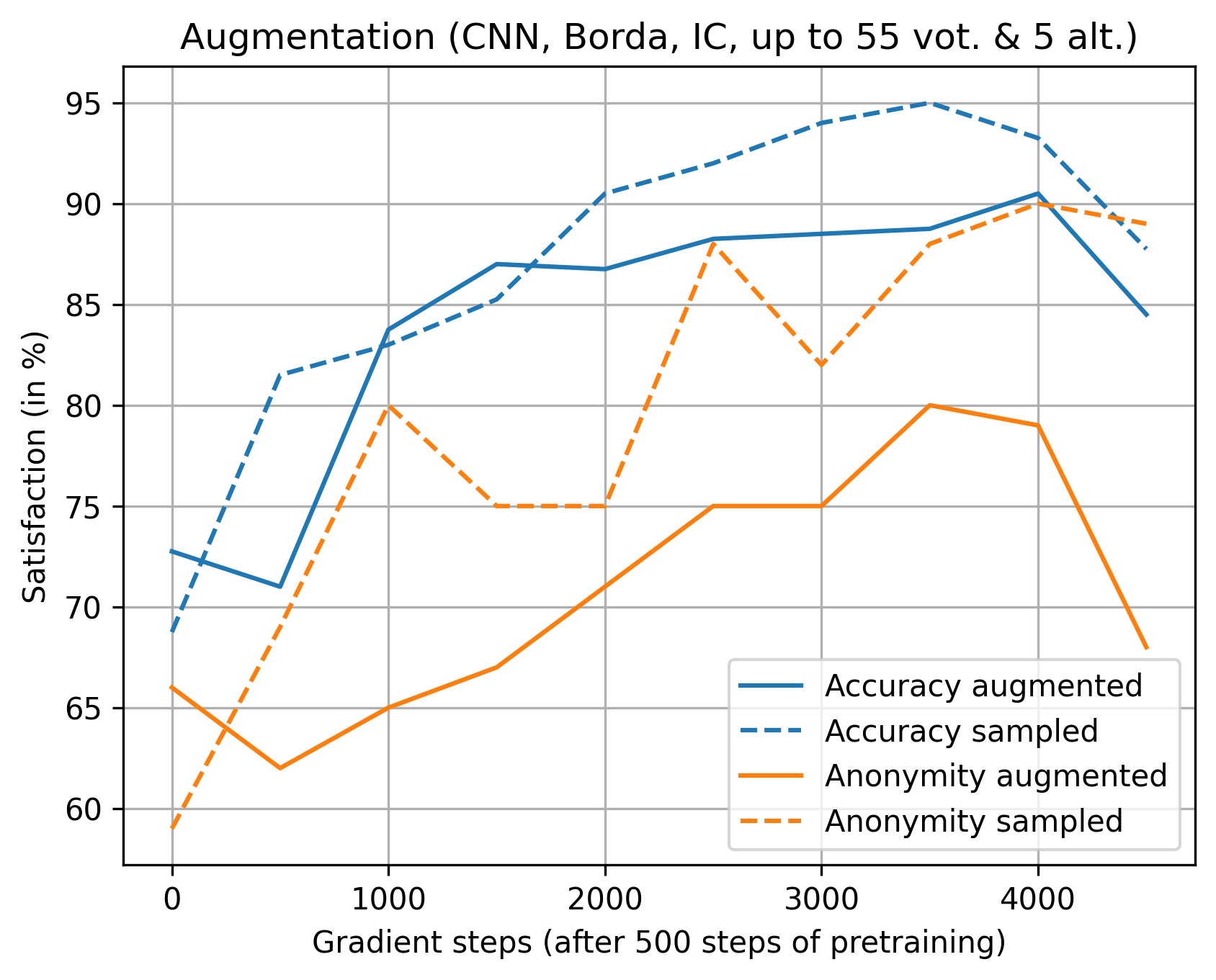}
\includegraphics[width=0.49\linewidth]{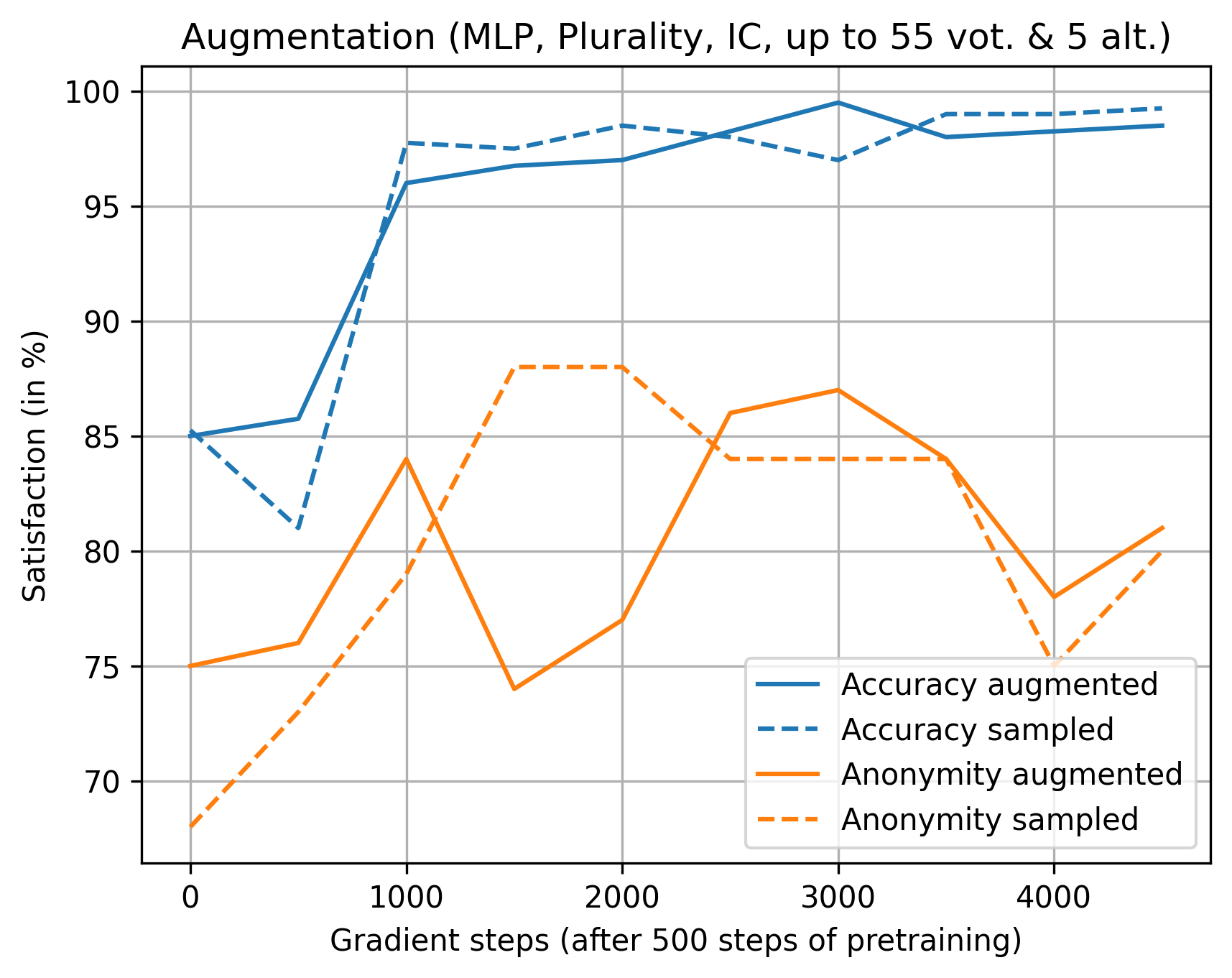}\\
\Description{Two additional plots describing further results of experiment~2, in its first version.}
\caption{The first version of experiment~2 (Section~\ref{ssec: experiment 2}), but for the anonymity axiom (instead of neutrality). Since the WEC is anonymous by design, this can only be tested for MLP and CNN.}
\label{fig: exp2 app initial anonymity}
\end{center}
\end{figure*}

\begin{figure*}
\begin{center}
\includegraphics[width=0.45\linewidth, trim = {0, 0, .85cm, 0}, clip]{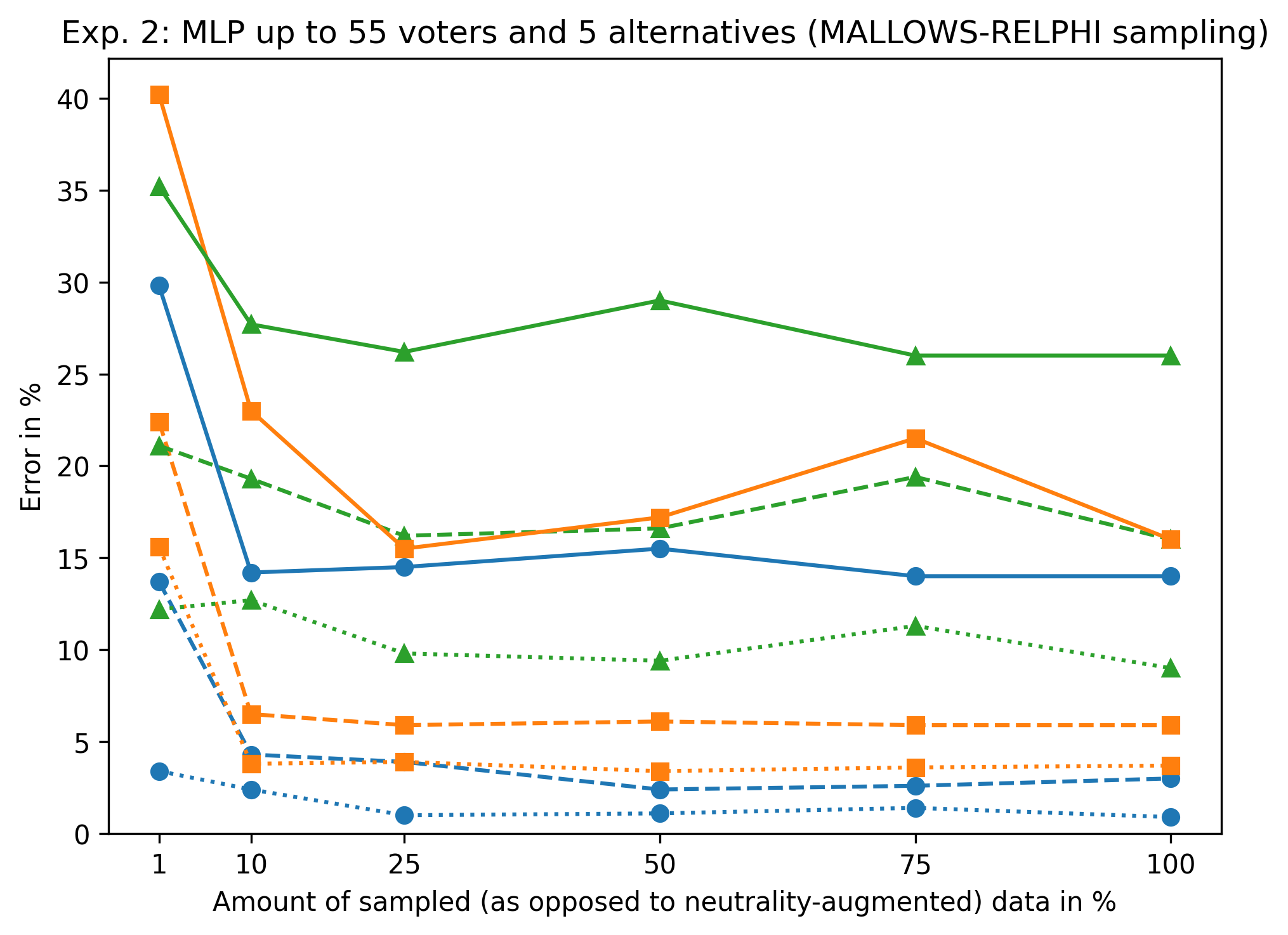}
\includegraphics[width=0.45\linewidth, trim = {0, 0, .85cm, 0}, clip]{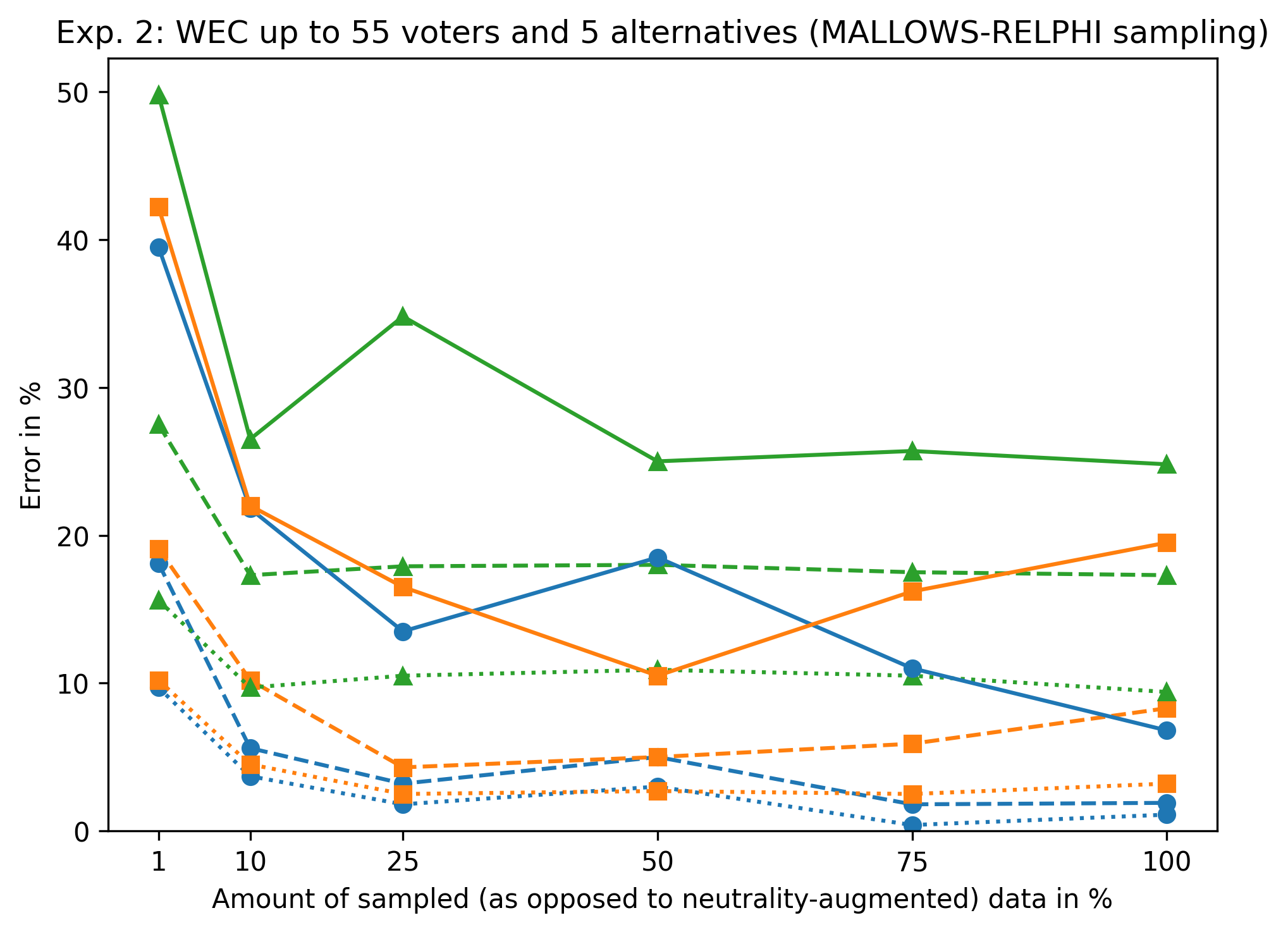}\\
\includegraphics[width=0.45\linewidth, trim = {0, 0, .85cm, 0}, clip]{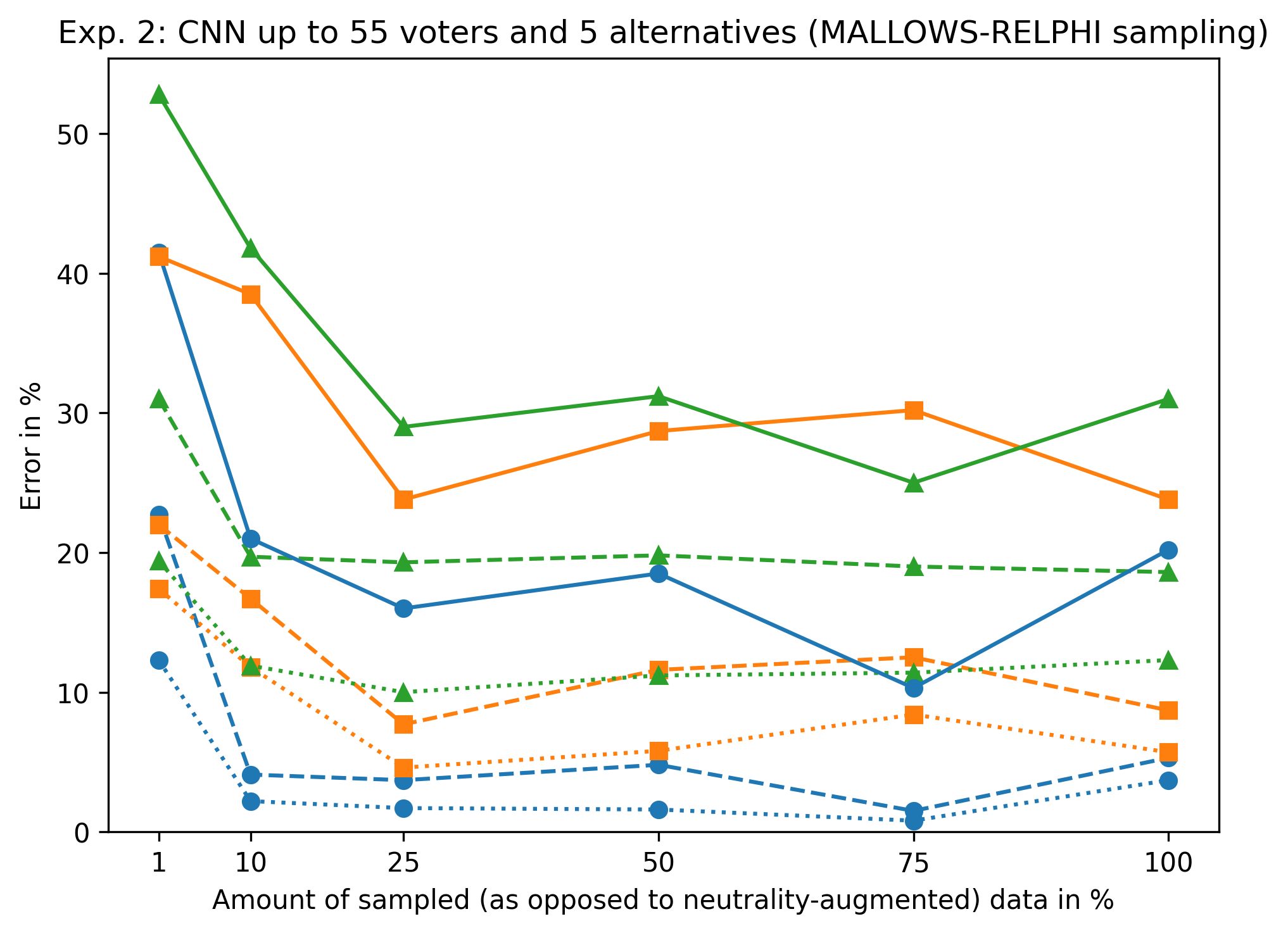}
\includegraphics[width=0.45\linewidth]{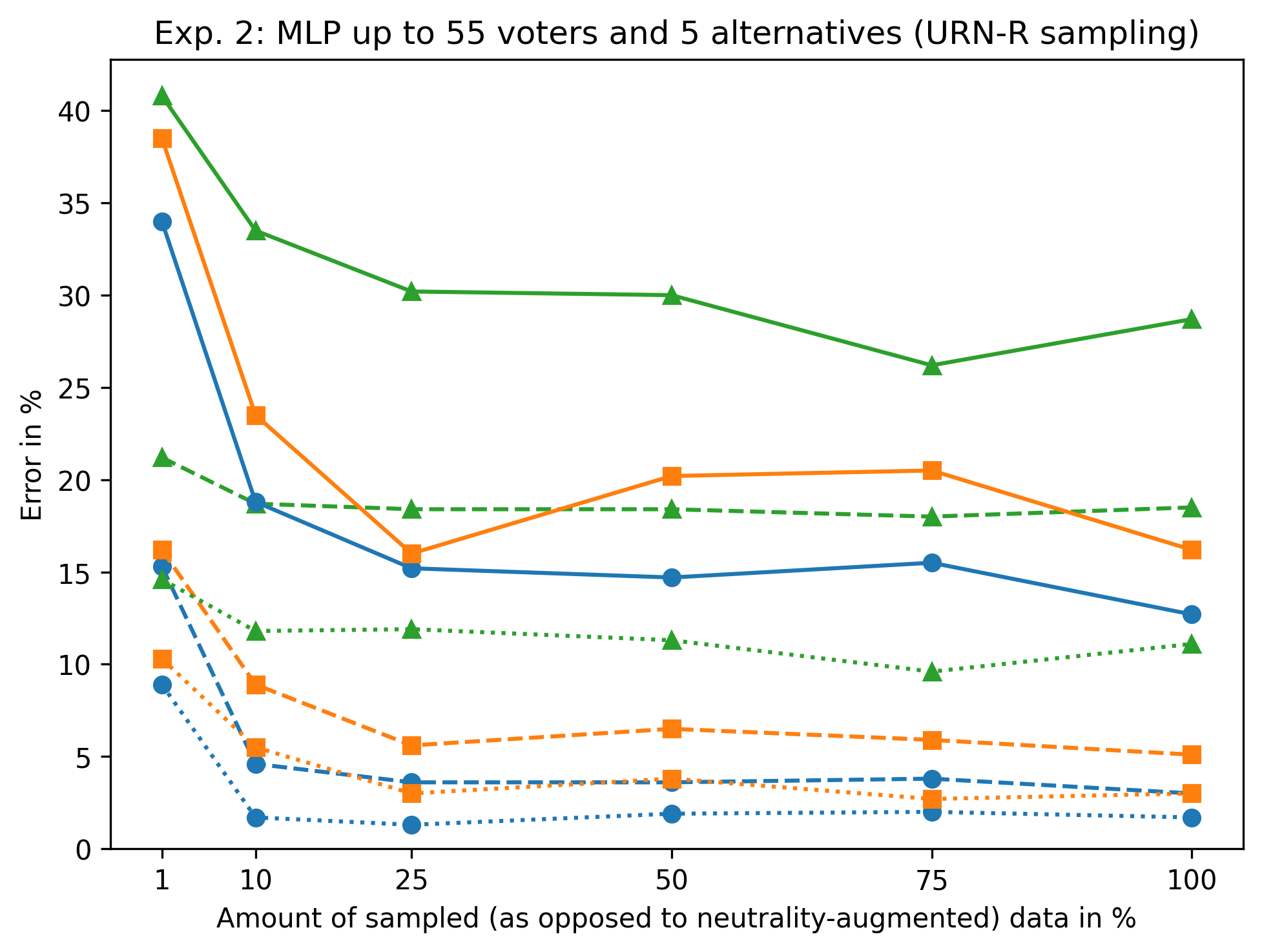}\\
\includegraphics[width=0.45\linewidth]{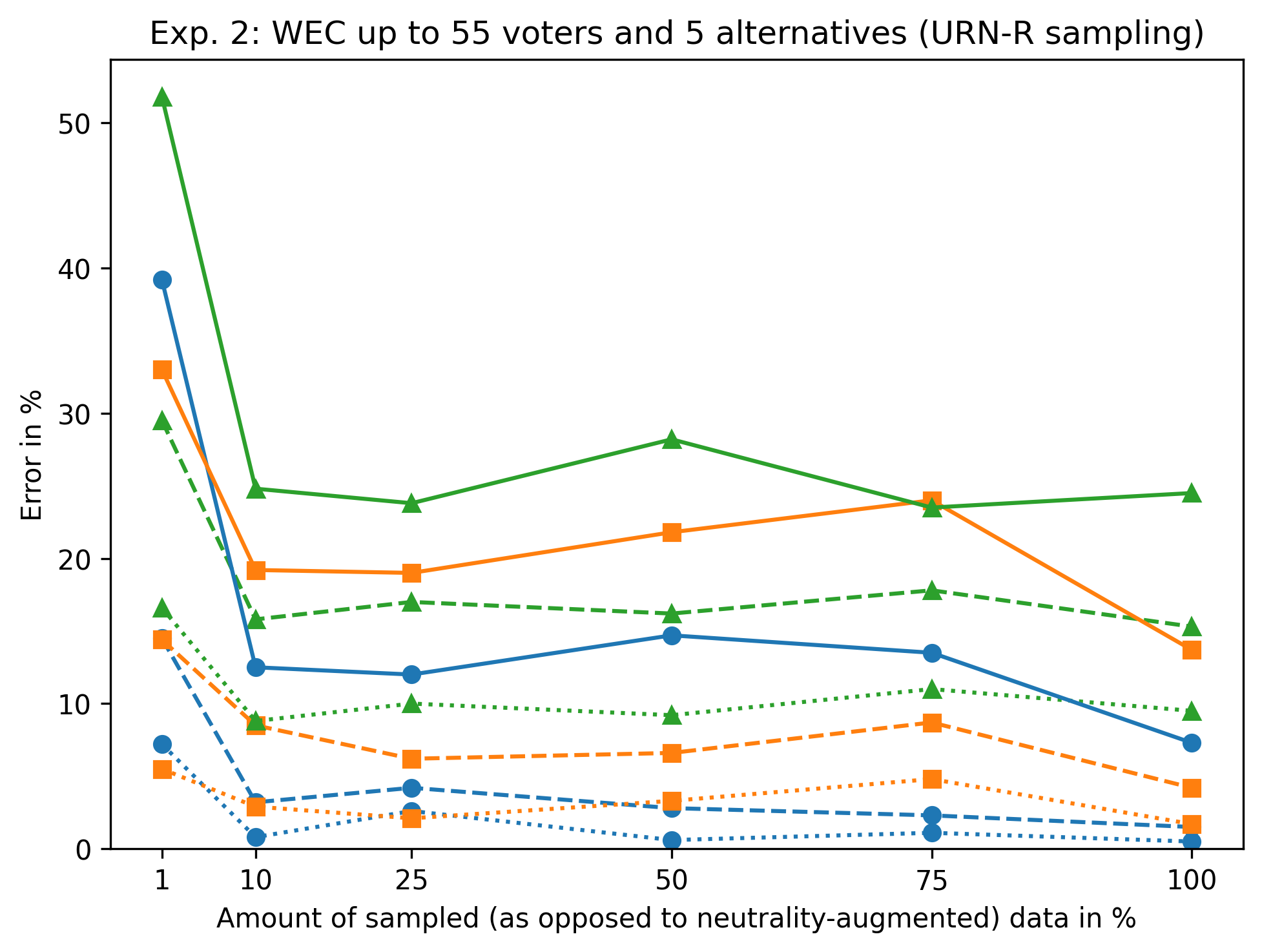}
\includegraphics[width=0.45\linewidth]{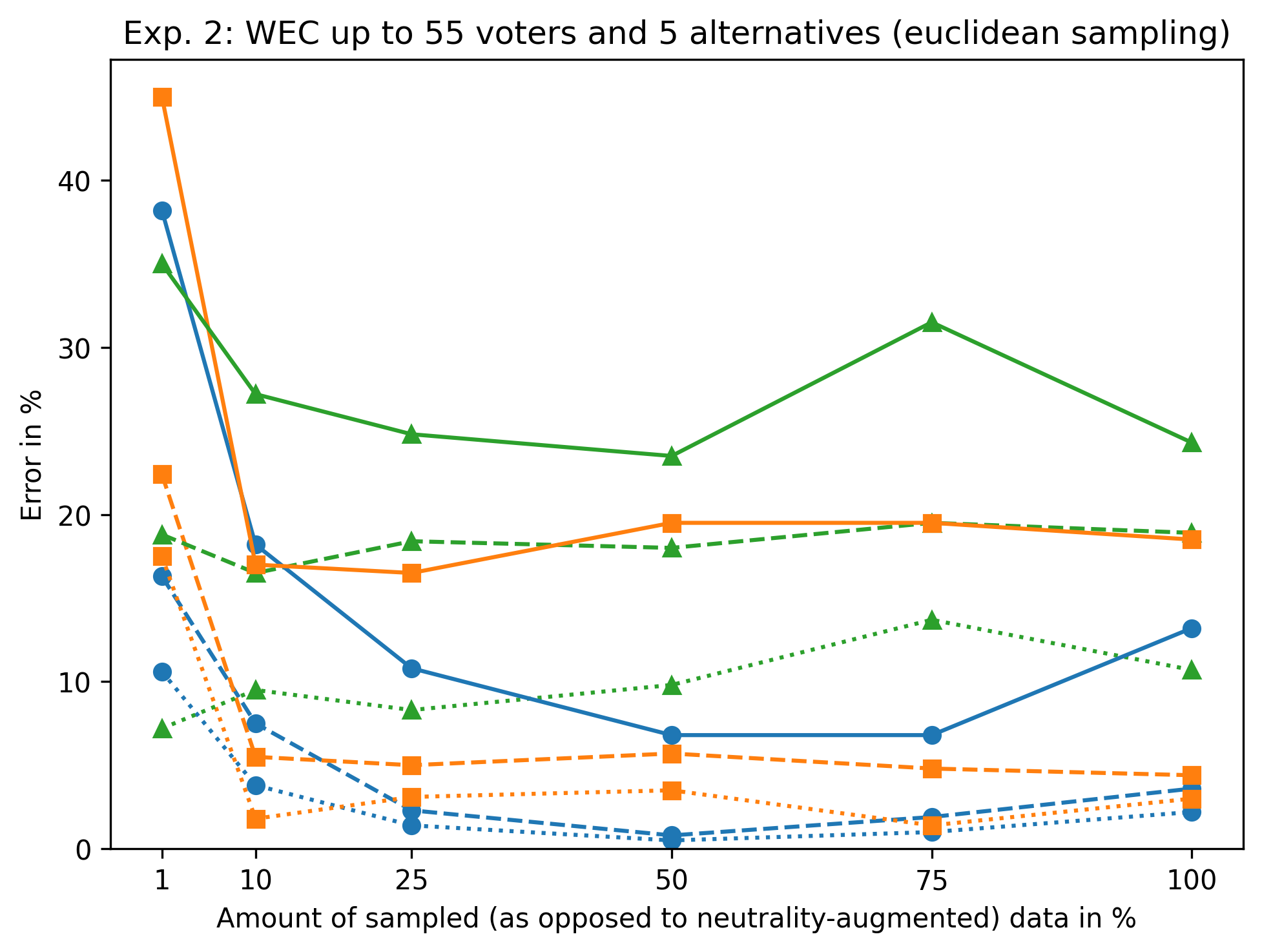}\\
\includegraphics[width=0.7\linewidth]{plot_2025-03-06_09-22-12_MLP_IC_neutrality_legend.png}
\Description{Six plots describing further results of experiment~2, in its second version.}
\caption{The second version of experiment~2 (Section~\ref{ssec: experiment 2}) with different distributions.}
\label{fig: exp2 app different distributions}
\end{center}
\end{figure*}

\begin{figure*}
\begin{center}
\includegraphics[width=0.49\linewidth]{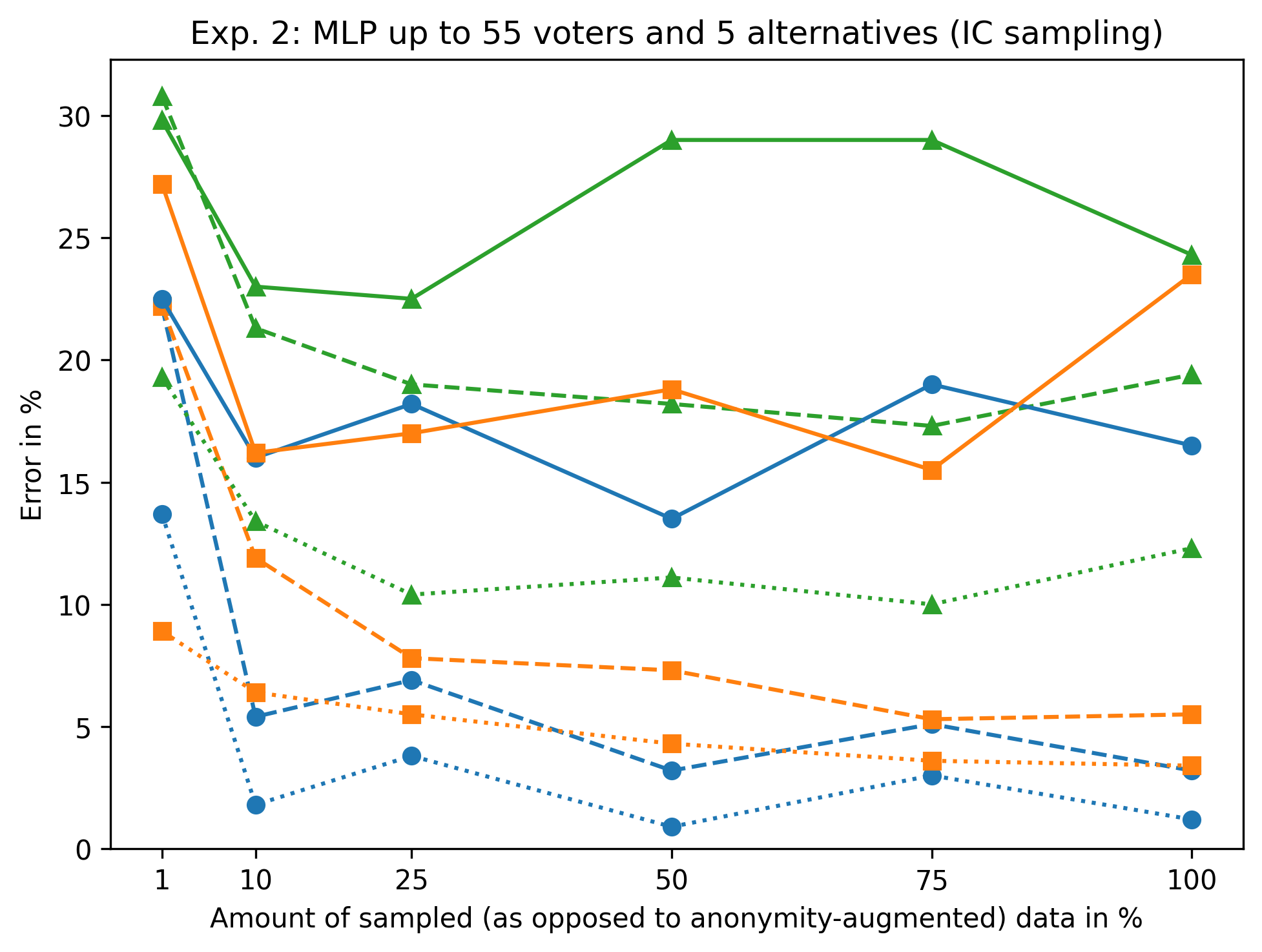}
\includegraphics[width=0.49\linewidth, trim = {0, 0, .85cm, 0}, clip]{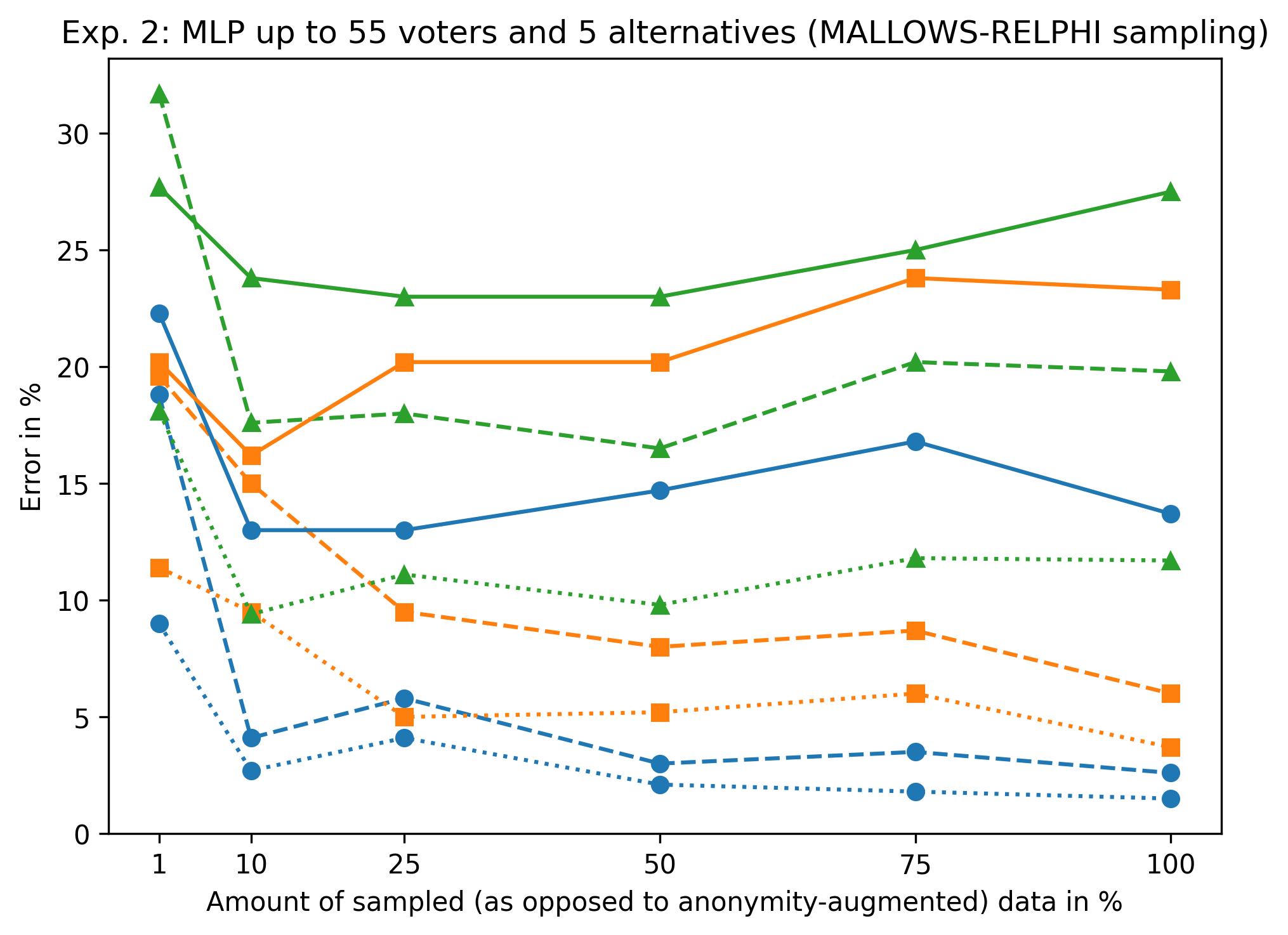}\\
\includegraphics[width=0.49\linewidth, trim = {0, 0, .85cm, 0}, clip]{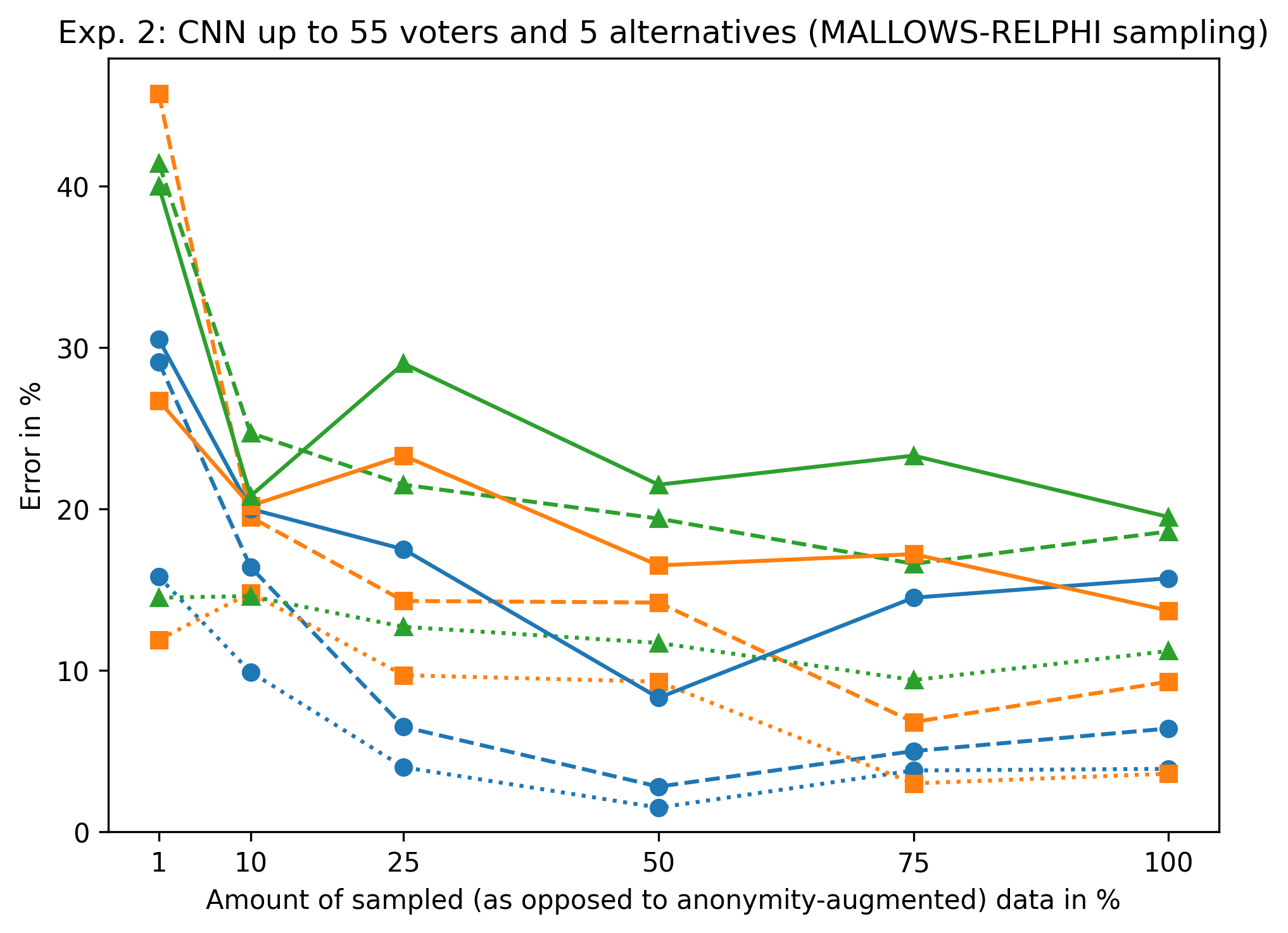}
\includegraphics[width=0.7\linewidth]{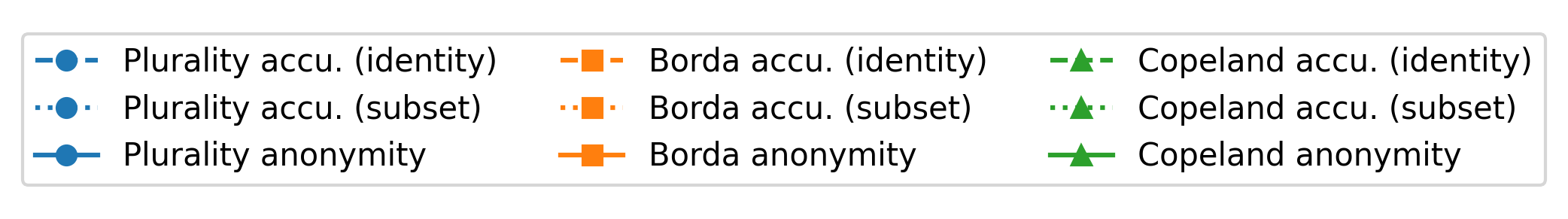}
\Description{Three additional plots describing further results of experiment~2, in its second version.}
\caption{The second version of experiment~2 (Section~\ref{ssec: experiment 2}), but for the anonymity axiom (instead of neutrality). Since the WEC is anonymous by design, this can only be tested for MLP and CNN.}
\label{fig: exp2 app anonymity}
\end{center}
\end{figure*}

We add further results to the `learning principles by example' experiment from Section~\ref{ssec: experiment 2}.
Figure~\ref{fig: exp2 app initial} and~\ref{fig: exp2 app initial anonymity} show different choices of architecture, rules, and distribution for the first version of the experiment.
Figure~\ref{fig: exp2 app different distributions} and~\ref{fig: exp2 app anonymity} show different choices of architecture, rules, and distribution for the second version of the experiment.

\section{Experiment 3}
\label{sec: app_exp3}

We add further results on the `rule synthesis guided by principles' experiment (Section~\ref{ssec: experiment 3}).

\subsection{More on the Experiment with IC}

Regarding difference-making profiles, Figure~\ref{fig: exp 3 differing profile IC} in the main text shows a profile where the model \emph{strongly disagrees} with its 5 closest voting rules, i.e., the model's winning set does not intersect the winning set of these rules.
Figure~\ref{fig: exp 3 weak differing profile IC} here shows a profile where the model \emph{weakly disagrees} with all considered voting rules, i.e., the model's winning set is not identical with the winning set of these rules.
(For the main text profile, the model also happens to weakly disagree with all considered rules.)
We found these profiles by going through 10,000 IC-sampled profiles and picking the weakly or strongly disagreeing profile with the smallest number of voters.

\floatstyle{boxed}
\restylefloat{figure}
\begin{figure}
\centering
\begin{footnotesize}
\begin{tabular}{cccccccc}
\toprule
  1    &  2    &  3    &  4    &  5    &  6    &  7    \\
\midrule
  $a$  &  $b$  &  $c$  &  $e$  &  $b$  &  $b$  &  $d$  \\
  $e$  &  $d$  &  $e$  &  $c$  &  $c$  &  $c$  &  $e$  \\
  $d$  &  $c$  &  $a$  &  $d$  &  $a$  &  $a$  &  $b$  \\
  $b$  &  $a$  &  $d$  &  $a$  &  $d$  &  $d$  &  $c$  \\
  $c$  &  $e$  &  $b$  &  $b$  &  $e$  &  $e$  &  $a$  \\
\bottomrule
\end{tabular} \\
\smallskip
\begin{tabular}{rl}
$\{ b,c \}$ & neutrality-averaged WEC, with sigmoids (rounded)
$a$:.38, 
$b$:.54, 
$c$:.50, 
$d$:.39, 
$e$:.39 
\\  
$\{ b \}$ & Plurality, Weak Nanson, Kemeny-Young\\
$\{ c \}$ & Borda, Copeland, Llull, Blacks, Coombs \\
$\{ d \}$ & Anti-Plurality, Baldwin \\
$\{ e \}$ & Instant Runoff TB \\
$\{ b, d\}$ & Stable Voting \\
$\{ b, c, d\}$ & Uncovered Set, Banks \\
$\{ b, c, e\}$ & PluralityWRunoff PUT \\
$\{a,b,c,d,e\}$ & Top Cycle \\
\end{tabular}
\end{footnotesize}
\Description{A profile on which the `WEC n' model weakly disagrees (IC sampling).}
\caption{IC sampling: Profile where the `WEC n' model weakly disagrees (i.e. non-identical winning sets) with existing voting rules.}
\label{fig: exp 3 weak differing profile IC}
\end{figure}

\floatstyle{plain}
\restylefloat{figure}

Table~\ref{tbl: exp 3 continued training} suggests that further optimization does not further improve axiom satisfaction. 

\begin{table*}
\begin{footnotesize}
\begin{center}
\begin{tabular}{lcccccc}
\toprule
                            &  Anon.\  &   Neut.\  &  Condorcet  &  Pareto  &  Indep.\  &  Average  \\
\midrule
 WEC n (NW, C, P, round 0)    &   100   &   100    &    97.5     &   100    &    46    &  88.7  \\
 WEC n (NW, C, P, round 1)    &   100   &   100    &     100     &   100    &   38.5   &  87.7  \\
 WEC n (NW, C, P, round 2)    &   100   &   100    &     100     &   100    &   34.8   &   87   \\
 WEC n (NW, C, P, I, round 3) &   100   &   100    &     100     &   100    &   31.8   &  86.3  \\
\bottomrule
\end{tabular}
\end{center}
\end{footnotesize}
\caption{IC sampling: The result of keeping on training an WEC model. Each round adds 20k gradient steps to the previous one. Round~1 is with a learning rate of $10^{-3}$, round~2 with $10^{-4}$, round~3 with $5*10^{-5}$, and round~4 the same but with added optimization of independence. 
}
\label{tbl: exp 3 continued training}
\end{table*}

Table~\ref{tbl: exp 3 averaged training IC} shows the statistical robustness of the axiom satisfaction achieved by the model. 

\begin{table*}
\begin{footnotesize}
\begin{center}
\begin{tabular}{lcccccc}
\toprule
                  &  Anon.  &   Neut.  &  Condorcet  &  Pareto  &  Indep.  &  Avg.  \\
\midrule
 Blacks           &   100   &   100    &     100     &   100    &  36.04   &  87.2  \\
 Stable Voting    &   100   &   100    &     100     &   100    &  40.48   &  88.1  \\
 Borda            &   100   &   100    &    93.82    &   100    &  37.72   & 86.32  \\
 Weak Nanson      &   100   &   100    &     100     &   100    &  38.28   & 87.68  \\
 Copeland         &   100   &   100    &     100     &   100    &  28.54   & 85.72  \\
 WEC n (NW, C, P) &   100   &   100    &    96.78    &   100    &   45.9   & 88.54  \\
\bottomrule
\end{tabular}
\end{center}
\end{footnotesize}
\caption{
IC sampling: Take the average over 5 runs of checking the axiom satisfaction of the `WEC n' model and its closest rules.
}
\label{tbl: exp 3 averaged training IC}
\end{table*}

\begin{figure*}
\begin{center}
\includegraphics[width=0.49\linewidth]{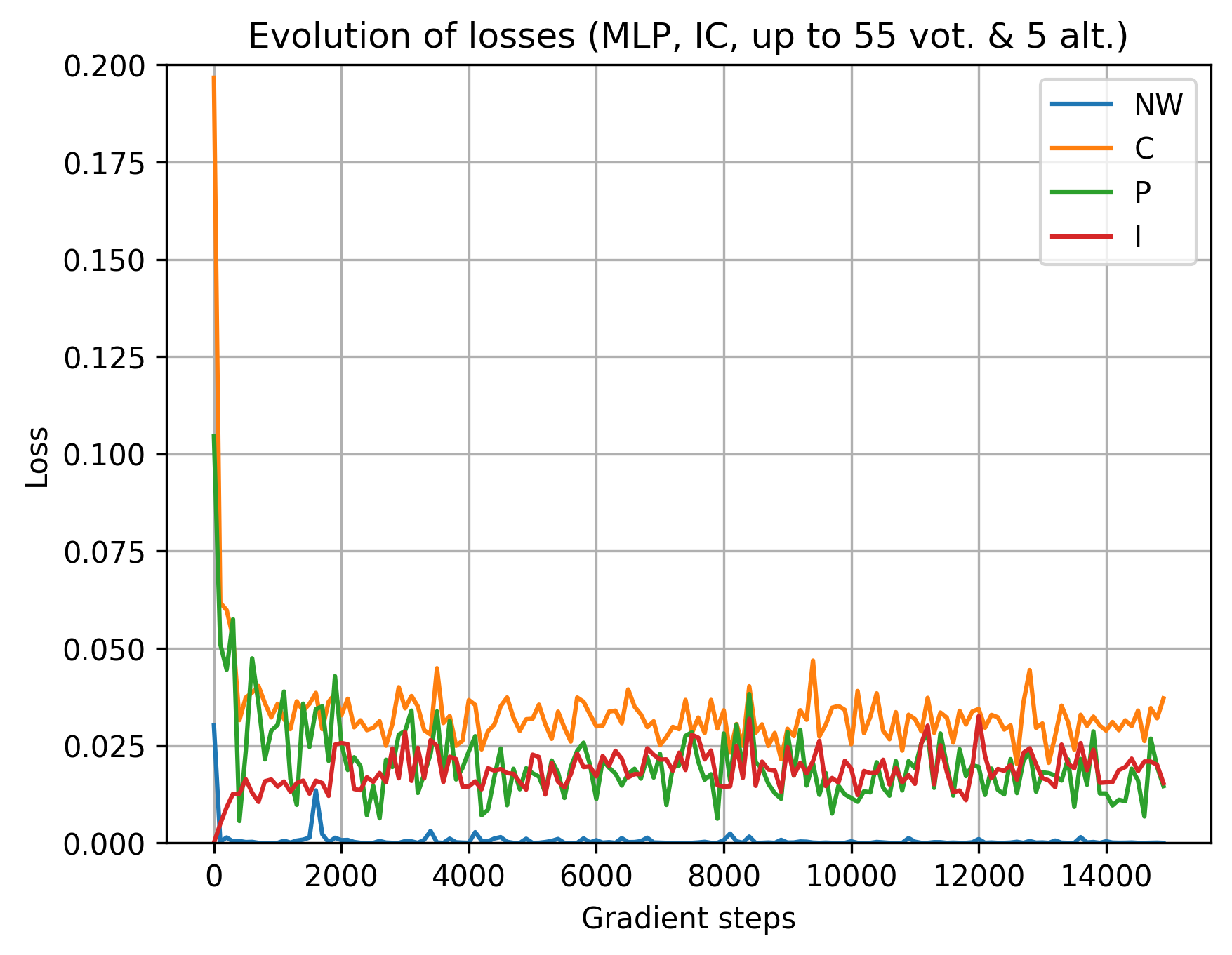}
\includegraphics[width=0.49\linewidth]{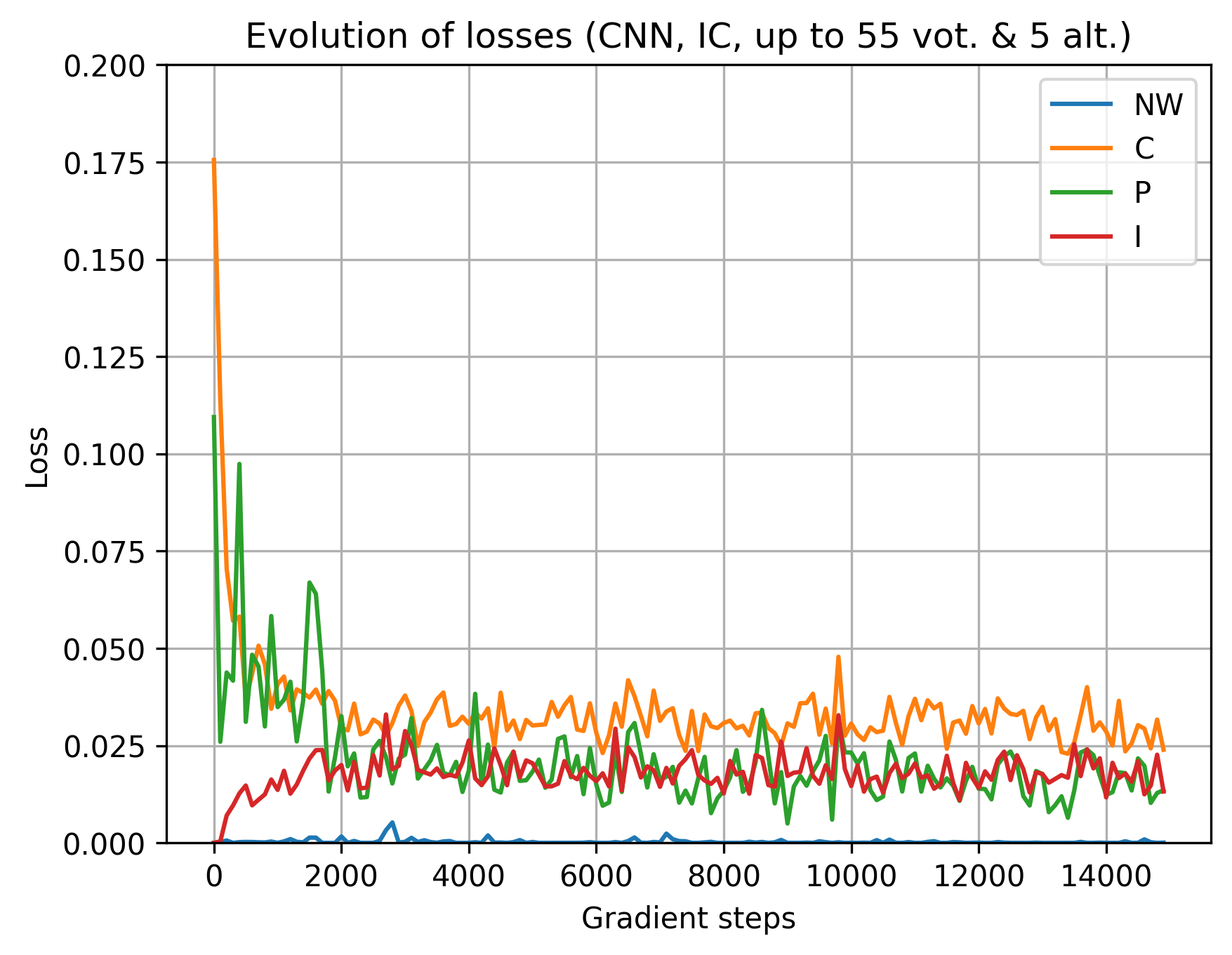}\\
\includegraphics[width=0.49\linewidth]{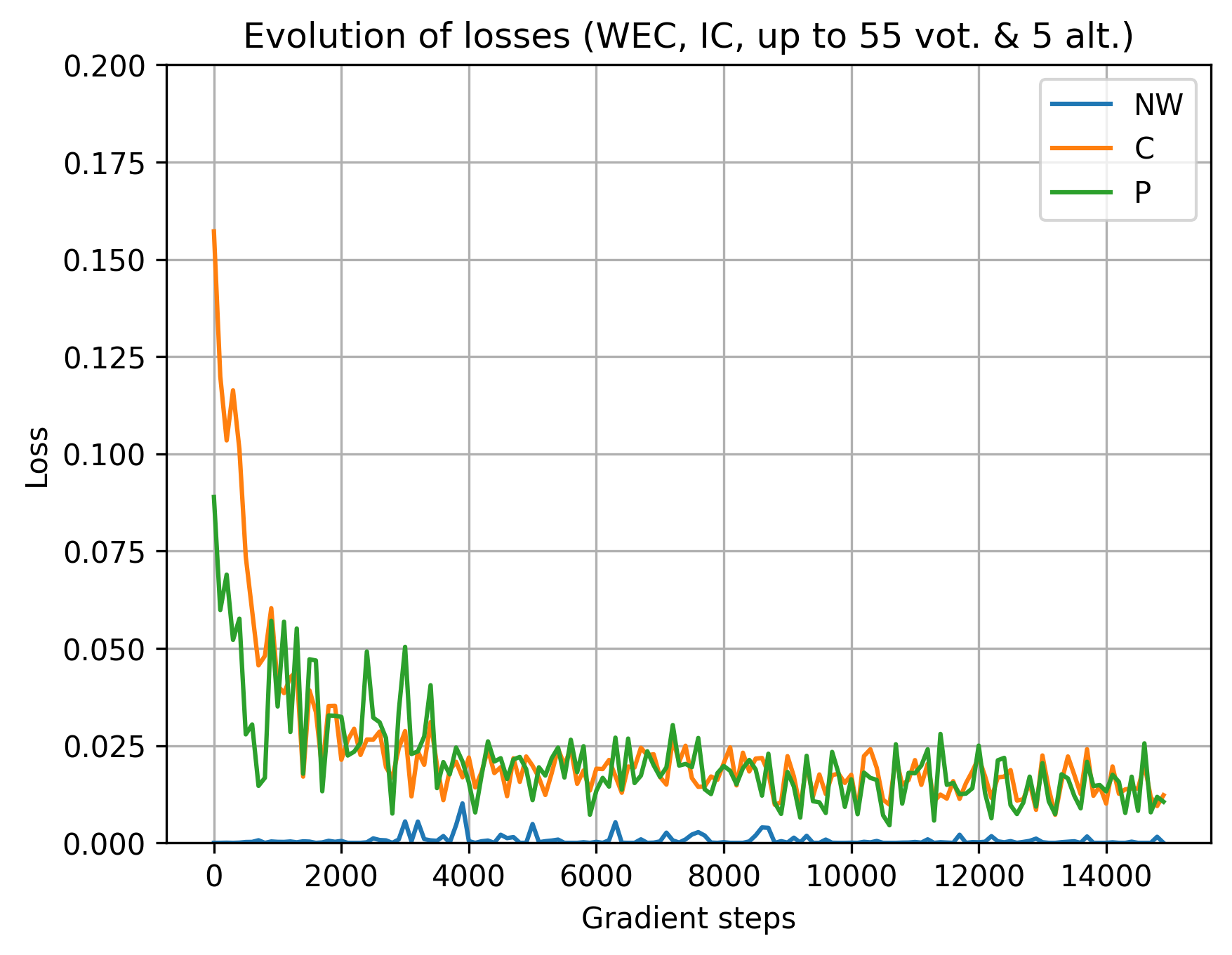}
\Description{Three plots describing the loss evolution during experiment~3.}
\caption{
The loss evolution during experiment~3. The run times of optimization were as follows: 5h~29min (MLP), 6h~08min (CNN), 2h~05min (WEC).
}
\label{fig: exp3 app loss evolution MLP CNN WEC}
\end{center}
\end{figure*}

Figure~\ref{fig: exp3 app loss evolution MLP CNN WEC} shows the evolution of the semantic losses during optimization and the run times, showing that the WEC not only achieves the best results but also does so most quickly.

\subsection{The Experiment with Mallows}

Table~\ref{tbl: exp 3 Mallows} shows the result of the experiment from Section~\ref{ssec: experiment 3} but with Mallows sampling (instead of IC sampling), using the parameters discussed in Section~\ref{ssec: voting theoretic parameters}.
Table~\ref{tbl: exp 3 similarities Mallows} shows how similar the best model is to its closest rules.
Figure~\ref{fig: exp 3 weak differing profile Mallows} presents a profile where the model differs from all considered rules.
Table~\ref{tbl: exp 3 averaged training Mallows} shows the statistical robustness of the axiom satisfaction achieved by the model.

\begin{table*}
\begin{footnotesize}
\begin{center}
\begin{tabular}{lcccccc}
\toprule
                         &  Anon.\  &   Neut.\  &  Condorcet  &  Pareto  &  Indep.\  &  Average  \\
\midrule
 Plurality               &   100   &   100    &     83.0      &   100    &   30.2   &  82.7  \\
 Borda                   &   100   &   100    &    92.8     &   100    &   32.8   &  85.1  \\
 Anti-Plurality          &   100   &   100    &    76.5     &   100    &   26.2   &  80.5  \\
 Copeland                &   100   &   100    &     100     &   100    &   27.8   &  85.5  \\
 Llull                   &   100   &   100    &     100     &   100    &   26.2   &  85.2  \\
 Uncovered Set           &   100   &   100    &     100     &   100    &   29.5   &  85.9  \\
 Top Cycle               &   100   &   100    &     100     &   100    &   25.2   &   85.0   \\
 Banks                   &   100   &   100    &     100     &   100    &   25.2   &   85.0   \\
 Stable Voting           &   100   &   100    &     100     &   100    &    39.0    &  87.8  \\
 Blacks                  &   100   &   100    &     100     &   100    &   33.8   &  86.8  \\
 Instant Runoff TB       &   100   &   100    &    96.8     &   100    &    29.0    &  85.2  \\
 PluralityWRunoff PUT    &   100   &   100    &     94.0      &   100    &    27.0    &  84.2  \\
 Coombs                  &   100   &   100    &    95.5     &   100    &   30.2   &  85.2  \\
 Baldwin                 &   100   &   100    &     100     &   100    &   39.2   &  87.9  \\
 Weak Nanson             &   100   &   100    &     100     &   100    &   33.8   &  86.8  \\
 Kemeny-Young            &   100   &   100    &     100     &   100    &   38.2   &  87.7  \\
\midrule 
 MLP p (NW, A, C, P, I)  &  78.8   &    76.0    &     94.0      &   100    &   38.8   &  77.5  \\
 MLP n (NW, A, C, P, I)  &  90.8   &   100    &    94.2     &   100    &   36.2   &  84.2  \\
 MLP na (NW, A, C, P, I) &  92.5   &   89.5   &    92.5     &   100    &   33.5   &  81.6  \\
 CNN p (NW, A, C, P, I)  &  80.5   &   68.8   &    94.5     &   100    &   38.8   &  76.5  \\
 CNN n (NW, A, C, P, I)  &  91.5   &   100    &     95.0      &   100    &   42.2   &  85.8  \\
 CNN na (NW, A, C, P, I) &  88.0    &   83.8   &     94.0      &   100    &   36.5   &  80.5  \\
 WEC p (NW, C, P)        &   100   &   65.5   &    91.8     &   100    &   37.8   &   79   \\
 WEC n (NW, C, P)        &   100   &   100    &     97.0      &   100    &    44.0    &  88.2  \\
\bottomrule
\end{tabular}
\end{center}
\end{footnotesize}
\caption{Mallows sampling: Axiom satisfaction of different rules (top part of the table) and models (bottom part of the table). Otherwise like Table~\ref{tbl: exp 3 IC} from the main text.
}
\label{tbl: exp 3 Mallows}
\end{table*}

\begin{table} 
\begin{footnotesize}
\begin{center}
\begin{tabular}{lcccccc}
\toprule
\emph{Identity accuracy}	&  WEC n  &  Stable Voting  &  Blacks  &  Borda  &  Weak Nanson  &  Copeland  \\
\hline
 WEC n         &   100   &      89.1       &   89.4   &  88.3   &     87.3      &    87.2    \\
 Stable Voting &  \phantom{89.1}   &       100       &  95.61   &  91.04  &     93.47     &   92.16    \\
 Blacks        &  \phantom{89.4}   &      \phantom{95.61}      &   100    &  95.43  &     91.71     &   90.82    \\
 Borda         &  \phantom{88.3}   &      \phantom{91.04}      &  \phantom{95.43}   &   100   &     87.14     &   86.25    \\
 Weak Nanson   &  \phantom{87.3}   &      \phantom{93.47}      &  \phantom{91.71}   &  \phantom{87.14}  &      100      &   92.08    \\
 Copeland      &  \phantom{87.2}   &      \phantom{92.16}      &  \phantom{90.82}   &  \phantom{86.25}  &     \phantom{92.08}     &    100     \\
\bottomrule
\end{tabular}

\medskip
\begin{tabular}{lcccccc}
\toprule
\emph{Subset accuracy}	&  WEC n  &  Stable Voting  &  Blacks  &  Borda  &  Weak Nanson  &  Copeland  \\
\hline
 WEC n         &   100   &      91.7       &   92.2   &  92.5   &     92.1      &    94.4    \\
 Stable Voting &  95.8   &       100       &  97.09   &  94.4   &     97.78     &   99.49    \\
 Blacks        &  95.8   &      96.63      &   100    &  97.31  &     95.96     &   97.97    \\
 Borda         &  94.2   &      92.06      &  95.43   &   100   &     91.39     &    93.4    \\
 Weak Nanson   &  92.7   &      94.5       &  93.02   &  90.33  &      100      &    97.4    \\
 Copeland      &  91.3   &      92.23      &   91.6   &  88.91  &     94.17     &    100     \\
\bottomrule
\end{tabular}
\end{center}
\end{footnotesize}
\caption{Mallows sampling: Similarities between the rules. Computed on 10,000 sampled profiles. Otherwise like Table~\ref{tbl: exp 3 similarities IC} from the main text.
}
\label{tbl: exp 3 similarities Mallows}
\end{table}

\floatstyle{boxed}
\restylefloat{figure}
\begin{figure}
\centering
\begin{footnotesize}
\begin{tabular}{cccc}
\toprule
  1    &  2    &  3    &  4    \\
\midrule
  $c$  &  $b$  &  $b$  &  $c$  \\
  $d$  &  $d$  &  $d$  &  $a$  \\
  $a$  &  $a$  &  $c$  &  $b$  \\
  $b$  &  $c$  &  $a$  &  $d$  \\
\bottomrule
\end{tabular}\\
\smallskip
\begin{tabular}{rl} 
$\{ b \}$ & neutrality-averaged WEC, with sigmoids (rounded)
$a$:.24, 
$b$:.51, 
$c$:.49, 
$d$:.32 
\\  
$\{ c \}$ & Instant Runoff \\
$\{ b, c \}$ & \parbox[t]{\textwidth - 5cm}{Plurality, Borda, Copeland, Llull, Uncovered Set, Stable Voting, Blacks, PluralityWRunoff PUT, Baldwin, Weak Nanson, Kemeny-Young}  \\
$\{ b, c, d \}$ & Banks  \\
$\{ a, b, c, d \}$ & Anti-Plurality, Top Cycle, Coombs \\
\end{tabular}
\end{footnotesize}
\Description{A profile on which the `WEC n' model weakly disagrees (Mallows sampling).}
\caption{Mallows sampling: Profile where the `WEC n' model weakly disagrees (i.e. non-identical winning sets) with existing voting rules.}
\label{fig: exp 3 weak differing profile Mallows}
\end{figure}
\floatstyle{plain}
\restylefloat{figure}

\begin{table*}
\begin{footnotesize}
\begin{center}
\begin{tabular}{lcccccc}
\toprule
                  &  Anon.  &   Neut.  &  Condorcet  &  Pareto  &  Indep.  &  Avg.  \\
\midrule
 Stable Voting    &   100   &   100    &     100     &   100    &  38.14   & 87.62  \\
 Blacks           &   100   &   100    &     100     &   100    &  34.04   & 86.84  \\
 Borda            &   100   &   100    &    94.16    &   100    &  35.02   & 85.84  \\
 Weak Nanson      &   100   &   100    &     100     &   100    &  37.18   & 87.46  \\
 Copeland         &   100   &   100    &     100     &   100    &   27.5   & 85.48  \\
 WEC n (NW, C, P) &   100   &   100    &    94.92    &   100    &  41.72   & 87.34  \\
\bottomrule
\end{tabular}
\end{center}
\end{footnotesize}
\caption{
Mallows sampling: Take the average over 5 runs of checking the axiom satisfaction of the `WEC n' model and its closest rules.
}
\label{tbl: exp 3 averaged training Mallows}
\end{table*}

\subsection{The Experiment with Urn}

Table~\ref{tbl: exp 3 Urn} shows the result of the experiment from Section~\ref{ssec: experiment 3} but with Urn sampling (instead of IC sampling).
Table~\ref{tbl: exp 3 similarities Urn} shows how similar the best model is to its closest rules.
Figure~\ref{fig: exp 3 strong differing profile Urn} presents a profile where the model strongly differs (i.e., had non-intersecting winning sets) from its five closest rules.
Table~\ref{tbl: exp 3 averaged training Urn} shows the statistical robustness of the axiom satisfaction achieved by the model.

\begin{table*}
\begin{footnotesize}
\begin{center}
\begin{tabular}{lcccccc}
\toprule
                         &  Anon.  &   Neut.  &  Condorcet  &  Pareto  &  Indep.  &  Avg.  \\
\midrule
 Plurality               &   100   &   100    &    84.2     &   100    &   24.5   &  81.8  \\
 Borda                   &   100   &   100    &    94.8     &   100    &    35    &  85.9  \\
 Anti-Plurality          &   100   &   100    &    76.8     &   100    &    25    &  80.3  \\
 Copeland                &   100   &   100    &     100     &   100    &   28.2   &  85.6  \\
 Llull                   &   100   &   100    &     100     &   100    &   27.3   &  85.5  \\
 Uncovered Set           &   100   &   100    &     100     &   100    &   25.2   &   85   \\
 Top Cycle               &   100   &   100    &     100     &   100    &   27.8   &  85.5  \\
 Banks                   &   100   &   100    &     100     &   100    &   27.5   &  85.5  \\
 Stable Voting           &   100   &   100    &     100     &   100    &    38    &  87.6  \\
 Blacks                  &   100   &   100    &     100     &   100    &   34.2   &  86.9  \\
 Instant Runoff TB       &   100   &   100    &     97      &   100    &   28.2   &   85   \\
 PluralityWRunoff PUT    &   100   &   100    &    94.2     &   100    &   26.2   &  84.1  \\
 Coombs                  &   100   &   100    &    96.5     &   100    &   28.5   &   85   \\
 Baldwin                 &   100   &   100    &     100     &   100    &    39    &  87.8  \\
 Weak Nanson             &   100   &   100    &     100     &   100    &   38.5   &  87.7  \\
 Kemeny-Young            &   100   &   100    &     100     &   100    &   39.2   &  87.9  \\
\midrule
 MLP p (NW, A, C, P, I)  &  79.8   &   74.2   &    92.8     &   100    &   34.8   &  76.3  \\
 MLP n (NW, A, C, P, I)  &   91    &   100    &    93.8     &   100    &   38.8   &  84.7  \\
 MLP na (NW, A, C, P, I) &  91.8   &    91    &    93.5     &   100    &    35    &  82.2  \\
 CNN p (NW, A, C, P, I)  &  82.8   &   73.8   &     94      &   100    &   37.8   &  77.6  \\
 CNN n (NW, A, C, P, I)  &   90    &   100    &    94.5     &   100    &   34.5   &  83.8  \\
 CNN na (NW, A, C, P, I) &   90    &   91.5   &    93.2     &   100    &   34.5   &  81.8  \\
 WEC p (NW, C, P)        &   100   &   75.2   &     93      &   100    &   37.8   &  81.2  \\
 WEC n (NW, C, P)        &   100   &   100    &    93.5     &   100    &    39    &  86.5  \\
\bottomrule
\end{tabular}
\end{center}
\end{footnotesize}
\caption{Urn sampling: Axiom satisfaction of different rules (top part of the table) and models (bottom part of the table). Otherwise like Table~\ref{tbl: exp 3 IC} from the main text.
}
\label{tbl: exp 3 Urn}
\end{table*}

\begin{table} 
\begin{footnotesize}
\begin{center}
\begin{tabular}{lcccccc}
\toprule
\emph{Identity accuracy}	&  WEC n  &  Blacks  &  Stable Voting  &  Borda  &  Copeland  &  Weak Nanson  \\
\midrule
 WEC n         &   100   &   88.9   &      88.5       &  88.2   &    86.5    &     86.3      \\
 Blacks        &  \phantom{88.9}   &   100    &      95.52      &  95.32  &   90.92    &     91.6      \\
 Stable Voting &  \phantom{88.5}   &  \phantom{95.52}   &       100       &  90.84  &   92.17    &     93.61     \\
 Borda         &  \phantom{88.2}   &  \phantom{95.32}   &      \phantom{90.84}      &   100   &   86.24    &     86.92     \\
 Copeland      &  \phantom{86.5}   &  \phantom{90.92}   &      \phantom{92.17}      &  \phantom{86.24}  &    100     &     92.09     \\
 Weak Nanson   &  \phantom{86.3}   &   \phantom{91.6}   &      \phantom{93.61}      &  \phantom{86.92}  &   \phantom{92.09}    &      100      \\
\bottomrule
\end{tabular}

\medskip
\begin{tabular}{lcccccc}
\toprule
\emph{Subset accuracy}	&  WEC n  &  Blacks  &  Stable Voting  &  Borda  &  Copeland  &  Weak Nanson  \\
\midrule
 WEC n         &   100   &   92.3   &      91.4       &  93.2   &    94.3    &     91.3      \\
 Blacks        &  95.1   &   100    &      96.44      &  97.48  &   98.07    &     95.4      \\
 Stable Voting &  94.9   &  97.03   &       100       &  94.51  &   99.47    &     97.48     \\
 Borda         &  93.9   &  95.32   &      91.76      &   100   &   93.39    &     90.72     \\
 Copeland      &  90.4   &  91.83   &      92.19      &  89.31  &    100     &     93.9      \\
 Weak Nanson   &  91.7   &  93.04   &      94.65      &  90.52  &   97.43    &      100      \\
\bottomrule
\end{tabular}
\end{center}
\end{footnotesize}
\caption{Urn sampling: Similarities between the rules. Computed on 10,000 sampled profiles. Otherwise like Table~\ref{tbl: exp 3 similarities IC} from the main text.
}
\label{tbl: exp 3 similarities Urn}
\end{table}

\floatstyle{boxed}
\restylefloat{figure}
\begin{figure}
\centering
\begin{footnotesize}
\begin{tabular}{ccccc}
\toprule
  1  &  2  &  3  &  4  &  5  \\
\midrule
  $b$  &  $a$  &  $b$  &  $d$  &  $e$  \\
  $c$  &  $c$  &  $e$  &  $a$  &  $c$  \\
  $d$  &  $e$  &  $d$  &  $c$  &  $a$  \\
  $e$  &  $d$  &  $a$  &  $b$  &  $b$  \\
  $a$  &  $b$  &  $c$  &  $e$  &  $d$  \\
\bottomrule
\end{tabular}\\
\smallskip
\begin{tabular}{rl} 
$\{ b \}$ & neutrality-averaged WEC, with sigmoids (rounded)
$a$:.44, 
$b$:.50, 
$c$:.48, 
$d$:.40, 
$e$:.44  
\\  
$\{ c \}$ & Blacks, Stable Voting, Borda, Copeland, Weak Nanson, Llull \\
$\{ b \}$ & Plurality, Instant Runoff TB \\
$\{ a \}$ & Baldwin \\
$\{ a, b \}$ & PluralityWRunoff PUT \\
$\{ a, c \}$ & Kemeny-Young \\
$\{ a, c, e \}$ & Uncovered Set, Banks\\
$\{ a, b, c, d, e \}$ & Anti-Plurality, Top Cycle, Coombs \\
\end{tabular}
\end{footnotesize}
\Description{A profile on which the `WEC n' model strongly disagrees (Urn sampling).}
\caption{Urn sampling: Profile where the `WEC n' model strongly disagrees (i.e. non-intersecting winning sets) with its 5 closest rules (among the remaining rules it only agrees with Plurality and Instant Runoff TB).}
\label{fig: exp 3 strong differing profile Urn}
\end{figure}
\floatstyle{plain}
\restylefloat{figure}

\begin{table*}
\begin{footnotesize}
\begin{center}
\begin{tabular}{lcccccc}
\toprule
                  &  Anon.  &   Neut.  &  Condorcet  &  Pareto  &  Indep.  &  Avg.  \\
\midrule
 Blacks           &   100   &   100    &     100     &   100    &  34.38   &  86.9  \\
 Stable Voting    &   100   &   100    &     100     &   100    &   39.2   & 87.84  \\
 Borda            &   100   &   100    &    93.46    &   100    &  35.66   & 85.84  \\
 Copeland         &   100   &   100    &     100     &   100    &   26.9   & 85.38  \\
 Weak Nanson      &   100   &   100    &     100     &   100    &  37.82   & 87.54  \\
 WEC n (NW, C, P) &   100   &   100    &    95.68    &   100    &  38.16   & 86.76  \\
\bottomrule
\end{tabular}
\end{center}
\end{footnotesize}
\caption{
Urn sampling: Take the average over 5 runs of checking the axiom satisfaction of the `WEC n' model and its closest rules.
}
\label{tbl: exp 3 averaged training Urn}
\end{table*}

\subsection{The Experiment with Euclidean}

Table~\ref{tbl: exp 3 Euclidean} shows the result of the experiment from Section~\ref{ssec: experiment 3} but with Euclidean sampling (instead of IC sampling).
Table~\ref{tbl: exp 3 similarities Euclidean} shows how similar the best model is to its closest rules.
Figure~\ref{fig: exp 3 weak differing profile Euclidean} presents a profile where the model differs from all considered rules.
Table~\ref{tbl: exp 3 averaged training Euclidean} shows the statistical robustness of the axiom satisfaction achieved by the model.

\begin{table*}
\begin{footnotesize}
\begin{center}
\begin{tabular}{lcccccc}
\toprule
                         &  Anon.  &   Neut.  &  Condorcet  &  Pareto  &  Indep.  &  Avg.  \\
\midrule
 Plurality               &   100   &   100    &    79.8     &   100    &   25.5   &   81   \\
 Borda                   &   100   &   100    &    93.8     &   100    &   37.2   &  86.2  \\
 Anti-Plurality          &   100   &   100    &    77.2     &   100    &    25    &  80.5  \\
 Copeland                &   100   &   100    &     100     &   100    &   29.5   &  85.9  \\
 Llull                   &   100   &   100    &     100     &   100    &   29.8   &   86   \\
 Uncovered Set           &   100   &   100    &     100     &   100    &   26.2   &  85.2  \\
 Top Cycle               &   100   &   100    &     100     &   100    &   28.7   &  85.8  \\
 Banks                   &   100   &   100    &     100     &   100    &    28    &  85.6  \\
 Stable Voting           &   100   &   100    &     100     &   100    &   39.5   &  87.9  \\
 Blacks                  &   100   &   100    &     100     &   100    &    37    &  87.4  \\
 Instant Runoff TB       &   100   &   100    &    96.8     &   100    &    30    &  85.4  \\
 PluralityWRunoff PUT    &   100   &   100    &    94.8     &   100    &   26.5   &  84.2  \\
 Coombs                  &   100   &   100    &     96      &   100    &   26.2   &  84.5  \\
 Baldwin                 &   100   &   100    &     100     &   100    &   40.2   &   88   \\
 Weak Nanson             &   100   &   100    &     100     &   100    &    40    &   88   \\
 Kemeny-Young            &   100   &   100    &     100     &   100    &   36.5   &  87.3  \\
\midrule 
 MLP p (NW, A, C, P, I)  &  79.2   &    78    &    92.2     &   100    &    36    &  77.1  \\
 MLP n (NW, A, C, P, I)  &  87.2   &   100    &    93.8     &   100    &   36.2   &  83.5  \\
 MLP na (NW, A, C, P, I) &  90.5   &   91.5   &    93.5     &   100    &   31.8   &  81.5  \\
 CNN p (NW, A, C, P, I)  &  83.2   &   74.5   &    93.2     &   100    &    35    &  77.2  \\
 CNN n (NW, A, C, P, I)  &  89.8   &   100    &     92      &   100    &   34.8   &  83.3  \\
 CNN na (NW, A, C, P, I) &   86    &   90.2   &    92.8     &   100    &   30.5   &  79.9  \\
 WEC p (NW, C, P)        &   100   &    75    &    96.8     &   100    &   38.5   &   82   \\
 WEC n (NW, C, P)        &   100   &   100    &    97.8     &   100    &   42.2   &   88   \\
\bottomrule
\end{tabular}
\end{center}
\end{footnotesize}
\caption{Euclidean sampling: Axiom satisfaction of different rules (top part of the table) and models (bottom part of the table). Otherwise like Table~\ref{tbl: exp 3 IC} from the main text.
}
\label{tbl: exp 3 Euclidean}
\end{table*}

\begin{table} 
\begin{footnotesize}
\begin{center}
\begin{tabular}{lcccccc}
\toprule
\emph{Identity accuracy}	&  WEC n  &  Stable Voting  &  Blacks  &  Weak Nanson  &  Copeland  &  Borda  \\
\hline
 WEC n         &   100   &      92.2       &   91.5   &     89.5      &    88.7    &  89.1   \\
 Stable Voting &  \phantom{92.2}   &       100       &  95.65   &     93.22     &   92.24    &  91.05  \\
 Blacks        &  \phantom{91.5}   &      \phantom{95.65}      &   100    &     91.37     &   90.85    &  95.4   \\
 Weak Nanson   &  \phantom{89.5}   &      \phantom{93.22}      &  \phantom{91.37}   &      100      &   92.46    &  86.77  \\
 Copeland      &  \phantom{88.7}   &      \phantom{92.24}      &  \phantom{90.85}   &     \phantom{92.46}     &    100     &  86.25  \\
 Borda         &  \phantom{89.1}   &      \phantom{91.05}      &   \phantom{95.4}   &     \phantom{86.77}     &   \phantom{86.25}    &   100   \\
\hline
\end{tabular}

\medskip
\begin{tabular}{lcccccc}
\toprule
\emph{Subset accuracy}	&  WEC n  &  Stable Voting  &  Blacks  &  Weak Nanson  &  Copeland  &  Borda  \\
\midrule
 WEC n         &   100   &      94.9       &   94.5   &     94.6      &    96.7    &  93.9   \\
 Stable Voting &  95.6   &       100       &  97.19   &     97.5      &   99.58    &  94.57  \\
 Blacks        &  94.7   &      96.69      &   100    &     95.61     &   97.92    &  97.38  \\
 Weak Nanson   &   92    &      94.24      &  92.82   &      100      &   97.53    &  90.2   \\
 Copeland      &  90.8   &      92.3       &  91.67   &     94.39     &    100     &  89.05  \\
 Borda         &  92.1   &      92.09      &   95.4   &     91.01     &   93.32    &   100   \\
\bottomrule
\end{tabular}
\end{center}
\end{footnotesize}
\caption{Euclidean sampling: Similarities between the rules. Computed on 10,000 sampled profiles. Otherwise like Table~\ref{tbl: exp 3 similarities IC} from the main text.
}
\label{tbl: exp 3 similarities Euclidean}
\end{table}

\floatstyle{boxed}
\restylefloat{figure}
\begin{figure}
\centering
\begin{footnotesize}
\begin{tabular}{cc}
\toprule
  1  &  2  \\
\midrule
  $d$  &  $b$  \\
  $a$  &  $e$  \\
  $c$  &  $a$  \\
  $b$  &  $c$  \\
  $e$  &  $d$  \\
\bottomrule
\end{tabular}\\
\smallskip
\begin{tabular}{rl} 
$\{ b \}$ & neutrality-averaged WEC, with sigmoids (rounded)
$a$:.34, 
$b$:.50, 
$c$:.13, 
$d$:.30, 
$e$:.13 
\\  
$\{a\}$ & Coombs \\
$\{d\}$ & Instant Runoff TB \\
$\{a,b\}$ & Weak Nanson \\
$\{b,d\}$ & Plurality, PluralityWRunoff PUT \\
$\{a,b\}$ & Borda, Copeland, Stable Voting, Blacks \\
$\{a, b, d\}$ & Llull, Uncovered Set, Banks, Baldwin, Kemeny-Young \\
$\{a,b,c\}$ & Anti-Plurality \\
$\{a,b,c,d,e\}$ & Top Cycle \\
\end{tabular}
\end{footnotesize}
\Description{A profile on which the `WEC n' model strongly disagrees (Euclidean sampling).}
\caption{Euclidean sampling: Profile where the `WEC n' model weakly disagrees (i.e. non-identical winning sets) with existing voting rules.}
\label{fig: exp 3 weak differing profile Euclidean}
\end{figure}
\floatstyle{plain}
\restylefloat{figure}

\begin{table*}
\begin{footnotesize}
\begin{center}
\begin{tabular}{lcccccc}
\toprule
                  &  Anon.  &   Neut.  &  Condorcet  &  Pareto  &  Indep.  &  Avg.  \\
\midrule
 Stable Voting    &   100   &   100    &     100     &   100    &  39.12   &  87.8  \\
 Blacks           &   100   &   100    &     100     &   100    &  34.98   & 86.98  \\
 Weak Nanson      &   100   &   100    &     100     &   100    &  39.32   & 87.86  \\
 Copeland         &   100   &   100    &     100     &   100    &  27.68   & 85.54  \\
 Borda            &   100   &   100    &    95.08    &   100    &  35.46   & 86.08  \\
 WEC n (NW, C, P) &   100   &   100    &    97.74    &   100    &  45.16   & 88.58  \\
\bottomrule
\end{tabular}
\end{center}
\end{footnotesize}
\caption{
Euclidean sampling: Take the average over 5 runs of checking the axiom satisfaction of the `WEC n' model and its closest rules.
}
\label{tbl: exp 3 averaged training Euclidean}
\end{table*}

\subsection{Ablation Study}
\label{ssec: app ablation study}

\begin{figure*}
\begin{center}
\includegraphics[width=0.7\linewidth]{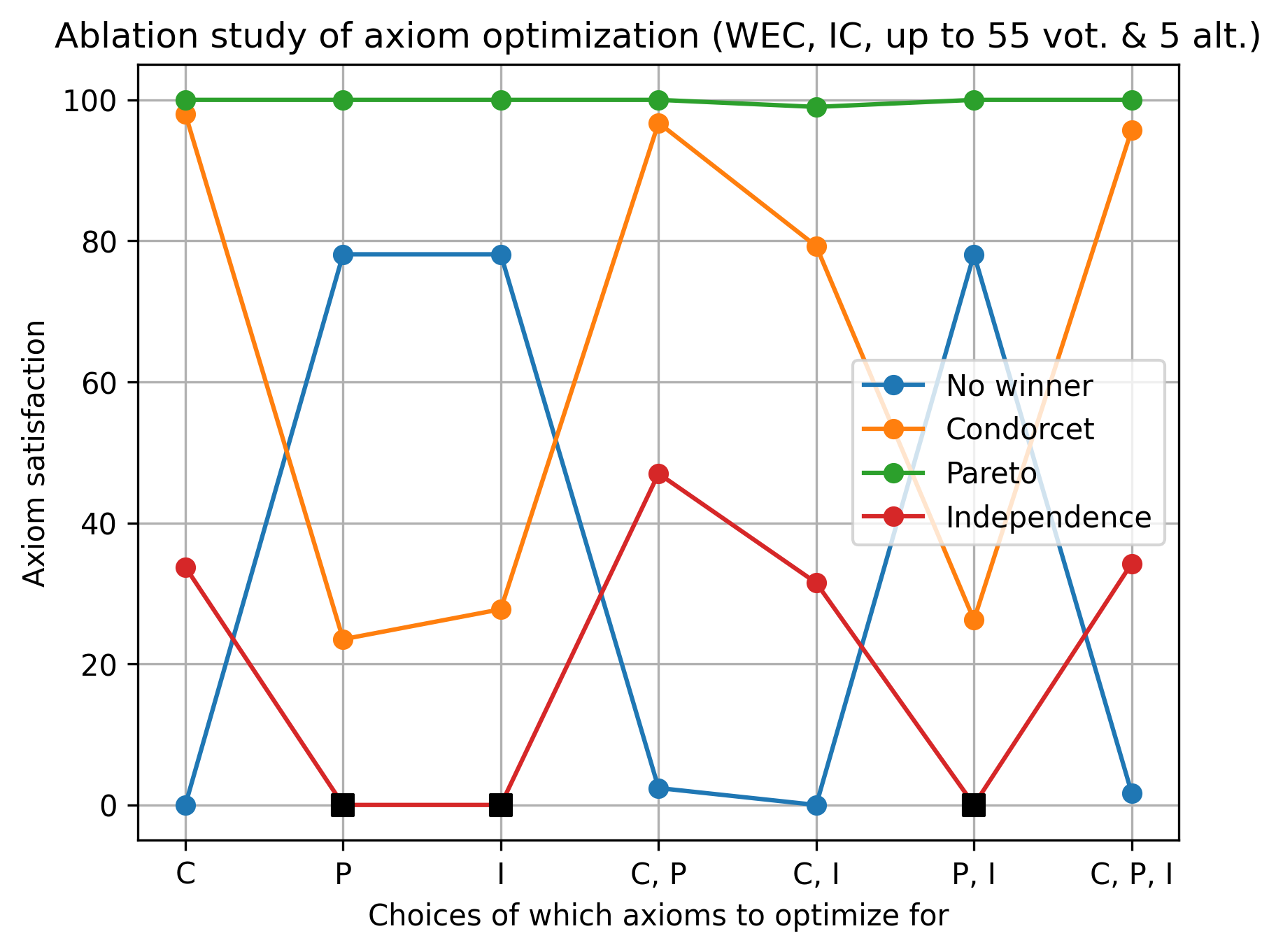}\\
\end{center}
\Description{A plot describing the ablation study in axiom optimization.}
\caption{Ablation study in axiom optimization with the neutrality-averaged WEC. For each possible (nonempty) choice of axioms to optimize for among Condorcet (C), Pareto (P), and independence (I), the achieved axiom satisfaction is shown. The `No winner' loss (NW) is always optimized for. Its reported satisfaction is $0$ if the model always outputs at least one winner. The axioms of anonymity and neutrality are not shown since they are satisfied by design. The black squares in the independence satisfaction indicate that the axiom was applicable on too few of the sampled test profiles to warrant an estimate. The best choice is C, P since it has the highest axiom satisfaction combined with a low `No winner' satisfaction.
}
\label{fig: exp3 app ablation study}
\end{figure*}

\begin{figure*}
\begin{center}
\includegraphics[width=0.43\linewidth]{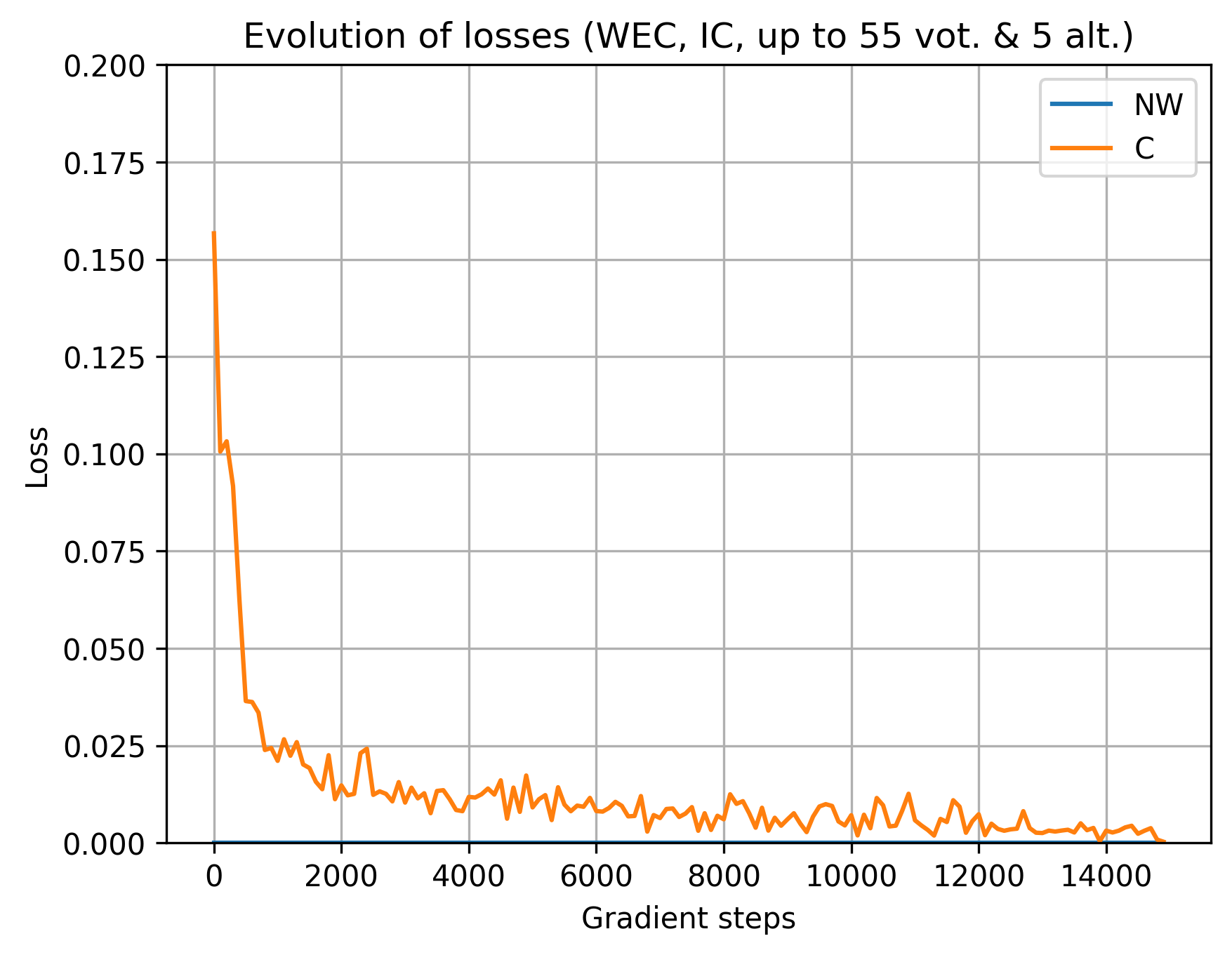}
\includegraphics[width=0.43\linewidth]{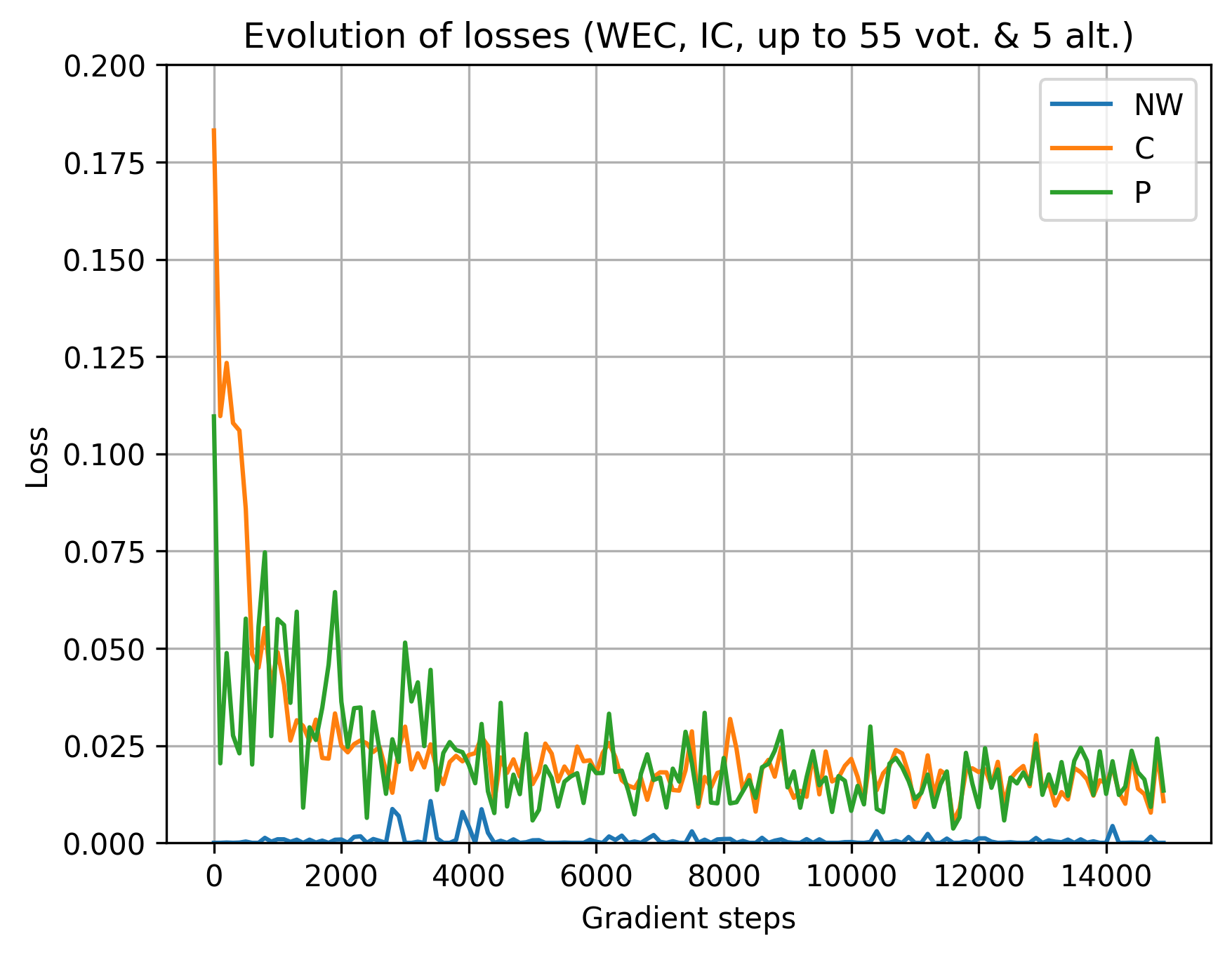}\\
\includegraphics[width=0.43\linewidth]{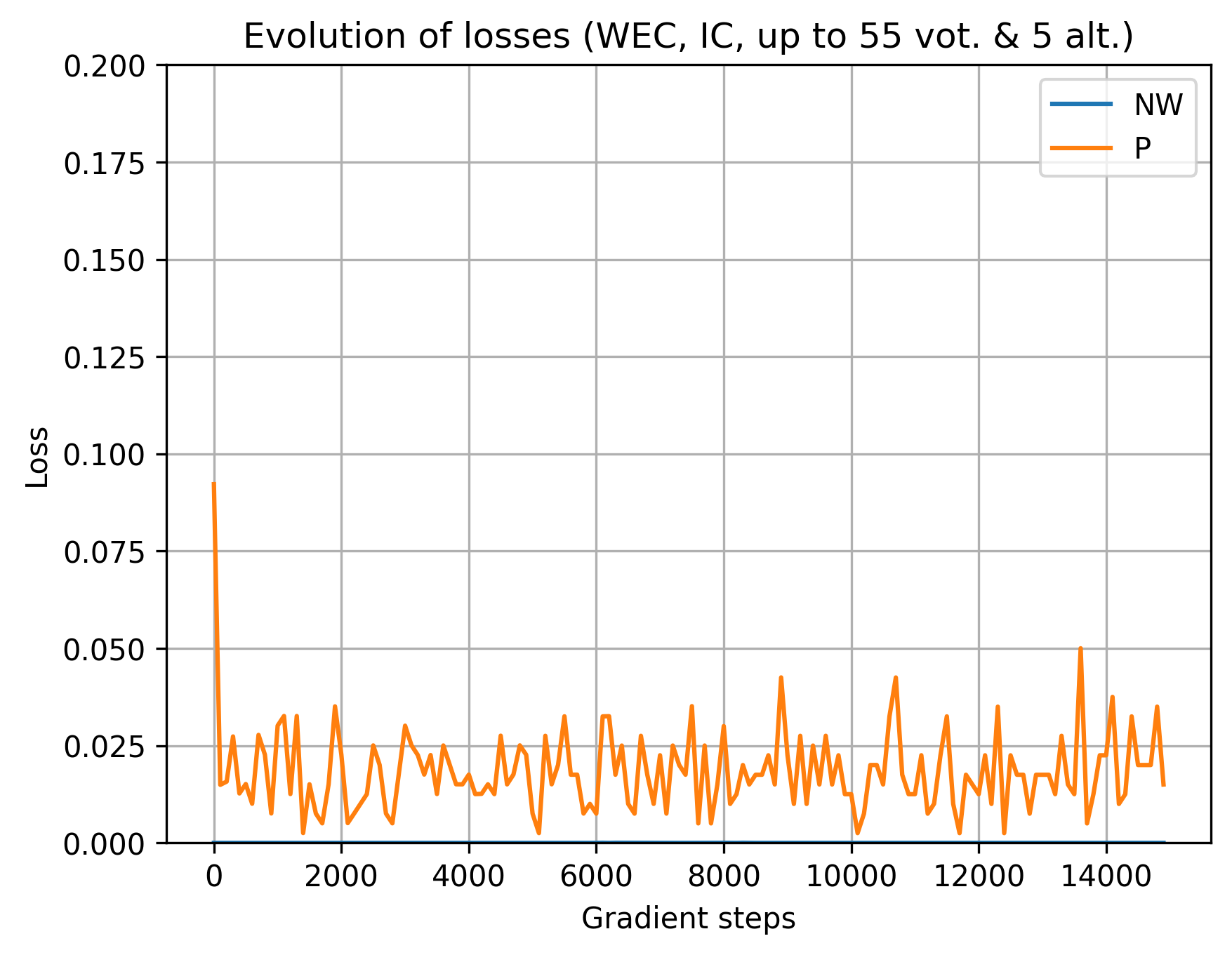}
\includegraphics[width=0.43\linewidth]{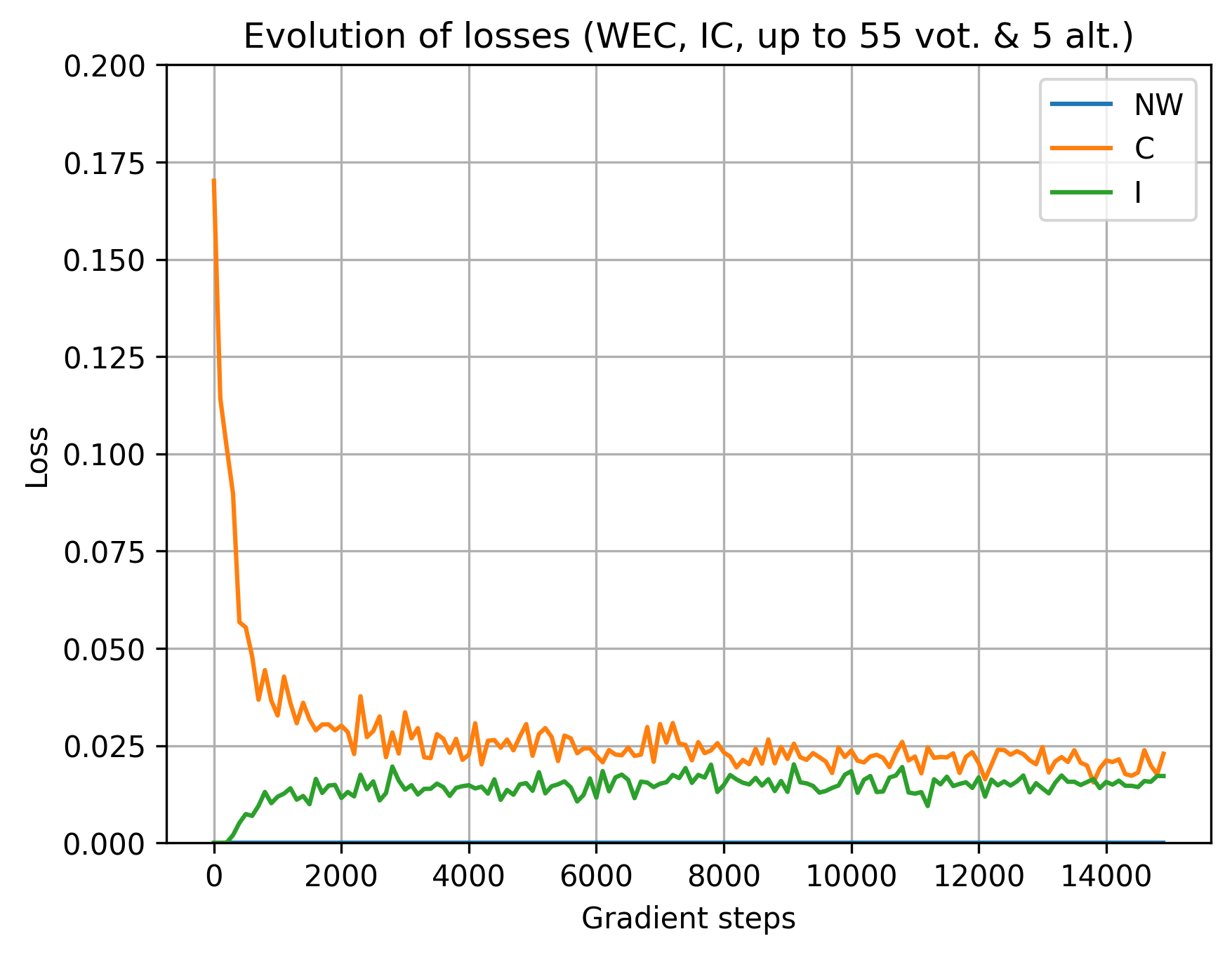}\\
\includegraphics[width=0.43\linewidth]{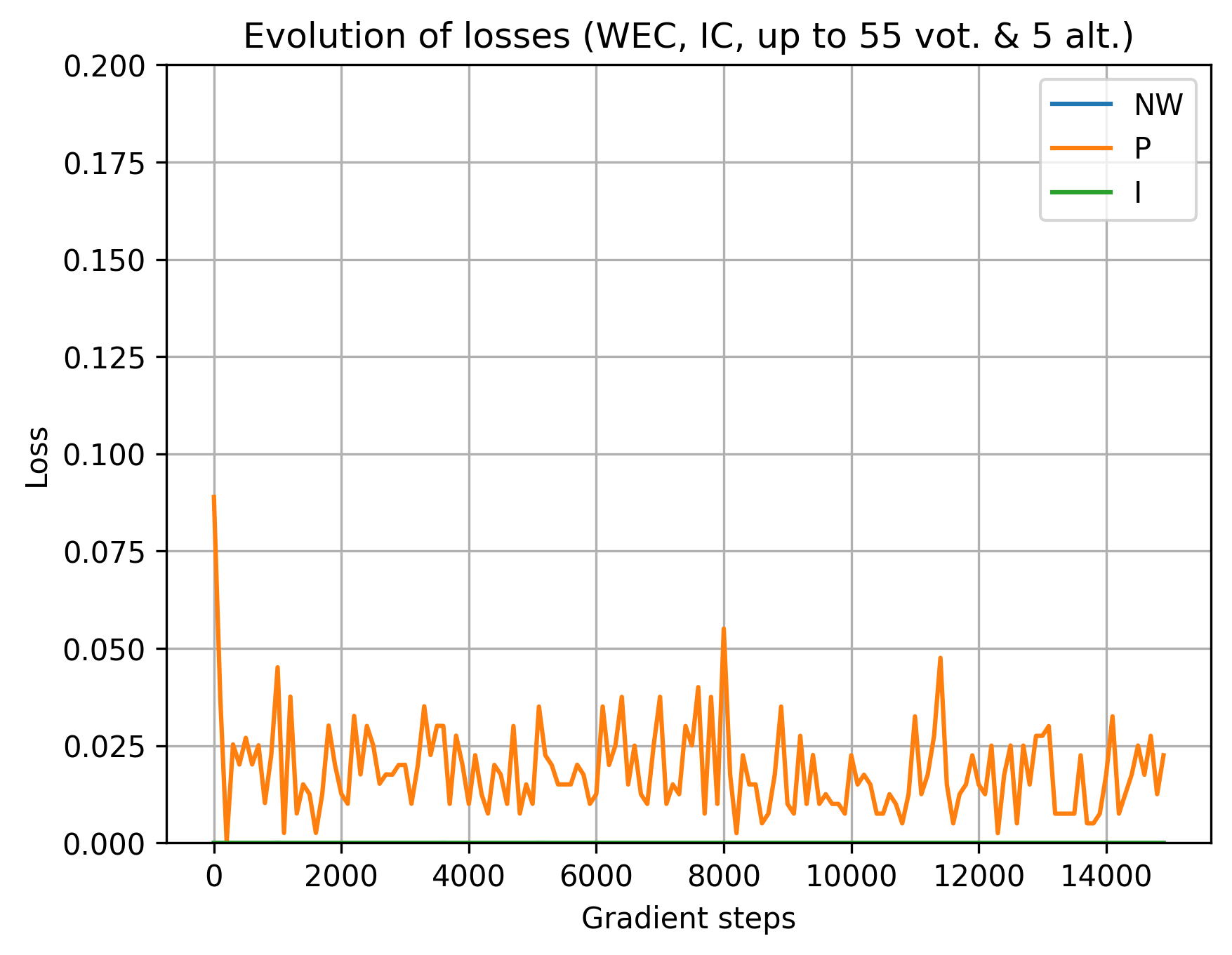}
\includegraphics[width=0.43\linewidth]{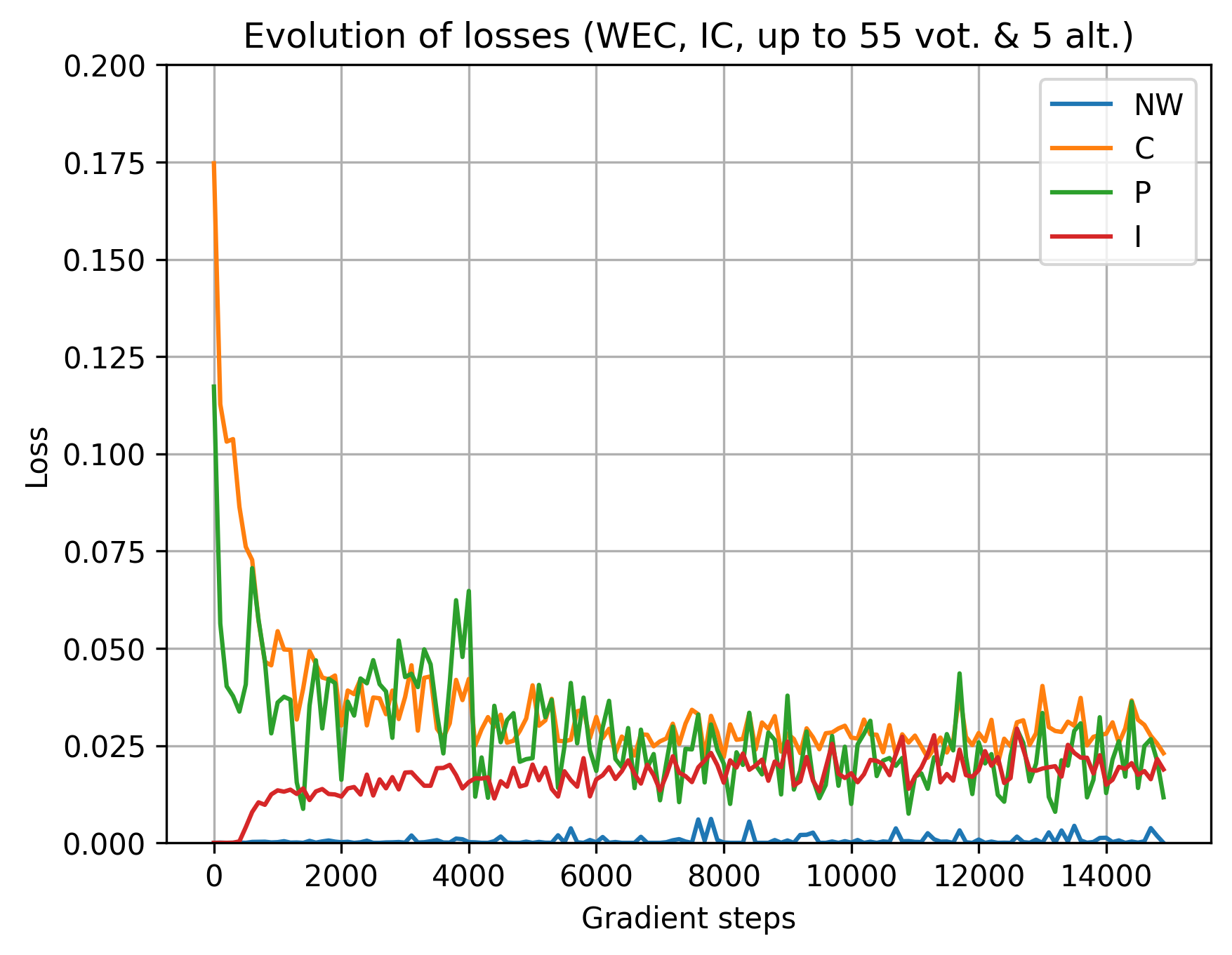}
\end{center}
\Description{Six plots describing the loss evolution during axiom optimization.}
\caption{The evolution of the losses during axiom optimization with the neutrality-averaged WEC, for each choice of which axioms to optimize among Condorcet (C), Pareto (P), and independence (I), with `no winner' (NW) always being optimized for. The loss curves for the NW$+$I-optimization are not shown, since they are so close to 0 that they are indistinguishable from the x-axis. In the other two cases that did not yield good axiom satisfaction---i.e., P and P, I in Figure~\ref{fig: exp3 app ablation study}---, the loss evolution also shows no convergence.}
\label{fig: exp3 app loss evolution}
\end{figure*}

Figure~\ref{fig: exp3 app ablation study} displays an ablation study in the choice of axioms to optimize for. For the best performing model, i.e., the neutrality-averaged WEC, there remain three axioms that can be optimized for: Condorcet (C), Pareto (P), and independence (I). The `No winner' loss (NW) is always needed to prevent the model from never outputting any winner. Which subset of the three axioms is the optimal choice for optimization? In the main text, we chose C and P. The figure shows that this choice indeed is the best one. 
Figure~\ref{fig: exp3 app loss evolution} shows, for each choice of axiom optimization, the evolution of the losses during training.

\FloatBarrier

\end{document}